\documentclass[acmsmall, screen, natbib, acmthm, nonacm=true]{acmart}

\usepackage{microtype}

\usepackage{bm}
\usepackage{mathrsfs} 
\usepackage{amsopn}
\usepackage{makecell}
\usepackage{booktabs}

\usepackage{amsmath,amsthm,amsfonts,mathtools}
\usepackage{bbm} %
\usepackage{enumitem}
\usepackage{natbib}
\usepackage{dsfont} %
\usepackage[normalem]{ulem} %
\usepackage{appendix}

\allowdisplaybreaks

\usepackage{graphicx}
\usepackage{wrapfig}
\usepackage{tikz} %
\usepackage{float}
\usepackage{caption}     % For custom captions
\usepackage{subcaption}

\newtheorem{theorem}{Theorem}
\newtheorem{corollary}{Corollary}
\newtheorem{lemma}{Lemma}

\newtheorem{proposition}{Proposition}
\newtheorem{definition}{Definition}

\newtheorem{assumption}{Assumption}

\global\long\def\d{\mathrm{d}}%
\global\long\def\E{\mathbb{E}}%
\global\long\def\R{\mathbb{R}}%

\global\long\def\law{\mathcal{L}}

\usepackage{xcolor}
\usepackage{todonotes} % 
\usepackage[showdeletions]{color-edits} % 
\addauthor[Yixuan]{yz}{blue} % 
\addauthor[Qiaomin]{qx}{red} % 
\addauthor[Yudong]{yc}{purple} %
\addauthor[Lucy]{lh}{magenta} % 
\begin{document}

\title{A Piecewise Lyapunov Analysis of Sub-quadratic SGD: Applications to Robust and Quantile Regression}

\author{Yixuan Zhang}
\affiliation{%
\institution{University of Wisconsin-Madison}
\department{Department of Industrial and Systems Engineering}
\city{Madison}
\state{Wisconsin}
\country{USA}}
\email{yzhang2554@wisc.edu}

\author{Dongyan (Lucy) Huo}
\affiliation{%
\institution{Cornell University}
\department{School of Operations Research and Information Engineering}
\city{Ithaca}
\state{New York}
\country{USA}}
\email{dh622@cornell.edu}

\author{Yudong Chen}
\affiliation{%
\institution{University of Wisconsin-Madison}
\department{Department of Computer Sciences}
\city{Madison}
\state{Wisconsin}
\country{USA}}
\email{yudong.chen@wisc.edu}

\author{Qiaomin Xie}
\affiliation{%
\institution{University of Wisconsin-Madison}
\department{Department of Industrial and Systems Engineering}
\city{Madison}
\state{Wisconsin}
\country{USA}}
\email{qiaomin.xie@wisc.edu} 

\begin{abstract}
Motivated by robust and quantile regression problems, {we investigate the stochastic gradient descent (SGD) algorithm} for minimizing an objective function $f$ that is locally strongly convex with a sub--quadratic tail. This setting covers many widely used online statistical methods. We introduce a novel piecewise Lyapunov function that enables us to handle functions $f$ with only first-order differentiability, which includes a wide range of popular loss functions such as Huber loss. Leveraging our proposed Lyapunov function, we derive finite-time moment bounds under general diminishing stepsizes, as well as constant stepsizes. We further establish the weak convergence, central limit theorem and bias characterization under constant stepsize, providing the first geometrical convergence result for sub--quadratic SGD. Our results have wide applications, especially in online statistical methods. In particular, we discuss two applications of our results. 1) Online robust regression: We consider a corrupted linear model with sub--exponential covariates and heavy--tailed noise. Our analysis provides convergence rates comparable to those for corrupted models with Gaussian covariates and noise. 2) Online quantile regression: Importantly, our results relax the common assumption in prior work that the conditional density is continuous and provide a more fine-grained analysis for the moment bounds.

\end{abstract}

\maketitle
\section{Introduction}

The problem of minimizing objective functions that are not strongly convex has garnered considerable attention across various domains, including modern statistical machine learning---such as matrix and tensor completion \cite{ge2015escaping, ge2016matrix, xia2021statistically}, deep neural networks \cite{lecun2015deep, jain2017non, maddox2019simple}, and robust statistics \cite{law1986robust, huber2011robust, loh2017statistical, loh2021scale}---as well as in optimization \cite{defazio2014saga, necoara2019linear} and stochastic approximation \cite{bach2013non}.

Specifically, in robust regression \cite{loh2017statistical,loh2021scale,bhatia2017consistent}, the goal is to recover the underlying model when the data is contaminated by outliers and/or corruption. In particular, it aims to find the global optimizer \(\theta^*\) of the population-level loss function:
\begin{align}\label{eq:robustintuition}
f(\theta) = \mathbb{E}[l(y - x^\top \theta)],
\end{align}
when given observations of independent and identically distributed (i.i.d.) data \(\{(x_n, y_n)\}_{n \geq 0}\) generated from a corrupted linear model: $y = x^\top \theta^* + \epsilon + s,$ where \(\epsilon\) represents the error and \(s\) denotes the corruption. In robust regression, the error $\epsilon$ is typically heavy--tailed without a second moment bound. It is well recognized that the classical squared loss function cannot handle heavy--tailed errors or corruption effectively. To encourage robustness, the field has been widely using loss functions \(l\) that have \emph{sub--quadratic tails} to assign less weight to outliers. Common robust loss functions include Huber~\cite{huber1992robust}, Pseudo-Huber loss~\cite{hartley2003multiple} and log-cosh loss~\cite{saleh2022statistical}. Importantly, the population-level loss $f$ inherits the sub--quadratic tail behavior of the loss function $l$. 
Meanwhile, we observe that \(f\) is strongly convex in a neighborhood of \(\theta^*\). The problem of minimizing objective functions with local strong convexity and sub--quadratic tails also arises in quantile regression~\cite{koenker2005quantile, hao2007quantile}. Quantile regression estimates the conditional quantile of the response given covariates, and is widely used in various applications~\cite{wang2014quantile,luo2013quantile,belloni2011ℓ}.

Motivated by the robust and quantile regression problems, we consider stochastic gradient descent (SGD) algorithms aimed at approximating the global minimum $\theta^* \in \mathbb{R}^d$ of a non-strongly convex function $f: \mathbb{R}^d \rightarrow \mathbb{R}$, using unbiased estimates of the function's gradients:
\begin{align}\label{eq:SGD}
    \theta_{n+1} = \theta_n - \alpha_n \left( \nabla f(\theta_n) + w_n(\theta_n) \right),
\end{align}
where $\nabla f$ denotes the population-level gradient of objective function $f$, $\{\alpha_n\}_{n \geq 0}$ are the stepsize sequence, and $\{w_n(\cdot)\}_{n \geq 0}$ are i.i.d. copies of a random field $w(\cdot)$ with $\mathbb{E}[w(\theta)] = 0$ for all $\theta \in \mathbb{R}^d$. In this work, we focus on a class of objective functions that are locally strongly convex around $\theta^*$ and exhibit a \textbf{sub--quadratic tail}. Throughout the paper, we refer to this function class as \emph{sub--quadratic functions} and, for brevity, we call the corresponding SGD procedure \emph{sub--quadratic SGD}. 

Sub--quadratic SGD covers many widely used online statistical algorithms, including those for online robust regression \cite{pesme2020online,godichon2024online} and online quantile regression \cite{chen2019quantile,wang2022renewable,jiang2022renewable,shen2024online}. Classical studies on robust regression \cite{bhatia2017consistent, loh2017statistical, loh2021scale} and quantile regression \cite{koenker2005quantile, hao2007quantile} primarily focus on processing complete datasets, which can be computationally inefficient and memory intensive. In many practical applications, however, data either arrives sequentially or is too large to be processed all at once, making SGD algorithms a natural and scalable alternative for the optimization problem. Research on online robust/quantile regression  remains relatively limited. Prior work on online robust regression \cite{pesme2020online} focused on corrupted linear models with Gaussian-distributed covariates and error. For online quantile regression, earlier work \cite{chen2019quantile,wang2022renewable,jiang2022renewable,shen2024online} assumed that the covariates \( x \) are at least sub--Gaussian and that the conditional distribution of the error \( \epsilon \) given \( x \) has a density that is continuous everywhere. In this work, by studying the class of sub--quadratic SGD, we introduce a unified framework for a more complete analysis of both online algorithms, and relax the assumptions imposed in previous work.

In Table~\ref{tab:summary}, we summarize some most related work on SGD. One line of work considers strongly convex objective functions $f$, where the behavior of the iterates $\{\theta_n\}_{n \geq 0}$ is well-understood under both diminishing stepsizes ($\alpha_n \to 0$) and constant stepsizes ($\alpha_n \equiv \alpha$). With diminishing stepsizes, studies  have demonstrated that setting $\alpha_n = \iota/(n + \kappa)$ achieves the optimal convergence rate of $\mathbb{E}[\|\theta_n - \theta^*\|^2] = \mathcal{O}(1/n)$ \cite{bottou2018optimization, chen2020finite}. Under a constant stepsize $\alpha$, the iterates $\{\theta_n\}_{n \geq 0}$  form a Markov chain that converges geometrically to a limiting random variable $\theta_\infty^{(\alpha)}$, which oscillates around $\theta^*$ \cite{dieuleveut2018bridging, zhang2024prelimit}. Notably, the asymptotic bias $\mathbb{E}[\theta_\infty^{(\alpha)}] - \theta^*$ is typically nonzero. When $\nabla f$ is continuously differentiable at $\theta^*$, this bias is proportional to the stepsize $\alpha$ \cite{dieuleveut2018bridging, huo2023bias, zhang2024constant}; when $\nabla f$ lacks continuous differentiability the bias is proportional to $\sqrt{\alpha}$ \cite{zhang2024prelimit}.

In contrast, our understanding of SGD for minimizing sub--quadratic functions $f$ is very limited. Prior work  \cite{kurdyka1998gradients,josz2023global} has primarily focused on deterministic gradient descent ($w(\cdot) \equiv 0$), providing convergence rates for $\|\theta_n - \theta^*\|$  with diminishing stepsizes. Recent work \cite{gadat2023optimal} considered sub--quadratic SGD with diminishing stepsizes and derived non-asymptotic moment bounds for $\mathbb{E}[\|\theta_n - \theta^*\|^p]$. However, their analysis requires twice differentiablility of the function $f$ and restrictive assumption on the noise sequence $\{w_n(\cdot)\}_{n \geq 0}$, due to the limitation of their proposed Lyapunov function. Beyond diminishing stepsizes, the study of constant stepsize sub--quadratic SGD via a Markov chain perspective remains scarce. In recent developments, \cite{qu2023computable} introduced a novel technique to analyze the weak convergence of Markov chains, validating their approach with specific examples of sub--quadratic SGD involving solely additive (i.e., $w(\theta)$ is independent of $\theta$) and heavy--tailed noise.

\begin{table}[h]
\centering
\begin{tabular}{|c|c|c|c|c|}
\toprule
Objective function $f$ & Noise $w(\theta)$& \makecell{Stepsize condition
\\ for moment bounds}& \makecell{Weak convergence\\with constant stepsize \\
$\theta_n \Rightarrow \theta_\infty^{(\alpha)}$} &Bias \\ \midrule
strongly convex~\cite{chen2020finite} & $\E[\|w(\theta)\|^2] <\infty$ & \makecell{$\alpha_n = \frac{\iota}{(n+\kappa)^\xi}$,\\ $\xi \in [0,1]$} &  \textemdash &  \textemdash \\ \hline
strongly convex~\cite{dieuleveut2018bridging} & $\E[\|w(\theta)\|^2] <\infty$ & $\alpha_n \equiv \alpha$ &  geometric &  $\Theta(\alpha)$ \\ \hline
\makecell{subquadratic, \\$f \in \mathcal{C}^2(\R^d)$~\cite{gadat2023optimal}} & sub--Gaussian& \makecell{$\alpha_n= \frac{\iota}{(n+\kappa)^\xi},$\\
$ \xi \in [1/2,1)$} &  \textemdash &  \textemdash \\ \hline \makecell{some special\\ subquadratic $f$~\cite{qu2023computable}}
 &  $\E[\|w(\theta)\|^2] =\infty$& \textemdash &  polynomial &  \textemdash \\ \hline
\makecell{subquadratic, \\$f \in \mathcal{C}^1(\R^d)$,\\ \textbf{This work}}  &  sub--exponential & \makecell{$\alpha_n = \frac{\iota}{(n+\kappa)^\xi}, $\\$\xi \in [0,1]$} &  geometric &  $\Theta(\alpha)$ \\ \bottomrule
\end{tabular}
\caption{Summary of most related work on SGD under different settings. We examine conditions on
the objective function $f$ and the noise $w(\theta)$, stepsize condition required for establishing moment bounds $\mathbb{E}[\|\theta_n - \theta^*\|^{2p}]$, and discuss
weak convergence result and bias characterization. Here, $\mathcal{C}^m(\mathbb{R}^d)$ denotes the class of real-valued functions on $\mathbb{R}^d$ that are $m$-times continuously differentiable.}
\label{tab:summary}
\end{table}

In this paper, we investigate sub--quadratic SGD, focusing on scenarios where the objective function \( f \) is only first-order differentiable and exhibits tails with at least linear growth. This function class covers many widely adopted loss functions in various domains.  %and provides more comprehensive analyses compared to previous studies \cite{yu2021analysis, gadat2023optimal, qu2023computable}. 
%Leveraging our analyses on sub--quadratic SGD, we relax restrictive assumptions considered in prior work on online robust regression and online quantile regression, and provide a more thorough analysis for both scenarios. 
Building upon this class of SGD, our main contributions are summarized as follows.%\paragraph{Main Contributions:}
\begin{itemize}[leftmargin=0.2in]
\item \textbf{A Novel Piecewise Lyapunov Function:} As elaborated in Section \ref{sec:challenge}, the key challenge of analyzing  the convergence of the iterates $\{\theta_n\}_{n \geq 0}$ is to construct an appropriate Lyapunov function. This task is complicated by the piecewise behavior of the objective function, which is locally strongly convex around $\theta^*$ but has a sub--quadratic tail. Prior work \cite{gadat2023optimal} develops a Lyapunov function for sub--quadratic SGD but imposes restrictive conditions on both the objective function and noise. In Section \ref{sec:Lyapunov}, we propose a novel piecewise Lyapunov function that effectively exploits the structure of sub--quadratic SGD without such restrictive assumptions. 
%The specifics of our Lyapunov function are detailed in Section \ref{sec:Lyapunov}. 
This new Lyapunov function allows us to derive the following analytical results.

\item \textbf{Finite-Time Moment Bounds:} We provide finite-time analysis of the moments \(\mathbb{E}[\|\theta_n - \theta^*\|^{2p}]\) for both diminishing and constant stepsizes. While \cite{gadat2023optimal} also obtain moment bounds, they only considered diminishing stepsizes $\alpha_n = \frac{\iota}{(n+\kappa)^\xi}$ with $\xi \in [1/2,1)$.  %and required the objective function \( f \) to be twice differentiable, along with restrictive assumptions on the noise sequence \(\{w_n(\cdot)\}_{n \geq 0}\). 
In contrast, with our new piecewise Lyapunov function, we are able to analyze a broader class of diminishing stepsizes with $\xi \in (0,1]$. In particular, we recover the results in \cite{gadat2023optimal}, without requiring the objective function to be twice differential or imposing restrictive assumptions on the noise sequence.
    %with  weaker assumptions on \( f \) and \(\{w_n(\cdot)\}_{n \geq 0}\). 
Our techniques also enables us to establish moment bounds under constant stepsizes (\(\alpha_n \equiv \alpha\)), and such bounds  play a crucial role in the fine-grained characterization of the Markov chain $\{\theta_n\}$. These results are presented in Section~\ref{sec:moment}.

    \item \textbf{Weak Convergence, Central Limit Theorem and Bias Characterization:} We investigate the SGD update \eqref{eq:SGD} with a constant stepsize, and  establish the weak convergence of the iterates \(\{\theta_n\}_{n \geq 0}\) to a limiting random variable \(\theta_\infty^{(\alpha)}\). % and provide an explicit geometric convergence rate. 
    In particular, leveraging the new piecewise Lyapunov function, we extend the drift and contraction (D\&C) technique \cite{qin2022geometric} to show that constant stepsize sub--quadratic SGD with sub--exponential noise  converges geometrically to \(\theta_\infty^{(\alpha)}\). To the best of our  knowledge, this is the first geometric convergence result for SGD applied to sub--quadratic functions. Having established weak convergence, we further prove a central limit theorem for the Markov chain \(\{\theta_n\}_{n \geq 0}\), which is crucial for statistical inference~\cite{li2023statistical}. Additionally, we characterize the asymptotic bias \(\mathbb{E}[\theta^{(\alpha)}_\infty] - \theta^*\) and show that it is proportional to the stepsize \(\alpha\) up to higher-order terms. These results are detailed in Section~\ref{sec:weakconvergence}.

    \item \textbf{Applications in  Robust Regression and Quantile Regression:} We apply our results on general sub--quadratic SGD to important statistical problems: online robust regression \cite{pesme2020online,godichon2024online} and online quantile regression \cite{chen2019quantile,wang2022renewable,jiang2022renewable,shen2024online}. For both settings, our new Lyapunov analysis allows us to relax restrictive assumptions considered in prior work and thereby provide a more comprehensive analysis. Specifically, our results are applicable to online robust regression with sub--exponential covariates and heavy--tailed error, and similarly to online quantile regression with sub--exponential covariates without requiring the continuity of the conditional density of the error \( \epsilon \). 
    %Compared to earlier work, we provide a more comprehensive analysis for both problems. 
    These findings are presented in Sections~\ref{sec:robustregression} and~\ref{sec:quantile}.

\end{itemize}

\subsection{Additional Related Work}\label{sec:relate}

\paragraph{Stochastic Gradient Descent and Stochastic Approximation}
The study of stochastic gradient descent (SGD) and stochastic approximation (SA) began with the seminal work of Robbins and Monro~\cite{robbins1951stochastic}. Early research focused on diminishing stepsizes~\cite{blum1954approximation, blum1954multidimensional}, proving almost sure asymptotic convergence for contractive SA and strongly convex SGD algorithms. Later, Ruppert~\cite{ruppert1988efficient} and Polyak~\cite{polyak1992acceleration} introduced the Polyak-Ruppert averaging technique to accelerate convergence. Recent works have explored non-asymptotic convergence with diminishing stepsizes. Studies in \cite{chen2022finite, chandak2022concentration, chen2024lyapunov} focus on contractive SA, establishing moment bounds on $\mathbb{E}[\|\theta_n - \theta^*\|^{2p}] $, and \cite{chen2023concentration} provides non-asymptotic confidence bounds on the estimation error. However, these works are limited when considering subquadratic SGD. Gadat et al.~\cite{gadat2023optimal} address diminishing stepsize subquadratic SGD, providing moment bounds for raw and averaged iterates under the assumption of twice differentiability of the objective function $f$. In contrast, we analyze subquadratic SGD requiring only once differentiability of $f$ by proposing a novel piecewise Lyapunov function.

There is growing interest in studying SGD and SA with constant stepsizes due to their ease of implementation and potential for faster convergence and robustness~\cite{dieuleveut2018bridging}. However, constant stepsizes eliminate the almost sure convergence guarantee present with diminishing stepsizes. Instead, convergence is to a limiting random variable $\theta^{(\alpha)}_\infty$~\cite{dieuleveut2018bridging, yu2021analysis, huo2023bias, zhang2024constant}, often exhibiting asymptotic bias where $\mathbb{E}[\theta^{(\alpha)}_\infty] \neq \theta^*$. For contractive and locally smooth SA updates, studies have shown that the asymptotic bias is of order $\Theta(\alpha)$~\cite{dieuleveut2018bridging, huo2023bias, zhang2024constant}. Zhang~\cite{zhang2024prelimit} investigates contractive but non-differentiable SA updates with specific structures around $\theta^*$, proving that the asymptotic bias is of order $\Theta(\sqrt{\alpha})$. In the constant stepsize regime, several works provide non-asymptotic moment bounds; for example, \cite{lakshminarayanan2018linear, mou2020linear} focus on linear SA, and \cite{chen2024lyapunov} uses the generalized Moreau envelope to analyze general contractive SA. However, limited prior work addresses constant stepsize subquadratic SGD. In this paper, we establish similar results for constant stepsize subquadratic SGD as those for constant stepsize strongly convex SGD and contractive SA.

\paragraph{Markov Chain Studies}
When considering SGD with a constant stepsize, the raw iterates \(\{\theta_n\}_{n \geq 0}\) form a time-homogeneous Markov chain in a general state space \cite{dieuleveut2018bridging, zhang2024prelimit}. Previous work focusing on Markov chains in general state spaces has proposed various techniques to verify convergence and provide convergence rates. Most convergence results are established by verifying drift and minorization (D\&M) conditions \cite{baxendale2005renewal, andrieu2015quantitative, yu2021analysis}. However, verifying D\&M conditions often relies on assuming that the density of the noise is lower bounded from 0 in a selected region, which may not hold when the noise term follows certain discrete distributions. Recently, \cite{qu2023computable} proposed a contractive drift (CD) condition, under which convergence with polynomial, subgeometric, and geometric rates can be verified. However, their analysis heavily depends on the smoothness of the update (see \cite[Assumption 2]{qu2023computable}) and it is not clear how to analyze their locally Lipschitz constant when the noise $w(\cdot)$ is not an additive nouse. In this work, we utilize an alternative framework proposed by \cite{qin2022geometric} that verifies drift and contraction (D\&C) conditions, which can be used to provide a geometric convergence rate of a Markov chain in the Wasserstein-1 distance.

\paragraph{Online Robust Regression and Online Quantile Regression.} Robust regression \cite{bhatia2017consistent, loh2017statistical, loh2021scale} and quantile regression \cite{koenker2005quantile, hao2007quantile} have a long-standing history in statistics. Notably, most previous methods are based on a batch framework, where the entire dataset is available before estimation begins. Recently, attention has shifted towards online methods. In online robust regression, previous work \cite{pesme2020online} studied the $\tilde{\eta}$-corrupted linear model with Gaussian covariates and noise, providing a convergence rate of $\mathcal{O}\left(\frac{d}{n(1-\tilde{\eta})^2}\right)$. For online quantile regression, \cite{shen2024online} considered the SGD update but restricted to Gaussian covariates, establishing a convergence rate of $\mathcal{O}(1/n)$. In this work, by utilizing results on sub--quadratic SGD, we provide more fine-grained analyses in both settings.

\section{Preliminaries}\label{sec:preliminary}
In this work, we study the SGD of sub--quadratic objective functions $f$ satisfying the following assumptions.
\begin{assumption}[Smoothness]\label{assumption:smooth}
There exists a constant \( L > 0 \) such that, for all \( \theta, \theta' \in \mathbb{R}^d \), 
\[
\|\nabla f(\theta) - \nabla f(\theta')\| \leq L \|\theta - \theta'\|.
\]
\end{assumption}

\begin{assumption}[Local Strong Convexity and sub--quadratic Tail]\label{assumption:hubor}
There exist $\theta^*\in \R^d$ and constants \( \mu, a, b, \Delta > 0 \) and \( k \in [1, 2) \) such that: %\qxcomment{$\exists \theta^*\in \R^d$?}
\begin{enumerate}
    \item \(\forall \theta', \theta'' \in \{ \theta \in \mathbb{R}^d : \|\theta - \theta^*\| \leq \Delta \} \),
    \[
    \langle \theta' - \theta'', \nabla f(\theta') - \nabla f(\theta'') \rangle \geq \mu \|\theta' - \theta''\|^2.
    \]
    \item \(\forall \theta \in \mathbb{R}^d \in \{\theta \in \R^d: \|\theta - \theta^*\| > \Delta \}\),
    \begin{align*}
        \|\nabla f(\theta)\| \leq a \|\theta - \theta^*\|^{k - 1} \quad \text{ and }  \quad \langle \theta - \theta^*, \nabla f(\theta) \rangle &\geq b \|\theta - \theta^*\|^k.
    \end{align*}
\end{enumerate}
\end{assumption}
A few remarks are in order. In Assumption~\ref{assumption:hubor}, the first condition requires that the objective function \( f \) is locally strongly convex in a neighborhood of \( \theta^* \). The second condition states that \( f \) grows at least as fast as \( \|\theta - \theta^*\|^{k} \). As an immediate consequence, Assumption~\ref{assumption:hubor} ensures that \( f \) has a unique minimum at \( \theta^* \). Moreover, many popular loss functions in robust statistics satisfy Assumptions~\ref{assumption:smooth} and~\ref{assumption:hubor}. For example, the Huber loss $l(t) = \begin{cases}
\frac{1}{2} t^2, & \text{if } |t| \leq \delta, \\
\delta |t| - \frac{1}{2} \delta^2, & \text{otherwise},
\end{cases}$ and the Pseudo-Huber loss
$l(t) = \delta^2 ( \sqrt{1 + ( t/\delta )^2 } - 1 )$~\cite{hartley2003multiple} satisfy the above two assumptions with \( k = 1 \). Additionally, the generalized Charbonnier loss $l(t) = \left( t^2 + c^2 \right)^{\alpha / 2}$~\cite{sun2010secrets} and Barron's general robust loss $l(t) = \frac{|\alpha - 2|}{\alpha} \Big( \big( \frac{(t / c)^2}{|\alpha - 2|} + 1 \big)^{\alpha / 2} - 1 \Big)$~\cite{barron2019general} satisfy the above two assumptions with \( k = \alpha \) when \( \alpha \in [1, 2) \). Furthermore, as we will see in Sections \ref{sec:robustregression} and \ref{sec:quantile}, online robust regression and online quantile regression can be reformulated as SGD for sub--quadratic functions with \( k = 1 \).

In this work, we focus on the parameter regime \( k \in [1, 2) \). We intentionally exclude \( k = 2 \) since the case \( k = 2 \) (i.e., one point convexity)  has been studied extensively \cite{dieuleveut2018bridging, yu2021analysis}, while analysis for \( k \in [1, 2) \) remains relatively scarce. Compared with the previous assumptions on the sub--quadratic objective function \cite{gadat2023optimal}, we relax the assumption that the objective function is twice differentiable by only requiring the first-order differentiability. Although we need an additional assumption that $\langle \theta - \theta^*, \nabla f(\theta) \rangle \geq b \|\theta - \theta^*\|^k$, we argue that this is verifiable in both online robust regression and quantile regression. Furthermore, our Assumption \ref{assumption:hubor} leads to a better designed Lyapunov function, which requires fewer additional assumptions about the objective function and the noise, as detailed in Section \ref{sec:Lyapunov}. Importantly, our Assumption \ref{assumption:hubor} allows us to include more commonly used objective functions that are only once differentiable, such as the widely used Huber loss.

To state the assumption on the noise \(\{w_n(\cdot)\}_{n \geq 0}\), we introduce the \(\psi_q\)-Orlicz space \cite{vershynin2018high}.

\begin{definition}[$\psi_q-$Orlicz Space]\label{def:orlicz}
Let $X$ be a real random variable in the \(\psi_q\)-Orlicz space, denoted by \(L_{\psi_q}\). Then, the following properties are equivalent; the parameters $K_{q,i} > 0, i\in [2]$ appearing in these
properties differ from each other by at most an absolute constant factor.
\begin{enumerate}
    \item $\mathbb{P}(|X| \geq t) \leq 2 \exp (-t^q / K_{q,0}^q), \/ \forall t \geq 0$.
    \item$(\mathbb{E}|X|^p)^{1 / p} \leq K_{q,1} p^{1/q}, \/  \forall p \geq 1$.
    \item $\mathbb{E} \exp (\lambda^q |X|^q) \leq \exp (K_{q,2}^q \lambda^q), \/ 
 \forall \lambda \text { such that } 0 \leq \lambda \leq \frac{1}{K_{q,2}} \text {. }$
\end{enumerate}
\end{definition}
In this paper, we consider i.i.d.\ noise sequence $\{w_n(\cdot)\}_{n \geq 0}$ that is uniformly in the $\psi_q-$Orlicz space for some $q \in (0,1]$, as stated in the following assumption.
\begin{assumption}[q]\label{assumption:uniformlyorlicz}
$\|w(\theta)\| $ is in $L_{\psi_q}$ with the  parameters $K_{q,i} > 0, i\in [2]$ for all $\theta \in \R^d.$
\end{assumption}

We note that \( L_{\psi_1} \) represents the class of all sub--exponential random variables, and \( L_{\psi_2} \) denotes the class of all sub--Gaussian random variables, and \( L_{\psi_{q_1}} \subset L_{\psi_{q_2}} \) whenever \( q_1 > q_2 \). 
In this work, we focus on objective functions that satisfy Assumption~\ref{assumption:hubor} with a parameter \( k \in [1, 2) \). Consequently, we require the noise sequence to fulfill Assumption~\ref{assumption:uniformlyorlicz}(q) with \( q = 2 - k \) by default. Throughout the remainder of the paper, when we refer to Assumption~\ref{assumption:uniformlyorlicz}, it specifically denotes the case where \( q = 2 - k \).  We say that a random variable $x$ is $\sigma-$sub--exponential if $x \in L_{\psi_1}$ with $K_{1,0} = \sigma.$

Furthermore, it is crucial to highlight that for SGD with strongly convex objective functions, previous studies \cite{merad2023convergence, zhang2024prelimit} only require that the expected squared norm of the noise satisfies \(\mathbb{E}[\|w(\theta)\|^2] \in \mathcal{O}(\|\theta - \theta^*\|^2 + 1)\) for all \(\theta \in \mathbb{R}^d\). In contrast, as we will expain in Section \ref{sec:Lyapunov}, achieving similar results as those for strongly convex SGD in the context of sub--quadratic SGD necessitates more stringent conditions beyond merely having finite moment bounds and Assumption \ref{assumption:uniformlyorlicz} is exactly the least assumption we need. Additionally, we remark that \cite{gadat2023optimal} imposes a more restrictive assumption on the noise sequence by requiring that \( \{w_n(\cdot)\}_{n \geq 0} \) uniformly resides in the \( L_{\psi_{4-2k}} \) space. In contrast, our approach only requires \( q = 2 - k < 4-2k\), thereby relaxing the assumption on the noise sequence compared to \cite{gadat2023optimal}. Moreover, as to be discussed in Sections \ref{sec:robustregression} and \ref{sec:quantile}, the noise for online robust regression and online quantile regression satisfies Assumption \ref{assumption:uniformlyorlicz}.

\subsection{Notations}
Let \( I_d \) represent the \( d \times d \) identity matrix, and let \( \mathbb{N} \) denote the set of natural numbers. The symbols \( \Sigma \) and \( \prod \) are used to indicate summation and product operations, respectively. When the lower index exceeds the upper index, i.e., \( a > b \), we define \( \sum_{i=a}^b = 0 \) and \( \prod_{i=a}^b = 1 \). For any \( n \in \mathbb{N} \), the notation \( [n] \) stands for the set \( \{0, 1, \ldots, n\} \). If \( x \in \mathbb{R}^d \) is a vector, \( \|x\| \) denotes its Euclidean norm, and if \( A \in \mathbb{R}^{d \times d} \) is a matrix, \( \|A\| \) represents its operator norm. A function \( g: \mathbb{R}^d \to \mathbb{R} \) is called 1-Lipschitz if for any \( \theta, \theta' \in \mathbb{R}^d \), \( |g(\theta) - g(\theta')| \leq \|\theta - \theta'\| \). We use $\otimes$ to denote the tensor product.

We denote \( \mathcal{P}_1(\mathbb{R}^d) \) as the space of integrable probability measures on \( \mathbb{R}^d \). For a random vector \( \theta \in \R^d\), \( \mathcal{L}(\theta) \) represents its distribution. The Wasserstein 1-distance between two distributions \( \mu \) and \( \nu \) in \( \mathcal{P}_1(\mathbb{R}^d) \) is defined as
\begin{align}\label{eq:wasserstein}
W_1(\mu, \nu) = \inf_{\xi \in \Pi(\mu, \nu)} \int_{\mathbb{R}^d} \|u - v\| \, d\xi(u, v) = \inf \left\{ \mathbb{E}\left[ \|\theta - \theta'\| \right] : \mathcal{L}(\theta) = \mu, \mathcal{L}(\theta') = \nu \right\},    
\end{align}
where \( \Pi(\mu, \nu) \) denotes the set of all joint distributions in \( \mathcal{P}(\mathbb{R}^d \times \mathbb{R}^d) \) with marginals \( \mu \) and \( \nu \).

For real-valued functions \( g_1(x), g_2(x) : \mathbb{R}^+ \to \mathbb{R}^+ \), we write \( g_1(x) \in o(g_2(x)) \) if \( \lim_{x \to \infty} \frac{g_1(x)}{g_2(x)} = 0 \). We denote \( g_1(x) \in \mathcal{O}(g_2(x)) \) if there exists a constant \( C > 0 \) such that \( g_1(x) \leq Cg_2(x) \) for all sufficiently large \( x \), and \( g_1(x) \in \Omega(g_2(x)) \) if \( g_1(x) \geq Cg_2(x) \). Finally, we have \( g_1(x) \in \Theta(g_2(x)) \) if \( g_1(x) \in \mathcal{O}(g_2(x)) \) and \( g_1(x) \in \Omega(g_2(x)) \).

\section{Challenges of Analyzing sub--quadratic SGD and A New Piecewise Lyapunov Function}\label{sec:Lyapunov}
In this section, we discuss the challenges associated with analyzing sub--quadratic SGD and provide an intuition behind our proposed Lyapunov function. For illustration purpose, we consider the following simple SGD algorithm with a constant stepsize $\alpha$ and only additive noise, as presented in \cite[Section 8.2]{qu2023computable}:
\begin{align}\label{eq:simpleSGD}
\theta_{n+1} = \theta_n - \alpha(f'(\theta_n) + w_n) =
\begin{cases}
\theta_n - \alpha(\theta_n + w_n) & \text{if } |\theta_n| < 1, \\
\theta_n - \alpha(\operatorname{sign}(\theta_n)|\theta_n|^{\beta-1} + w_n) & \text{if } |\theta_n| \geq 1,
\end{cases}
\end{align}
where $\beta \in [1,2)$ and $\{w_n\}_{n \geq 0}$ denotes the i.i.d.\ zero mean additive noise sequence independent of $\theta$. The corresponding objective function is defined as
\begin{align}\label{eq:simpleobjective}
f(\theta) = 
\begin{cases}
 \theta^2/2 & \text{if } |\theta| < 1, \\
|\theta|^\beta/\beta - 1/\beta + 1/2 & \text{if } |\theta| \geq 1.
\end{cases}
\end{align}
It is easy to verify that the objective function \eqref{eq:simpleobjective} satisfies Assumptions \ref{assumption:smooth} and \ref{assumption:hubor}.
\subsection{Limitations of Prior Work and Challenges of Analyzing sub--quadratic SGD}\label{sec:challenge}

%\lhcomment{Global strongly convex, well understand; local + tail sub--quadratic -- not well-understood. Tail sub--quadratic, robustness regression, bounded gradient, inherent feature. Intuitive, geometric rate, verified for GD (deterministic). But with noise/fluctuation, not sure. In fact, if too heavy tailed, verified by QYL's paper, problematic, no geometric convergence. If not heavy--tailed (inherent feature), then conjecture geometric, but not clear.}

For the simple example in \eqref{eq:simpleobjective}, when $\beta = 2$, the objective function is global strongly convex. The corresponding SGD dynamic \eqref{eq:simpleSGD} is well studied~\cite{dieuleveut2018bridging}, where the iterates $\{\theta_n\}$ converge geometrically to a limiting random variable $\theta^{(\alpha)}_\infty$, assuming the noise has a finite second moment ($\mathbb{E}[|w_0|^2] < \infty$). Meanwhile, for functions with local strong convexity and sub--quadratic tail ($\beta\in [1,2)$), our understanding of the SGD convergence is much limited. For the special case of deterministic gradient descent without noise ($w_n \equiv 0$), one can show that the iterates converge geometrically to $\theta^*=0$ due to the local strong convexity of $f$, achieving Q-convergence of order 1\footnote{A sequence $\{\theta_n\}$ converges to $L$ with Q-convergence of order 1 if $0 < \lim_{n\to\infty} \frac{|\theta_{n+1} - L|}{|\theta_n - L|} < 1$.}.

However, challenges arise with the presence of gradient noise $\{w_n\}$. It remains unclear whether the SGD update~\eqref{eq:simpleSGD} can still achieve geometric weak convergence as the deterministic case or strongly convex setting. Interestingly, recent work~\cite{qu2023computable} showed that when the noise is heavy--tailed with $\mathbb{E}[|w_0|^\gamma] < \infty$ for $\gamma \in (1,2]$ and $\gamma + \beta \geq 3$, the iterates $\{\theta_n\}$ converge weakly to a stationary distribution at a \textbf{polynomial} rate of $\mathcal{O}\big(n^{-\frac{\gamma + \beta - 3}{2 - \beta}}\big)$. For the special case $\beta=1$ and $\gamma=2,$ they establish an nonexplicit convergence rate $o(1)$, which is argued to be non-improvable to any polynomial rate. The distinct behaviors of this class of SGD demonstrate the compounding effect of the noise and the sub--quadratic tail on the convergence rate. 

On the other hand, in many applications with sub--quadratic objective functions, such as robust regression and quantile regression (cf.~Sections~\ref{sec:robustregression}--\ref{sec:quantile}), the gradient noise is inherently light-tailed, such as sub--exponential distribution. Intuitively, when the noise $w_n$ is light-tailed with higher-order moments, the iterate $\theta_n$ tends to move closer towards the local strongly convex region compared to the heavy--tailed case. This raises the question of whether the iterates $\{\theta_n\}$ can converge to a stationary distribution at a faster, possibly \textbf{geometric}, rate when the noise is light-tailed. We remark that the method in~\cite{qu2023computable} is limited to the heavy--tailed noise with $\gamma \in (1,2]$ and it is unclear how to extend their approach to the light-tailed case.

To analyze the dynamic of a stochastic sequence $\{\theta_n\}$, a common approach is to investigate the drift of an appropriately chosen Lyapunov function to bound the moments of the error $|\theta_n-\theta^*|$~\cite{srikant2019td_learning,huo2023bias,chen2024lyapunov,chen2022finite}. The key challenge here is the construction of a proper Lyapunov function. Specifically, to establish geometric convergence of the iterates $\{\theta_n\}$, the Lyapunov function $V:\R^d\rightarrow \R^+$ is expected to satisfy the following drift condition:
\begin{align}\label{eq:drift}
\E[V(\theta_1)] \leq (1 - \alpha \nu) V(\theta_0) + \mathcal{O}(\alpha^2), \quad \forall \theta_0 \in \mathbb{R}.
\end{align}
Lyapunov functions with the above property are crucial for deriving geometric moment bounds of strongly convex case~\cite{chen2020finite, chen2022finite}, as well as for establishing geometric weak convergence of Markov chain by verifying the drift and contraction (D\&C) condition~\cite{qin2022geometric}. While there are other techniques developed for proving geometric convergence, they have limitations as discussed in Section~\ref{sec:relate}.

To gain intuition on identifying Lyapunov functions of property \eqref{eq:drift} for sub--quadratic SGD, let us consider the simple example \eqref{eq:simpleSGD}. By taking Taylor expansion of $V(\theta_1)$, we note that there exists a random variable $\lambda \geq 0$ depending on $\theta_0$ and $w_0$ such that 
\begin{align}
\E[V(\theta_1)] &= V(\theta_0) -  \alpha \E[V'(\theta_0) (f'(\theta_0) + w_0)] + \alpha^2/2 \E[V''(\theta_0 - \alpha\lambda(f'(\theta_0) + w_0))(f'(\theta_0) + w_0)^2]\nonumber\\
&= V(\theta_0) -  \alpha V'(\theta_0) f'(\theta_0)  +\frac{\alpha^2}{2} \E[V''(\theta_0 - \alpha\lambda(f'(\theta_0) + w_0))(f'(\theta_0) + w_0)^2]. \label{eq:taylor}
\end{align}
Combining equation \eqref{eq:taylor} and our goal \eqref{eq:drift}, we aim to find a Lyapunov function that satisfy
\begin{enumerate}
    \item $V'f' \in \Omega (V);$
    \item $\E[V''(\lambda \theta_1 + (1-\lambda)\theta_0)(f'(\theta_0) + w(\theta_0))^2] \in \mathcal{O}(V(\theta_0)+1);$
    \item Require minimal additional assumptions on the objective function $f$ and the noise $w$.
\end{enumerate}
Given condition (1), we argue that the classical polynomial Lyapunov function $V(\theta) = |\theta|^{2p}$ for strongly convex case can not be applied to sub--quadratic SGD, since $V'(\theta)f'(\theta) = 2p|\theta|^{2p+\beta-2} \notin \Omega(V(\theta)) $ when $|\theta|>1$. In fact, to ensure condition (1) holds, we need to analyze (1) in two regions. 

\paragraph{Region 1 (local strong convexity): \( |\theta| < 1 \).} In this case, (1) simplifies to: $V'(\theta) \theta \propto V(\theta),$ which implies \( V(\theta) \propto \theta^{2p} \) by solving the ordinary differential equation (ODE) \( y' x = 2 p y \) for any \( p \in \mathbb{N}^+ \).
\paragraph{Region 2 (sub--quadratic tail): \( |\theta| \geq 1 \). }Here, (1) becomes: $V'(\theta) \operatorname{sign}(\theta) |\theta|^{\beta -1} \propto V(\theta),$ leading to \( V(\theta) \propto \exp\left( \frac{\theta^{2 - \beta}}{2 - \beta} \right) \) by solving the ODE \( y' \operatorname{sign}(x) |x|^{\beta -1} = y \).

Importantly, we observe that for sub--quadratic SGD, the Lyapunov function has to admit an exponential tail with the order of at least $\exp(|\theta|^{2-\beta})$ in Region 2. Consequently, to ensure $\E[V(\theta_1
)] = \E[V(\theta_0 - \alpha (\nabla f(\theta_0) + w_0))]$ well-defined, the noise should at least satisfy $\E[\exp(|w_0|^{2-\beta})] < \infty.$

There are two approaches to define a Lyapunov function that satisfies the above properties in the two distinct regions. One option is to construct a unified Lyapunov function, which offers great convenience for analysis. 
However, such a function may fail to capture key behaviors of the SGD across different regions and often require additional assumptions. For instance,
recent work \cite{gadat2023optimal} considered a similar class of sub--quadratic functions $f$ and introduced a unified Lyapunov function of the form \( V(\theta) = f(\theta) \exp(\phi(f(\theta))) \). Since their Lyapunov function depends on $f$, they require the twice differentiability of $f$ to perform the Taylor expansion \eqref{eq:taylor}. Moreover, they need to assume $\E[\exp(|w_0|^{4-2\beta})] < \infty$ to ensure condition (2) holds. The alternative approach is to define a piecewise Lyapunov function, which offers more flexibility in capturing the specific behaviors of SGD in each region. The key challenge with this method lies in the analysis of condition (2), as it becomes difficult to determine which region the random variable $\theta_0 - \alpha\lambda(f'(\theta_0) + w_0)$ falls into.  In this work, we \textbf{piecewisely} define the Lyapunov function, and address the challenge by carefully analyzing the impact of the noise on the drift of the Lyapunov function; see Section \ref{sec:pivot} for details.

\subsection{A New Piecewise Lyapunov Function}\label{sec:newLyapunov}

The discussions in Section \ref{sec:challenge} provide insights into constructing an appropriate piecewise Lyapunov function for a general sub--quadratic SGD \eqref{eq:SGD}. Building on the discussion and under Assumption  \ref{assumption:hubor}, the suitable piecewise Lyapunov function \( V \) should be formulated as:
\begin{align*}
V(\theta) = 
\begin{cases}
\exp\left( r_1 \|\theta - \theta^*\|^{2 - k} \right) - r_2, & \text{if } \|\theta - \theta^*\| > \Delta, \\
r_3 \|\theta - \theta^*\|^2, & \text{if } \|\theta - \theta^*\| \leq \Delta.
\end{cases}
\end{align*}
To ensure that the Lyapunov function \( V \) is twice differentiable everywhere—an essential requirement for analyzing the remaining terms in the Taylor expansion \eqref{eq:taylor}—we carefully select the constants \( r_1 \), \( r_2 \), and \( r_3 \) so that \( V \) is continuous and has continuous first and second derivatives. These constants are uniquely determined, leading to the following Lyapunov functions.

For all \( k \in [1,2) \) consistent with Assumption \eqref{assumption:hubor}, we define:
\begin{align}\label{eq:V}
V_{k,0}(\theta) = 
\begin{cases}
\exp\big( \frac{k \|\theta - \theta^*\|^{2 - k}}{(2 - k) \Delta^{2 - k}} \big) - (1 - k/2) \exp( \frac{k}{2 - k} ), & \text{if } \|\theta - \theta^*\| > \Delta, \\
\frac{k \exp( k/(2 - k)) \|\theta - \theta^*\|^2}{2 \Delta^2}, & \text{if } \|\theta - \theta^*\| \leq \Delta.
\end{cases}
\end{align}
Subsequently, we define:
\begin{align}\label{eq:Vp}
V_{k,p}(\theta) = \|\theta - \theta^*\|^p \cdot V_{k,0}(\theta), \quad \forall p \geq 0,
\end{align}
which serves as a Lyapunov function for analyzing the higher moment bounds of the SGD iterates.

\subsection{Pivot Results}\label{sec:pivot}
In this subsection, we explore key properties of our proposed Lyapunov functions \( V_{k,p} \). We present two essential pivot results that are crucial in establishing our main findings in Section \ref{sec:mainresults}.

We summarize important properties of the Lyapunov function \( V_{k,p} \) in the following lemma.

\begin{lemma}\label{lem:V}
Given $V_{k,p}$ defined in equation \eqref{eq:Vp}, we have
\begin{enumerate}
\item $V_{k,p}(\theta) \geq \frac{k\exp(k/(2-k))\|\theta-\theta^*\|^{2+p}}{2\Delta^2}, \quad \forall \theta \in \R^d.$
    \item $V_{k,p}(\cdot)$ is twice differentiable everywhere for all $k \in [1,2)$ and $p \geq 0$.
    \item There exist some constants $c_{k},c_{k}'\geq 0$ that depend only on $k$ and $\Delta$ such that for all $\theta\in \R^d,$
\begin{align*}
\|V_{k,p}''(\theta)\| \leq& c_{k}(1 + p)^2\|\theta-\theta^*\|^{p}\exp\bigg(\frac{k\|\theta-\theta^*\|^{2-k}}{(2-k)\Delta^{2-k}}\bigg),\\
\|V_{k,p}''(\theta)\| \leq& c_{k}'(1 + p)^2\|\theta-\theta^*\|^{p+2-2k}\exp\bigg(\frac{k\|\theta-\theta^*\|^{2-k}}{(2-k)\Delta^{2-k}}\bigg).
\end{align*}
    \item Under Assumption \ref{assumption:hubor} with $k \in [1,2)$, for all $\theta \in \R^d$, we have
    \begin{align*}
    \langle V_{k,p}'(\theta), \nabla f(\theta)\rangle \geq \min\Big\{\frac{b k}{\Delta^{2-k}}, \mu(2+p)\Big\}V_{k,p}(\theta).
    \end{align*}
\end{enumerate}
\end{lemma}
Several important observations are worth mentioning. First, Lemma~\ref{lem:V}--(2) and Lemma~\ref{lem:V}--(4) confirm that our proposed Lyapunov function \( V_{k,p} \) is indeed a proper choice, i.e., satisfying the condition (1) in Section \ref{sec:challenge}. In certain situations, it is necessary to establish an upper bound on higher-order moments, such as \( \|\theta - \theta^*\|^{2+p} \). Lemma~\ref{lem:V}--(1) allows us to achieve this by directly bounding \( V_{k,p}(\theta) \). Furthermore, when performing the Taylor expansion as outlined in equation \eqref{eq:taylor}, it is essential to carefully bound the second-order derivative terms. Lemma~\ref{lem:V}--(3) provides tight upper bounds for these terms. Notably, in Lemma~\ref{lem:V}--(3), we present two upper bounds: the first bound is tighter when \( \|\theta - \theta^*\| \leq \Delta \), and the second bound becomes tighter when \( \|\theta - \theta^*\| \geq \Delta \). Proof of Lemma \ref{lem:V} is provided in Appendix \ref{sec:V}. 

Our two pivot results are stated below.

\begin{proposition}\label{prop:V1} Under Assumptions~\ref{assumption:smooth} and \ref{assumption:hubor} with $k \in [1,2)$, there exists $\alpha_{k,0}>0$  such that
    \begin{align*}
\E\left[V_{k, 0}\big(\theta-\alpha(\nabla f(\theta) + w(\theta))\big)\right] \leq (1-\alpha\mu_{k,0})V_{k,0}(\theta) + \alpha^2c'_{k,0}, \quad \forall \theta\in \R^d,\alpha \leq \alpha_{k,0}, 
\end{align*}
where $\mu_{k,0} = \min(\frac{b k}{2\Delta^{2-k}},  \mu )$ and $c_{k,0}'$ is a constant independent of $\alpha$.
\end{proposition}

\begin{proposition}\label{prop:V2}
Under Assumptions~\ref{assumption:smooth} and \ref{assumption:hubor} with $k \in [1,2)$, $\forall p \geq 2$, there exists $\alpha_{k,p}>0$ such that
    \begin{align*}
\E\left[V_{k, p}\big(\theta-\alpha(\nabla f(\theta) + w(\theta))\big)\right] \leq (1-\alpha\mu_{k,p})V_{k, p}(\theta) + \alpha^2c_{k,p}V_{k, p-2}(\theta) + \alpha^{p+2}c_{k,p}', \forall \theta\in \R^d, \alpha \leq \alpha_{k,p},
\end{align*}
where $\mu_{k,p} =\min(\frac{b k}{2\Delta^{2-k}},  \frac{\mu(2+p)}{2})$ and $c_{k,p},c_{k,p}'$ are  constants not depending  on $\alpha$.
\end{proposition}

Proposition \ref{prop:V1} demonstrates a one-step contraction of Lyapunov function $V_{k,0}$ up to a higher-order bias term. {We note that we cannot let $\Delta \to \infty$ in Proposition~\ref{prop:V1} 
to recover the results of strongly convex SGD, because we restrict $k \in [1,2)$, 
which implies $\Delta < \infty$. } Importantly, Proposition \ref{prop:V1} is crucial for deriving second-order moment bounds (cf.\ Section \ref{sec:moment}), as well as fine-grained analysis of the Markov chain under a constant stepsize, including establishing weak convergence, the central limit theorem, and bias characterization results (cf.\ Section \ref{sec:weakconvergence}). The proof of Proposition \ref{prop:V1} depends on Lemma \ref{lem:V} and a precise discussion of the value of noise $w(\theta)$, and is provided in Appendix \ref{sec:V1}.

Although Proposition \ref{prop:V2} does not offer an exact one-step contraction, it establishes a recursive relationship among \( V_{k,p}(\theta_{n+1}) \), \( V_{k,p}(\theta_{n}) \), and \( V_{k,p-2}(\theta_{n}) \), which would allow us to upper bound $\E[V_{k,p}(\theta_n)]$ by employing induction on $p$ and $n$. We emphasize that Propositions \ref{prop:V1} and \ref{prop:V2} together enable us to derive higher moment bounds (detailed in Section \ref{sec:moment}). The proof of Proposition \ref{prop:V2} is provided in Appendix \ref{sec:V2}.

\section{Main Results}\label{sec:mainresults}

%\qxcomment{General comment for this section: Overall our results resemble what has been established for strongly convex setting. Is there anything new we can say for the sub--quadratic setting?}

In this section, we present the main results for the sub--quadratic SGD defined in Section \ref{sec:preliminary}. In Section \ref{sec:moment}, we analyze the moment bounds under both constant stepsize  and diminishing stepsize regimes. In Section \ref{sec:weakconvergence}, we focus on constant stepsize sub--quadratic SGD and examine the weak convergence, central limit theorem and bias characterization of the Markov chain \(\{\theta_n\}_{n \geq 0}\). Surprisingly, we highlight that by our proposed Lyapunov function, we achieve the common results for strongly convex SGD under the sub--quadratic SGD setting.

\subsection{Finite-Time Moment Bound}\label{sec:moment}
In this subsection, we explore the finite-time moment bounds for constant and diminishing stepsizes.
\subsubsection{Moment Bound with Constant Stepsize}
For the SGD update \eqref{eq:SGD} with a constant stepsize (\(\alpha_n \equiv \alpha\)),  the following theorem provides the finite-time bounds for the \(2p\)-th moment of the error $\|\theta_n -\theta^*\|$.

\begin{theorem}[Moment Bounds with Constant Stepsize]\label{thm:constantmoment} Consider dynamic \eqref{eq:SGD} with $\alpha_n \equiv \alpha$.
Under Assumption~\ref{assumption:smooth}--\ref{assumption:uniformlyorlicz} with $k \in [1,2)$,  $\forall p \in \mathbb{N}, \theta_0\in \R^d$, there exists $\bar{\alpha}_{k,p}> 0$ such that when $\alpha \leq \bar{\alpha}_{k,p}$,
    \begin{align*}
\E[\|\theta_n -\theta^*\|^{2p+2}] \leq \frac{2\Delta^2V_{k,2p}(\theta_0)}{k\exp(k/(2-k))}\cdot (1-\alpha\mu_{k,2p})^n + \alpha^{p+1}d_{k,2p}, \quad \forall n \geq \frac{-p\ln(\alpha)}{\alpha \mu_{k,0}},
\end{align*}
where $\mu_{k,2p} =\min(\frac{b k}{2\Delta^{2-k}}, \mu(1+p))$, $V_{k,2p}(\cdot)$ is defined in  \eqref{eq:Vp} and $d_{k,2p}$ is a  constant independent of $\alpha$.
\end{theorem}

Theorem \ref{thm:constantmoment} indicates that with a constant stepsize, the $2p$-order moment \(\E[\|\theta_n - \theta^*\|^{2p}]\) is upper bounded by two terms: one that converges to zero geometrically fast, with a rate scaling with the stepsize $\alpha$; another representing an order-\(\mathcal{O}(\alpha^p)\) bias that does not vanish with the iteration $n$. Interestingly, this result resembles the behavior of constant stepsize SGD for strongly convex objective functions \cite{chen2020finite, zhang2024prelimit}. To the best of our knowledge, this is the first result that establishes high moment bounds for constant stepsize SGD with sub--quadratic objective functions. It is worth pointing out that Theorem \ref{thm:constantmoment} plays a crucial role in analyzing the Markov chain $\{\theta_n\}_{n \geq 0}$ induced by constant stepsize sub--quadratic SGD in the next subsection. Theorem \ref{thm:constantmoment} is proved by applying Propositions \ref{prop:V1} and \ref{prop:V2} on the drift guarantee of our Lyapunov functions to bound $V_{k,2p}$, followed by using Lemma~\ref{lem:V}--(1) of $V_{k,2p}$ to derive the moment bounds. The full proof can be found in Appendix~\ref{sec:constantmoment}.
\subsubsection{Moment Bounds with Diminishing Stepsize}

We next consider the SGD update \eqref{eq:SGD} under a general class of diminishing stepsize with the form $\alpha_n = \frac{\iota}{(n+\kappa)^\xi}$. The following theorem provides the finite-time second moment bounds.
\begin{theorem}[Moment Bounds with Diminishing Stepsize]\label{thm:diminishingmoment}Consider dynamic \eqref{eq:SGD} with $\alpha_n =  \frac{\iota}{(n+\kappa)^{\xi}}$, under Assumptions~\ref{assumption:smooth}--\ref{assumption:uniformlyorlicz} with $k \in [1,2)$, 
$\forall \iota >0$, there exists $\kappa_\iota>0$, such that when we choose $\kappa \geq \kappa_\iota$, $\forall \theta_0\in \R^d,$ we have
\begin{enumerate}
    \item When $\xi = 1$ and $\iota>1/\mu_{k,0}$, for all $n \geq 0$, we have 
    \begin{align*}
    \E[\|\theta_n-\theta^*\|^2] \leq \frac{2\Delta^2V_{k,0}(\theta_{0})}{k\exp(k/(2-k))}\Big(\frac{\kappa}{n+\kappa}\Big)^{\iota\mu_{k,0}}+\frac{8e\iota^2\Delta^2c_{k,0}'}{(\iota\mu_{k,0}-1)k\exp(k/(2-k))}\cdot\frac{1}{n+\kappa}.
    \end{align*}
    \item When $\xi \in (0,1)$, for all $n \geq 0$, we have
    \begin{align*}
    \E[\|\theta_n-\theta^*\|^2] \leq& \frac{2\Delta^2V_{k,0}(\theta_{0})}{k\exp(k/(2-k))}\exp\Big(-\frac{\mu_{k,0}\iota}{1-\xi}\big((n+\kappa)^{1-\xi}-\kappa^{1-\xi}\big)\Big)\\
    &+\frac{4\iota\Delta^2c_{k,0}'}{\mu_{k,0}k\exp(k/(2-k))}\cdot\frac{1}{(n+\kappa)^\xi}.
    \end{align*}
\end{enumerate}
Here $\mu_{k,0} =\min(\frac{b k}{2\Delta^{2-k}}, \mu)$ and $V_{k,0}(\cdot)$ is defined in equation \eqref{eq:V}.
\end{theorem}

Theorem \ref{thm:diminishingmoment} examines how the convergence rate of \(\E[\|\theta_n - \theta^*\|^{2}]\) is influenced by \(\xi\) and \(\iota\) in the stepsize $\alpha_n = \frac{\iota}{(n+\kappa)^\xi}$. Specifically, by setting \(\alpha_n = \frac{\iota}{n+\kappa}\) with \(\iota > 1/\mu_{k,0}\), we obtain the optimal convergence rate of \(\mathcal{O}(1/n)\). In contrast, when \(\xi \in (0,1)\), the convergence rate becomes sub-optimal at \(\mathcal{O}(1/n^\xi)\), but this rate comes with greater robustness, since it does not depend on the choice of \(\iota\). Similar convergence results have been established for SGD with strongly convex functions \cite{chen2020finite}. 

Recent work \cite{gadat2023optimal} also studied sub--quadratic SGD and provided upper bounds on \(\E[\|\theta_n - \theta^*\|^{2p}]\) and \(\E[\|\hat{\theta}_n - \theta^*\|^2]\) under diminishing stepsizes, where \(\hat{\theta}_n := \frac{1}{n} \sum_{t=0}^{n-1} \theta_t\) is the average of the iterates. They further showed that the convergence rate of \(\E[\|\hat{\theta}_n - \theta^*\|^2]\) attains the Cramér-Rao lower bound (CRLB) \cite{rao1992information}.
 It is important to note that \cite{gadat2023optimal} imposes more restrictive assumptions on both the objective function \(f\) and the noise sequence \(\{w_n(\cdot)\}_{n \geq 0}\), requiring $f$ being twice differentiable and $\E[\exp(|w_0|^{4-2\beta})] < \infty$. Additionally, they only consider \(\xi\in[1/2, 1)\), whereas we address the convergence rate for the full range of \(\xi \in (0, 1]\). Furthermore, we remark that in \cite{gadat2023optimal}, the upper bounds on \(\E[\|\theta_n - \theta^*\|^{2p}]\) and \(\E[\|\hat{\theta}_n - \theta^*\|^2]\) are derived based on their key Theorem 11(i). In our work, Propositions \ref{prop:V1} and \ref{prop:V2} on the drift guarantees of our proposed Lyapunov functions play a crucial role in proving Theorems \ref{thm:constantmoment} and \ref{thm:diminishingmoment}, and importantly they imply Theorem 11(i) of \cite{gadat2023optimal}. Therefore, in Theorem \ref{thm:diminishingmoment}, we provide only the second moment bound for raw iterates $\theta_n$. The higher moment bounds and the second moment bounds for the averaged iterates can be directly obtained by using our Propositions \ref{prop:V1} and \ref{prop:V2}, and following the line of argument in \cite{gadat2023optimal}. We will discuss the second moment bounds on the averaged iterates for applications in robust regression and quatile regression in Sections \ref{sec:robustregression} and \ref{sec:quantile}. %Theorem \ref{thm:diminishingmoment} is derived by following the Proposition \ref{prop:V1} and the moment bounds analysis for strongly convex SGD \cite{chen2020finite}. 
 The proof of Theorem \ref{thm:diminishingmoment} can be found in Appendix \ref{sec:diminishingmoment}.

%\qxcomment{As we emphasize the importance of the Lyapunov function for the analysis and the pivot results, it would be nice to discuss how we use these results to derive results in Theorem 1 and 2.}

\subsection{Weak Convergence, Central Limit Theorem and Bias Characterization}\label{sec:weakconvergence}
In this subsection, we study the fluctuations of $\{\theta_n\}_{n \geq 0}$ of sub--quadratic SGD with constant stepsize. 

Since the noise sequence \(\{w_n\}_{n \geq 0}\) are independently and identically distributed and the stepsize $\alpha_n\equiv \alpha$ does not depends on the time step $n$, the sequence of iterates \(\{\theta_n\}_{n \geq 0}\) forms a time-homogeneous Markov chain. Our first goal here is to establish a weak convergence result showing that the Markov chain \(\{\theta_n\}_{n \geq 0}\) converges to a limiting stationary distribution in the Wasserstein-1 distance (\(W_1\)). To this end, we require some additional assumptions.

\begin{assumption}\label{assumption:lipnoise}
There exists $c_w>0$ such that $W_1(\law(w(\theta)), \law(w(\theta'))) \leq c_w\|\theta-\theta'\|,$ $\forall \theta, \theta' \in \R^d.$ 
\end{assumption}

%\qxcomment{Change the constant $W$ to $c_W$ to avoid the confusion with the Wasserstein distance.}
Assumption \ref{assumption:lipnoise} ensures that the variation in the random field \(w(\cdot)\), as measured by the \(W_1\) metric, is controlled by the change in the parameter \(\theta\). This is a common assumption for studying weak convergence in the Wasserstein distance. For SGD with strongly convex objective functions, the authors of \cite{merad2023convergence} assume \(W_2^2(\mathcal{L}(w(\theta)), \mathcal{L}(w(\theta')))\in \mathcal{O}(\|\theta - \theta'\|^2)\) and argue that the co-coercivity in expectation used in \cite{dieuleveut2018bridging} implies their assumption in the linear regression setting. We point out that our Assumption \ref{assumption:lipnoise} is weaker because \(W_1(\mu, \nu)^2 \leq {W_2^2(\mu, \nu)}\) for all \(\mu, \nu \in \mathcal{P}_1(\mathbb{R}^d)\).

\begin{assumption}\label{assumption:A2}
Consider the same $\Delta>0$ in Assumption \ref{assumption:hubor}. There exist $r, \bar{\alpha}>0$ such that $r\bar{\alpha} <1$ and for any two initial points $\theta$ and $\theta'$ with $\theta, \theta' \in \{\theta\in \R^d: \|\theta-\theta^*\| \leq \Delta\}$, we have
\begin{align*}
W_1(\law(\theta-\alpha(\nabla f(\theta) + w(\theta))), \law(\theta'-\alpha(\nabla f(\theta') + w(\theta')))) \leq (1-\alpha r)\|\theta-\theta'\|, \quad \forall   \alpha \leq \bar{\alpha}.
\end{align*}
\end{assumption}
Assumption \ref{assumption:A2} indicates that, given two initial points in a neighborhood around \(\theta^*\), the Wasserstein distance between their subsequent iterates shrinks compared to the Euclidean distance between the initial points. We remark that if Assumption \ref{assumption:hubor} holds and \(\E[\|w(\theta) - w(\theta')\|^2] \in \mathcal{O}(\|\theta - \theta'\|^2)\), then Assumptions \ref{assumption:lipnoise} and \ref{assumption:A2} are readily satisfied. We need Assumption \ref{assumption:A2} since we employ the drift and contraction (D\&C) condition technique for Markov chain convergence analysis \cite{qin2022geometric}, where Assumption \ref{assumption:A2} plays a crucial role in verifying this condition.

Several other conditions have been considered to establish weak convergence results. The drift and minorization (D\&M) condition \cite{meyn1994computable, rosenthal1995minorization, douc2004practical} requires a restrictive minorization condition on the noise, which in general does not hold for discretely distributed noise sequences. Recently, \cite{qu2023computable} introduced a contractive drift (CD) condition and applied their framework to specific sub--quadratic SGD algorithms with only additive noise. Their framework heavily relies on an accurate estimate of the smoothness of the noisy gradient \( \nabla f(\theta) + w(\theta) \). However, when considering a general SGD where \( w(\cdot) \) is multiplicative noise, it becomes unclear how to precisely bound the local Lipschitz constant of the noisy gradient. In this work, we employ the drift and contraction (D\&C) condition approach and establish the following theorem.

\begin{theorem}[Weak Convergence]\label{thm:convergence}
Under Assumptions \ref{assumption:smooth}--\ref{assumption:A2} with $k \in [1,2)$, there exists $\bar{\alpha}_k\geq 0$ such that when $\alpha \leq \bar{\alpha}_k$, there exists a unique limit random variable $\theta_\infty^{(\alpha)}$ such that $\forall \theta_0 \in \R^d,$
\begin{align*}
W_1(\law(\theta_n), \law(\theta_\infty^{(\alpha)})) \leq \frac{\Delta(V_{k,0}(\theta_0)+2)}{\sqrt{k\exp(k/(2-k))}} (1-\rho)^n,\quad \forall n \geq0,
\end{align*}
where $V_{k,0}(\cdot)$ is defined in equation \eqref{eq:V} and $\rho \in \Theta(\alpha)$. Furthermore, $\law(\theta_\infty^{(\alpha)})$ is also the stationary distribution of the Markov chain $\{\theta_n\}_{n \geq 0}$.
\end{theorem}
Theorem \ref{thm:convergence} indicates that the Markov chain \(\{\theta_n\}_{n\geq0}\) generated by the constant stepsize sub--quadratic SGD  converges geometrically to a stationary distribution. Similar weak convergence results have been established for SGD with strongly convex objective functions~\cite{dieuleveut2018bridging, huo2023bias, zhang2024prelimit}. However, the method used in strongly convex setting can not be applied to the sub-quadradic case as discussed in Section \ref{sec:challenge}. While the recent work \cite{qu2023computable} investigates the weak convergence of sub--quadratic SGD with a constant stepsize, they focus on some specific objective functions and only consider additive noise. It remains unclear how to generalize their technique to the general sub--quadratic setting. In our work, by leveraging our newly introduced Lyapunov function \(V_{k,p}(\cdot)\) in Section \ref{sec:Lyapunov}, we establish the weak convergence of sub--quadratic SGD. The proof of Theorem \ref{thm:convergence} is provided in Appendix \ref{sec:convergence}.

One immediate implication of the weak convergence result is the establishment of the central limit theorem (CLT) for the Markov chain \(\{\theta_n\}_{n \geq 0}\), as presented in the following theorem.

\begin{theorem}[Central Limit Theorem]\label{thm:clt}
Under the same setting as Theorem \ref{thm:convergence}, for any 1--Lipschitz $g(\cdot) :\R^d\to \R $, define $S_n(g) = \sum_{t = 0}^{n-1}\big(g(\theta_t) - \E_{\theta \sim \law(\theta_\infty^{(\alpha)})}[g(\theta)]\big)$.
We then have that $\sigma^2(g)=\lim _{n \rightarrow \infty} \frac{1}{n} \E_{\theta_0 \sim \law(\theta_\infty^{(\alpha)})}[S_n^2(g)] $ exists and is finite.  Furthermore, for $\law(\theta_\infty^{(\alpha)})$-almost every point $\theta_0\in \R^d$,
\begin{align*}
S_n(g)/\sqrt{n} \Rightarrow \mathcal{N}\left(0, \sigma^2(g)\right), \quad \text { as } n \rightarrow \infty.
\end{align*}

\end{theorem}
The proof of Theorem \ref{thm:clt} is carried out by verifying Conditions A1 and A2 provided in \cite{jin2020central}, following the weak convergence result. Establishing this CLT is crucial for uncertainty quantification and statistical inference \cite{li2023statistical}. Similar results have been established for SGD with quadratic tails \cite{yu2021analysis, merad2023convergence} and for Q--learning \cite{zhang2024constant, xie2022statistical}. The detailed proof is provided in Appendix \ref{sec:clt}.

Building upon the weak convergence result, and under the additional assumption that the objective function is differentiable to a higher order, we can further characterize the asymptotic bias \(\mathbb{E}[\theta^{(\alpha)}_\infty] - \theta^*\), as presented in the following corollary.
\begin{corollary}[Bias Characterizaion]\label{co:bias}
Under the same setting as Theorem \ref{thm:convergence} and further assuming that the objective function $f$ is three times differentiable, then there exists $\bar{\alpha}_k' \geq 0$, such that 
\begin{align*}
\E[\theta_\infty^{(\alpha)}] = \theta^* + \alpha B + \mathcal{O}(\alpha^{3/2}),
\end{align*}
where $B = f''(\theta^*)^{-1}f'''(\theta^*)(f^{\prime \prime}\left(\theta^*\right) \otimes I_d+I_d \otimes f^{\prime \prime}(\theta^*))^{-1}S(\theta^*)$ and $S(\theta) = \E[w(\theta)w(\theta)^T]$.
\end{corollary}
We emphasize that employing a constant stepsize in sub--quadratic SGD leads the raw iterates \(\{\theta_n\}_{n \geq 0}\) to converge to a limiting random variable \(\theta_\infty^{(\alpha)}\) at a geometric rate, as stated in Theorem \ref{thm:convergence}. This convergence rate surpasses the \(\mathcal{O}(1/n)\) rate achieved with diminishing stepsizes, as presented in Theorem \ref{thm:diminishingmoment}. However, using a constant stepsize induces an asymptotic bias \(\mathbb{E}[\theta_\infty^{(\alpha)}] - \theta^*\) that is in general not zero. As demonstrated in Corollary \ref{co:bias}, this asymptotic bias is proportional to \(\alpha\) up to higher-order terms. This finding has important algorithmic implications for reducing the bias to higher orders of \(\alpha\) through Richardson-Romberg (RR) extrapolation technique \cite{hildebrand1987introduction}, as discussed in prior work \cite{huo2023bias, zhang2024constant, zhang2024prelimit, huo2024collusion}. Therefore, by applying RR extrapolation to constant stepsize sub--quadratic SGD, one can achieve fast convergence with a reduced bias term.

Leveraging Theorem \ref{thm:constantmoment} alongside Fatou's lemma \cite[Theorem 1.6.5]{durrett2019probability}, we can show that \(\E[\|\theta_\infty^{(\alpha)} - \theta^*\|^2] \in \mathcal{O}(\alpha)\) and \(\E[\|\theta_\infty^{(\alpha)} - \theta^*\|^4] \in \mathcal{O}(\alpha^2)\). Consequently, Corollary \ref{co:bias} follows directly by building on Theorems \ref{thm:constantmoment} and \ref{thm:convergence}, as well as by following the proof argument for \cite[Theorem 4]{dieuleveut2018bridging}. Therefore, we omit the proof of Corollary \ref{co:bias}.

\section{Application to Online Robust Regression}\label{sec:robustregression}

In this section, we examine our main results in Section \ref{sec:mainresults} in the context of online robust regression.

\subsection{Model Setup}
 We assume that we have access to i.i.d. data sequence $\{(x_n,y_n)\}_{n \geq 0}$ from the online oblivious response corruption model \cite{pesme2020online}:
\begin{align}
    y = x^T\theta_{\operatorname{reg}}^* +\epsilon + s ,\label{eq:robustmodel}
\end{align}
where $\theta_{\operatorname{reg}}^* \in \R^d$ is the true parameter we wish to recover, $x \in \R^d$ is the zero mean covariate, $\epsilon \in \R$ is the zero mean noise  and $s\in \R$ is the corruption, and $x,\epsilon,s$ are  independent with each other.

Given the differentiable loss function $l(\cdot): \R \to \R^+,$ the population-level loss and gradient are defined as:
\begin{align*}
f_{\operatorname{reg}}(\theta) = \E[l( y -  x^T \theta)] \quad \text{ and } \quad \nabla f_{\operatorname{reg}}(\theta) = -\E[l'( y -  x^T \theta)x].
\end{align*}
It is easy to verify that $\nabla f_{\operatorname{reg}}(\theta_{\operatorname{reg}}^*) = \E[l'(\epsilon+s)x] =\E[l'(\epsilon+s)]\E[x]= 0$. Given the i.i.d. data sequence $\{(x_n,y_n)\}_{n \geq 0}$, we consider the online robust regression that performs the following iterative update: 
\begin{align}\label{eq:robust}
\theta_{n+1} &= \theta_n + \alpha_n l'( y_n - x_n^T \theta_n) x_n.
\end{align}
For model $\eqref{eq:robustmodel}$, we consider the following assumption.
\begin{assumption}\label{assumption:data1}
The covariate $x$, the noise $\epsilon$ and the corruption $s$ are independent random variables that satisfy the following properties:
\begin{enumerate}
    \item $x$ is a zero mean random variable such that $\E[xx^T] = I_d$ and $\|x\|$ is $\sigma_x$-sub--exponential.
    \item $\epsilon$ is a zero mean random variable such that  $\E[|\epsilon|] < \infty.$
    \item $s$ is a random variable such that $\E[|s|] <\infty$ 
\end{enumerate}
\end{assumption}
We note that the first two conditions in Assumption \ref{assumption:data1} are standard in robust regression \cite{loh2017statistical, loh2021scale}. These conditions focus on settings where the covariate \(x\) is isotropically distributed and sub--exponential, and the noise \(\epsilon\) possesses only a finite first moment. This framework includes many heavy--tailed noise distributions, such as \(\alpha\)-stable distributions for \(\alpha \in (1,2]\). We highlight that the assumption of isotropically distributed can be easily relaxed to $\E[xx^T]$ being positive definite. For the ease of exposition, we focus on the setting $\E[xx^T] = I_d$. 

Regarding corruption \(s\), offline robust regression typically assumes \(\eta\)-corruption, where at most an \(\eta\) fraction of the offline dataset is corrupted. In contrast, the online oblivious response corruption model~\cite{pesme2020online} assumes that \(s\) follows a specific distribution and is independent of \((x, \epsilon)\).

Furthermore, Assumption \ref{assumption:data1}--(3) ensures that the population-level loss \(f_{\operatorname{reg}}(\theta)\) is well-defined for all \(\theta \in \mathbb{R}^d\), particularly for loss functions with at least linear growth. Prior work~\cite{pesme2020online} requires \(\mathbb{E}[s \operatorname{erf}(s/c)] < \infty\), where \(\operatorname{erf}(\cdot)\) is the Gaussian error function. Since \(s \operatorname{erf}(s/c) \in \Theta(|s|)\), this condition is equivalent to our assumption \(\mathbb{E}[|s|] < \infty\). Additionally, \cite{pesme2020online} assumes that both covariates and noise are Gaussian, whereas we consider broader distribution classes.

For the loss function $l(\cdot)$, we make the following assumption.
\begin{assumption}\label{assumption:l}
The loss function $l(\cdot)$ satisfies the following properties:
\begin{enumerate}
\item There exists $L_l \geq 0$ such that $    |l'(t)-l'(t')| \leq L_l|t-t'|, \quad \forall t, t' \in \R.$
    \item $l'(\cdot)$ is non-decreasing, $l'(0)=0$ and there exists $a_l>0$ such that $        |l'(t)| \leq a_l, \quad \forall t \in \R.$
\item There exist $\Delta_l, \mu_l >0$ such that for all $t,t' \in \{t\in \R:|t| \leq \Delta_l\}$ and $t \geq t'$,
\begin{align*}
    l'(t)-l'(t') \geq \mu_l(t-t') .
\end{align*}
\end{enumerate}
\end{assumption}
By Assumption \ref{assumption:l}, our analysis focuses on a class of loss functions that are both locally strongly convex and exhibit linear growth. Notably, this class cover many widely-used robust loss functions, including the Huber loss \cite{huber1992robust}, pseudo-Huber loss \cite{hartley2003multiple}, and log-cosh loss \cite{saleh2022statistical}.

Lastly, we impose an additional assumption that the corrupted noise term \(\epsilon + s\) has a strictly positive probability mass within the strongly convex region of the loss function \( l(\cdot) \).
\begin{assumption}\label{assumption:data2}
There exists $\Delta_{\epsilon,s} < \Delta_l$ such that $\mathbb{P}(|\epsilon+s| \leq \Delta_{\epsilon,s})>0.$
\end{assumption}
A few remarks are in order. Assumption \ref{assumption:data2} is satisfied when $P(s \neq 0)<1$ and the noise $\epsilon$ follows a continuous distribution, which is common in robust regression setting \cite{bhatia2017consistent,dalalyan2019outlier, pesme2020online}. Furthermore, by defining $\Tilde{\eta} = \mathbb{P}(|\epsilon+s| \geq \Delta_l)$, we note that Assumption \ref{assumption:data2} holds if and only if $\Tilde{\eta} <1.$ Here we call $\Tilde{\eta}$ as the effective outlier proportion as defined in \cite{pesme2020online}. 

It is clear that the online robust regression update  \eqref{eq:robust} can be cast as an SGD update as in \eqref{eq:SGD},
\begin{align*}
\theta_{n+1} = \theta_n - \alpha_n \big( \nabla f_{\operatorname{reg}}(\theta_n) +  w_{\operatorname{reg},n}(\theta_n) \big),
\end{align*}
%\begin{align*}
%\theta_{n+1} = \theta_n - \alpha_n \big( \nabla f_{\operatorname{reg}}(\theta_n) +  \underbrace{(-\nabla f_{\operatorname{reg}}(\theta_n)-l'( y_n - x_n^T \theta_n) x_n)}_{w_{\operatorname{reg},n}(\theta)} \big),
%\end{align*}
where the noise sequence  $\{w_{\operatorname{reg},n}(\theta)\}_{n \geq 0} \overset{\operatorname{i.i.d.}}{\sim} w_{\operatorname{reg}}(\theta) = -\nabla f_{\operatorname{reg}}(\theta)-l'( y - x^T \theta) x$. 
\subsection{Main Results for Online Robust Regression}
We verify that the population-level gradient $\nabla f_{\operatorname{reg}}(\theta)$ and the noise term $w_{\operatorname{reg}}(\cdot)$ satisfy the assumptions required for the main results in Section \ref{sec:mainresults}, as stated in the following theorem.

\begin{theorem}\label{thm:robust}
Under Assumptions~\ref{assumption:data1}--\ref{assumption:data2},  the online robust regression update \eqref{eq:robust} can be reformulated as a sub--quadratic SGD satisfying Assumptions \ref{assumption:smooth}--\ref{assumption:A2}.
Specifically, Assumptions \ref{assumption:smooth}--\ref{assumption:A2} hold with $L = L_l\E[\|x\|^2], \mu = \frac{\mu_l}{2}\mathbb{P}(|\epsilon+s| \leq \Delta_{\epsilon,s}), a =a_l\E[\|x\|] , b \in \mathcal{O}(\frac{1}{\E[\|x\|^4]}), \Delta = \frac{\Delta_l-\Delta_{\epsilon,s}}{\sigma_x\ln(8\E[\|x\|^4])}, k =1 ,c_w =2L_\tau\sqrt{\E[\|x\|^4]} $ and $ r = \frac{\mu_l}{4}\mathbb{P}(|\epsilon+s| \leq \Delta_{\epsilon,s})$.
\end{theorem} 

The proof of Theorem \ref{thm:robust} is provided in Appendix \ref{sec:proofrobust}. We highlight that verifying the last condition of Assumption \ref{assumption:hubor} (i.e., $\langle \theta - \theta^*_{\operatorname{reg}}, \nabla f_{\operatorname{reg}}(\theta) \rangle \geq b \|\theta - \theta^*_{\operatorname{reg}}\|$ when $\|\theta-\theta^*_{\operatorname{reg}}\| \geq \Delta$) is the most challenging part and the proof can be outlined in the following three main steps:
\begin{enumerate}[leftmargin=*]
    \item Prove that $\langle \theta - \theta^*_{\operatorname{reg}}, \nabla f_{\operatorname{reg}}(\theta) \rangle > 0$ for all $\theta \neq \theta^*_{\operatorname{reg}}$.
    \item Show that there exist $b'>0$ and {$\Delta' \gg \Delta$} such that $\langle \theta - \theta^*_{\operatorname{reg}}, \nabla f_{\operatorname{reg}}(\theta) \rangle \geq b' \|\theta - \theta^*_{\operatorname{reg}}\|$ when $\|\theta-\theta^*_{\operatorname{reg}}\| \geq \Delta'.$
    \item By step (1), we have $\min_{\theta\in \R^d : \Delta\leq \|\theta-\theta^*_{\operatorname{reg}}\| \leq \Delta'}\langle \theta - \theta^*_{\operatorname{reg}}, \nabla f_{\operatorname{reg}}(\theta) \rangle>0$. Then, there always exists
    \begin{align*}
        b = \min \bigg\{ b', \frac{\min_{\theta\in \R^d : \Delta\leq \|\theta-\theta^*_{\operatorname{reg}}\| \leq \Delta'}\langle \theta - \theta^*_{\operatorname{reg}}, \nabla f_{\operatorname{reg}}(\theta) \rangle}{\Delta'} \bigg\}>0
    \end{align*} 
    such that  $\langle \theta - \theta^*_{\operatorname{reg}}, \nabla f_{\operatorname{reg}}(\theta) \rangle \geq b \|\theta - \theta^*_{\operatorname{reg}}\|$ when $\|\theta-\theta^*_{\operatorname{reg}}\| \geq \Delta$.
\end{enumerate}

Building upon Theorem \ref{thm:robust}, we investigate the main results from Section \ref{sec:mainresults} in the context of online robust regression \eqref{eq:robust}, as presented in the following corollaries.
\begin{corollary}[Moment bounds]\label{co:robustmoments}
Consider the dynamic \eqref{eq:robust} under Assumptions \ref{assumption:data1}--\ref{assumption:data2}. We have:
\begin{enumerate}
    \item When $\alpha_n \equiv \alpha$, $\forall p \in \mathbb{N}$, there exists $\alpha_{\operatorname{reg}, p} \geq 0$ such that when $\alpha \leq \alpha_{\operatorname{reg}, p}$, we have
    \begin{align*}
\E[\|\theta_n -\theta^*_{\operatorname{reg}}\|^{2p+2}] \leq \frac{2(\Delta_l-\Delta
_{\epsilon,s})^2V_{1,2p}(\theta_0)}{e\sigma_x^2\ln^2(8\E[\|x\|^4])}\cdot (1-\alpha\mu_{\operatorname{reg}, 2p})^n + \alpha^{p+1}d_{\operatorname{reg}, 2p}, \/ \forall \theta_0 \in \R^d, n \geq \frac{-p\ln(\alpha)}{\alpha \mu_{\operatorname{reg}, 0}}.
\end{align*}
\item When $\alpha_n = \frac{\iota}{n+\kappa}$ with $\iota > 1/\mu_{\operatorname{reg}, 0}$, there exist $\kappa_\iota > 0$ such that when we choose $\kappa \geq \kappa_\iota$, for all $n\geq 0$ and $\theta_0 \in \R^d$, we have
\begin{align*}
\E[\|\theta_n-\theta^*_{\operatorname{reg}}\|^2] \leq \frac{2(\Delta_l-\Delta
_{\epsilon,s})^2V_{1,0}(\theta_0)}{e\sigma_x^2\ln^2(8\E[\|x\|^4])}(\frac{\kappa}{n+\kappa})^{\iota\mu_{\operatorname{reg}, 0}} + \frac{8e\iota^2(\Delta_l-\Delta
_{\epsilon,s})^2c_{1,0}'}{(\iota\mu_{\operatorname{reg}, 0}-1)e\sigma_x^2\ln^2(8\E[\|x\|^4])}\cdot\frac{1}{n+\kappa}.
\end{align*}
\item When $\alpha_n = \frac{\iota}{(n+\kappa)^{\xi}}$ with $\xi \in (0,1)$,  for all $n\geq 0$ and $\theta_0 \in \R^d$,
\begin{align*}
\E[\|\theta_n-\theta^*_{\operatorname{reg}}\|^2] \leq& \frac{2(\Delta_l-\Delta
_{\epsilon,s})^2V_{1,0}(\theta_0)}{e\sigma_x^2\ln^2(8\E[\|x\|^4])}\exp\Big(-\frac{\mu_{\operatorname{reg},0}\iota}{1-\xi}\big((n+\kappa)^{1-\xi}-\kappa^{1-\xi}\big)\Big)\\
    &+\frac{4\iota(\Delta_l-\Delta
_{\epsilon,s})^2c_{1,0}'}{\mu_{\operatorname{reg},0}e\sigma_x^2\ln^2(8\E[\|x\|^4])}\cdot\frac{1}{(n+\kappa)^\xi}.
\end{align*}
\item When $\alpha_n = \frac{\iota}{n^{\xi}}$ with $\xi \in [1/2,1)$,  if $l(\cdot)$ is twice differntiable, for all $n\geq 0$ and $\theta_0 \in \R^d$,
\begin{align*}
\E[\|\hat{\theta}_n-\theta^*_{\operatorname{reg}}\|^2] \leq \frac{d\E[l'(\epsilon+s)^2]}{n\E[l''(\epsilon+s)]^2} + \mathcal{O}\left(\frac{1}{n^{(\xi+1/2)\wedge(2-\xi)}}\right),
\end{align*}
where $\frac{d\E[l'(\epsilon+s)^2]}{n\E[l''(\epsilon+s)]^2}$ is the CRLB of robust regression with loss function $l(\cdot)$ and a sample size $n$.
\end{enumerate}
Here $\mu_{\operatorname{reg}, 2p} \in \mathcal{O}(\mu_l(1-\tilde{\eta})(1+p))$, $V_{1,2p}(\cdot)$ is defined in equation \eqref{eq:Vp}, $c_{1,0}'$ is defined in Proposition \ref{prop:V1} and $d_{\operatorname{reg}, 2p}$ is a constant not depending  on $\alpha$.
\end{corollary}

The first three statements in Corollary \ref{co:robustmoments} on the moment bounds of raw iterates $\theta_n$ follow directly from Theorems \ref{thm:constantmoment}, \ref{thm:diminishingmoment}, and \ref{thm:robust}. The last statement on the mean-square error of the averaged iterate $\hat{\theta}_n$ is obtained by using Propositions \ref{prop:V1} and \ref{prop:V2}, along with the verification of Assumption \( H_S \) introduced in \cite{gadat2023optimal}. To the best of our knowledge, statements (1)--(3) provide the first results on non-asymptotic higher moment bounds under constant stepsizes and second moment bounds under general diminishing stepsizes for online robust regression. Additionally, by the definition of $\mu_{\operatorname{reg},2p}$, we conclude that the larger the effective outlier proportion $\tilde{\eta}$ is, the slower the algorithm \eqref{eq:robust} converges, which aligns with the intuition that the more dispersed the distribution of outliers, the more difficult it is for the algorithm to converge. Proof of Corollary \ref{co:robustmoments} are provided in Appendix~\ref{sec:quantilehs}. 

It is worth pointing out that the last statement of Corollary \ref{co:robustmoments} combined with Assumption \ref{assumption:l} allows us to derive the upper bound $\E[\|\hat{\theta}_n - \theta^*_{\operatorname{reg}}\|^2] \in \mathcal{O}(\frac{d}{n(1-\tilde{\eta})^2}).$ This is because $l'$ is non-decreasing, \( |l'| < \infty \) and \( |l''(x)| \geq \mu_l \) when $l''(\cdot)$ exists and \( |x| \leq \Delta_l \). This bound is consistent with the convergence rate reported in \cite{pesme2020online} for a more restrictive setting. %\qxcomment{If the bound already appeared in prior work, why we want to include this part here?}\yzcomment{a more general setting can achieve the same result?}

\begin{corollary}[Weak Convergence, Central Limit Theorem and Bias Characterization]\label{co:robustweak}
Consider the dynamic \eqref{eq:robust} under Assumptions \ref{assumption:data1}--\ref{assumption:data2} and $\alpha_n \equiv \alpha$. We have the following:
\begin{enumerate}
    \item There exists $\alpha_{\operatorname{reg}}>0$ such that when $\alpha \leq \alpha_{\operatorname{reg}}$, there exists a unique limit random variable $\theta_\infty^{(\alpha)}$ such that $\forall \theta_0 \in \R^d$,
    \begin{align*}
W_1(\law(\theta_n),\law(\theta_\infty^{(\alpha)})) \leq \frac{(\Delta_l-\Delta
_{\epsilon,s})(V_{1,0}(\theta_0)+2)}{\sqrt{e}\sigma_x\ln(8\E[\|x\|^4])}(1-\rho)^n,\quad \forall n \geq0,
    \end{align*}
    where $\rho \in \Theta(\alpha)$ and $\law(\theta_\infty^{(\alpha)})$ is also the stationary distribution of the Markov chain $\{\theta_n\}_{n \geq 0}$.
    \item For $\law(\theta_\infty^{(\alpha)})$-almost every point $\theta_0 \in \R^d$, we have the CLT as stated in Theorem \ref{thm:clt}.
    \item If the regression function $l(\cdot)$ is three times differentiable, we have
\begin{align*}
\E[\theta_\infty^{(\alpha)}] = \theta^*_{\operatorname{reg}} + \alpha B + \mathcal{O}(\alpha^{3/2}), \quad \text{with }  B = -\frac{1}{2}\E[l'''(\epsilon+s)]\E[l'(\epsilon+s)^2]\E[x\|x\|^2]/\E[l''(\epsilon+s)]^2.
\end{align*}
\end{enumerate}
\end{corollary}

Notably, Corollary \ref{co:robustweak} provides the first Markov chain analysis of constant stepsize online robust regression. In particular, the weak convergence and CLT results could be potentially leveraged for statistical inference tasks, such as constructing confidence intervals. The final statement ensures that we can apply Richardson--Romberg extrapolation technique, as discussed under Corollary \ref{co:bias}, to construct iterates that not only converge geometrically but also achieve a reduced bias. Corollary \ref{co:robustweak} follows directly from Theorems \ref{thm:convergence}--\ref{thm:robust} and Corollary \ref{co:bias}. Therefore, we omit the proof of Corollary \ref{co:robustweak}.

\section{Application to Online Quantile Regression}\label{sec:quantile}
In this section, we apply our main results in Section \ref{sec:moment} to the context of online quantile regression.
\subsection{Model Setup}
 For a given $\tau \in (0,1)$, we assume that we have access to i.i.d. data sequence $\{x_n,y_n\}_{n \geq 0}$ from the following classical quantile regression model \cite{chen2019quantile}
\begin{align}\label{eq:quantilemodel}
y = x^T\theta^*_\tau+ \epsilon,
\end{align}
where the covariate $x$ is a random variable supported on $\Omega_x \subseteq \R^d$ and the error $\epsilon \in \R$ satisfies
\begin{align*}
\mathbb{P}(\epsilon \leq 0 \mid x) = \tau, \quad \forall x \in \Omega_x,
\end{align*}
which implies that $x^T\theta^*_\tau$ is the $\tau$--quantile of $y$ conditioned on $x.$ 

We denote $F_x(\cdot)$ to be the cumulative distribution function (CDF) of $\epsilon$ given x. Consequently, we have $F_x(0) = \tau$ for all $x \in \Omega_x$. We focus on the setting where the covariate $x$ and the conditional CDF $F_x(\cdot)$ satisfies the following assumption.
\begin{assumption}\label{assumption:quantile}
The covariate $x$ and the conditional cumulative distribution function $F_x(\cdot)$ satisfy the following properties:
\begin{enumerate}
    \item $\E[xx^T] = I_d$ and $\|x\|$ is $\sigma_x$--sub--exponential. 
      \item There exists $L_\tau \geq 0$ such that for all $t, t' \in \R$, $   | F_x(t) - F_x(t') | \leq L_\tau | t - t' |, \quad \forall x \in \Omega_x.$
    \item There exist $\Delta_\tau, \mu_\tau>0$ such that for all $t, t' \in \{ t \in \mathbb{R}: | t | \leq \Delta_\tau \}$ and $t \geq t'$, 
    $$
    F_x(t) - F_x(t') \geq \mu_\tau (t - t') , \quad \forall x \in \R^d.
    $$
\end{enumerate}
\end{assumption}

Similar to the robust regression setting in Section \ref{sec:robustregression}, here we assume $x$ to be isotropically distributed (i.e. $\E[xx^T] = I_d$) for the ease of exposition and it is easy to generalize $\E[xx^T]$ to be positive definite. We note that  Assumption \ref{assumption:quantile}-(1) is common in classical quantile regression, while previous work further assumes $\|x\|$ is either bounded \cite{fernandes2021smoothing} or sub--Gaussian \cite{chen2019quantile, shen2024online}. In contrast, we only require $\|x\|$ to be sub--exponential. Assumption \ref{assumption:quantile}--(2) requires $F_x(\cdot)$ being uniformly Lipschitz continuous for all $x \in \Omega_x$. {Previous work \cite{fernandes2021smoothing, wang2022renewable, jiang2022renewable, shen2024online} further assumes the existence and continuity of conditional density function $p_x(\cdot)$ of $F_x(\cdot)$ with respect to the Lebesgue measure} and the uniformly boundness of  $p_x(\cdot)$, which implies Assumption \ref{assumption:quantile}--(2). The last condition guarantees the uniformly strong monotonicity of $F_x(\cdot)$ in a neighborhood of 0. Such a locally strong monotonicity condition is also standard \cite{fernandes2021smoothing, wang2022renewable, jiang2022renewable, shen2024online} and can be easily verified if the density $p_x(\cdot)$   is uniformly lower bounded by a positive constant within a neighborhood of 0. Notably, when Assumption \ref{assumption:quantile} holds, $0$ is the unique point such that $F_x(0) = \tau$ for all $x \in \Omega_x.$

Importantly, thanks to our proposed piecewise Lyapunov functions \eqref{eq:V} 
and \eqref{eq:Vp}, {Assumption~\ref{assumption:quantile} does not require 
the continuity of the conditional density function $p_x(\cdot)$. This flexibility 
allows for many conditional distributions of $\epsilon$ whose support is bounded 
and whose density $p_x(\cdot)$ may be discontinuous at the support boundaries. 
Another example is a mixture of a continuous distribution and a point mass, 
commonly observed in applications such as survival analysis~\cite{klein2006survival}, econometrics~\cite{wooldridge2010econometric}, or 
insurance modeling~\cite{klugman2012loss}. In such cases, the density is discontinuous at 
the point mass. In contrast to our Assumption~\ref{assumption:quantile}, most prior work relies on the continuity of 
$p_x(\cdot)$}, and some work~\cite{costa2021non,shen2024online} further assumes 
that $\epsilon$ has a finite first moment.
 
 The online quantile regression update \cite{shen2024online} is expressed as:
\begin{align}
\theta_{n+1} = \theta_n - \alpha_n ( \mathds{1}_{\{ y_n - x_n^T\theta_n \leq 0 \}} - \tau )x_n.\label{eq:quantile}
\end{align}
We define the population-level gradient $\nabla f_\tau(\theta):= \E[( \mathds{1}_{\{ y - x^T\theta \leq 0 \}} - \tau )x]$. After taking the conditional expectation over $\epsilon$ given $x$, by model \eqref{eq:quantilemodel}, we have 
\begin{align*}
\nabla f_\tau(\theta) = \E[\E[( \mathds{1}_{\{ y - x^T\theta \leq 0 \}} - \tau )x|x]] = \E[(F_x(x^T\theta-x^T\theta^*_\tau)-\tau)x].
\end{align*}
It is easy to verify that $\nabla f_\tau(\theta^*_\tau) = 0$ and we can further reinterpret  \eqref{eq:quantile} as a SGD update as in \eqref{eq:SGD} with the noise sequence  $\{w_{\tau,n}(\theta)\}_{n \geq 0} \overset{\operatorname{i.i.d.}}{\sim} w_\tau(\theta) = (\mathds{1}_{\{y - x^T\theta \leq 0\}} - \tau)x - \nabla f_\tau(\theta)$. 

We note that when \( x \equiv 1 \), the update equation~\eqref{eq:quantile} simplifies to the recursive quantile estimation \cite{costa2021non}. Consequently, our results in Section~\ref{sec:quantileresults} also apply to the recursive quantile estimation setting.

\subsection{Main Results for Online Quantile Regression}\label{sec:quantileresults}
  We verify that the population-level gradient $\nabla f_\tau(\cdot)$ and the noise term $w_\tau(\cdot)$ satisfy the assumptions required for the main results in Section \ref{sec:moment}, as stated in the following theorem.
\begin{theorem}\label{thm:quantile}
Under Assumption~\ref{assumption:quantile}, the online quantile regression update \eqref{eq:quantile} can be reformulated as a sub--quadratic SGD satisfying Assumptions \ref{assumption:smooth}--\ref{assumption:uniformlyorlicz}.
Specifically, Assumptions \ref{assumption:smooth}--\ref{assumption:uniformlyorlicz} hold with with $L = L_\tau\E[\|x\|^2], \mu = \mu_\tau/2, a =(1+\tau)\E[\|x\|] , b \in \mathcal{O}(1/\E[\|x\|^4]),  \Delta = \frac{\Delta_\tau}{\sigma_x\ln(8\E[\|x\|^4])}$ and $k =1$.
\end{theorem} 

The proof of Theorem \ref{thm:quantile} is in Appendix \ref{sec:quantileproof}. Since the update \eqref{eq:quantile} is not smooth with respect to $\theta$, it is unclear whether constant stepsize online quantile regression exhibits weak convergence; we discuss why this may not hold in Appendix \ref{sec:quantileweak}. Nonetheless, building on Theorem \ref{thm:quantile}, we extend the main results from Section \ref{sec:moment} to online quantile regression \eqref{eq:quantile}, as shown in the following corollary.

\begin{corollary}[Moment bounds]\label{co:quantilemoments}
Consider the dynamic \eqref{eq:quantile} under Assumption \ref{assumption:quantile}. We then have:
\begin{enumerate}
    \item When $\alpha_n \equiv \alpha$, $\forall p \in \mathbb{N}$, there exists $\alpha_{\tau, p} \geq 0$ such that when $\alpha \leq \alpha_{\tau, p}$,
    \begin{align*}
\E[\|\theta_n -\theta^*_\tau\|^{2p+2}] \leq \frac{2\Delta_\tau^2V_{1,2p}(\theta_0)}{e\sigma_x^2\ln(8\E[\|x\|^4])^2}\cdot (1-\alpha\mu_{\tau,2p})^n + \alpha^{p+1}d_{\tau, 2p}, \quad \forall n \geq \frac{-p\ln(\alpha)}{\alpha \mu_{\tau,0}}.
\end{align*}
\item When $\alpha_n = \frac{\iota}{n+\kappa}$ with $\iota > 1/\mu_{\tau,0}$, there exist $\kappa_\iota > 0$ such that when we choose $\kappa \geq \kappa_\iota$, for all $n\geq 0$ and $\theta_0 \in \R^d$,
\begin{align*}
\E[\|\theta_n-\theta^*_\tau\|^2] \leq \frac{2\Delta_\tau^2V_{1,0}(\theta_0)}{e\sigma_x^2\ln(8\E[\|x\|^4])^2}(\frac{\kappa}{n+\kappa})^{\iota\mu_{\tau,0}} + \frac{8e\iota^2\Delta_\tau^2c_{1,0}'}{(\iota\mu_{\tau,0}-1)e\sigma_x^2\ln(8\E[\|x\|^4])^2}\cdot\frac{1}{n+\kappa}.
\end{align*}
\item When $\alpha_n = \frac{\iota}{(n+\kappa)^{\xi}}$ with $\xi \in (0,1)$,  for all $n\geq 0$ and $\theta_0 \in \R^d$,
\begin{align*}
\E[\|\theta_n-\theta^*_\tau\|^2] \leq& \frac{2\Delta_\tau^2V_{1,0}(\theta_{0})}{e\sigma_x^2\ln(8\E[\|x\|^4])^2}\exp\Big(-\frac{\mu_{\tau,0}\iota}{1-\xi}\big((n+\kappa)^{1-\xi}-\kappa^{1-\xi}\big)\Big)\\
    &+\frac{4\iota\Delta_\tau^2c_{1,0}'}{\mu_{\tau,0}e\sigma_x^2\ln(8\E[\|x\|^4])^2}\cdot\frac{1}{(n+\kappa)^\xi}.
\end{align*}
\item When $\alpha_n = \frac{\iota}{n^{\xi}}$ with $\xi \in [1/2,1)$,  if the conditional density $p_x(\cdot)$ is continuous, for all $n\geq 0$ and $\theta_0 \in \R^d$,
\begin{align*}
\E[\|\hat{\theta}_n-\theta^*_\tau\|^2] \leq \frac{\tau(1-\tau)\operatorname{Tr}(\E[p_x(0)xx^T]^{-1}\cdot\E[p_x(0)xx^T]^{-1})}{n} + \mathcal{O}\left(\frac{1}{n^{(\xi+1/2)\wedge(2-\xi)}}\right),
\end{align*}
where the first term is the CRLB of quantile regression with a sample size $n$.
\end{enumerate}
Here $\mu_{\tau,2p} \in \mathcal{O}(\mu_\tau(1+p)/2)$, $V_{1,2p}(\cdot)$ is defined in equation \eqref{eq:Vp}, $c_{1,0}'$ is defined in Proposition \ref{prop:V1} and $d_{\tau, 2p}$ is a constant not depending  on $\alpha$.
\end{corollary}

%In addition to the discussions in Corollary~\ref{co:robustmoments}, 
To our best knowledge, only statement (2) of Corollary~\ref{co:quantilemoments} has been established in recent work \cite{shen2024online} under the assumption that the covariate \( x \) is Gaussian distributed. Through Corollary~\ref{co:quantilemoments}, we provide a more complete analysis of online quantile regression under different stepsizes and weaker assumptions. By the definition of $\mu_{\tau,2p}$ and the first three statements, we conclude that if the conditional CDF $F_x$ is flatter around $0$, $\mu_\tau$ would be smaller and leads to a slower convergence rate under both constant and diminishing stepsizes. Notably, statement (4) is the \emph{first} result on the convergence of the averaged iterates with a convergence rate that meets the CRLB for online quantile regression. Under Assumption~\ref{assumption:quantile}, we can further lower bound \( p_x(0) \) by \( \mu_\tau \), yielding 
\begin{align*}
\E[\|\hat{\theta}_n - \theta^*_\tau\|^2] \in \mathcal{O}\Big(\frac{d \tau (1 - \tau)}{n \mu_\tau^2}\Big).
\end{align*}
We also note that, \cite{costa2021non} considered recursive quantile estimation, a special instance of quantile regression, and established statement (4) by requiring \( \epsilon \) to have a finite second moment. In this work, when considering recursive quantile estimation (\( x \equiv 1 \)), by defining \( p_1(0) \) as the density of \( \epsilon \) at 0, we obtain $\E[(\hat{\theta}_n - \theta^*_\tau)^2] \in \mathcal{O}\Big(\frac{ \tau (1 - \tau)}{n p_1(0)^2}\Big)$, which aligns with the results in \cite{costa2021non}, but we require a weak assumption on the error $\epsilon.$  Proof of Corollary \ref{co:quantilemoments} is provided in Appendix~\ref{sec:quantilehs}.

\section{Numerical Experiments}
In this section, we present numerical experiments for both online robust regression, as defined in equation~\eqref{eq:robust}, and online quantile regression, as defined in equation~\eqref{eq:quantile}.

For online robust regression, we consider the model \( y = x\theta^*_{\operatorname{reg}} + \epsilon + s \), where \( \mathbb{P}(x = 3) = 0.25 \), \( \mathbb{P}(x = -1) = 0.75 \), \( \theta^*_{\operatorname{reg}} = 0 \), \( \epsilon \) follows a Student's \( t \)-distribution \( t_\nu \) with degrees of freedom \( \nu = 1.1 \), and \( s = 0.01 \). The loss function uses the pseudo-Huber loss \( l(t) = \sqrt{1 + t^2} - 1 \).

For online quantile regression, we consider the model \( y = x\theta^*_\tau + \epsilon \), where \( x \sim \mathcal{N}(0,1) \), \( \theta^*_\tau = 0 \), and \( \epsilon \sim \operatorname{Cauchy}(-1,1) \). Consequently, for all \( x \in \mathbb{R} \), we have \( \mathbb{P}(y \leq 0 \mid x) = \mathbb{P}(\epsilon \leq 0) = 0.75 \), and we perform online quantile regression with \( \tau = 0.75 \).

\begin{figure}[htbp]
    \centering
    % First Row
    \begin{subfigure}[t]{0.23\textwidth}
        \centering
        \includegraphics[width=1.05\textwidth]{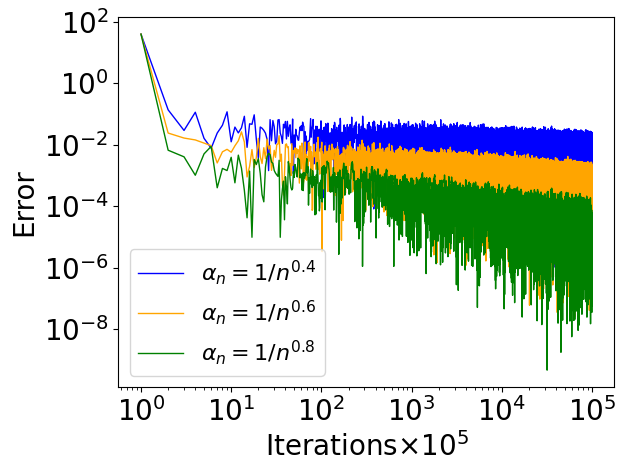}
        \caption{Online robust regression error}
        \label{fig:robustdimi}
    \end{subfigure}
    \hfill
    \begin{subfigure}[t]{0.23\textwidth}
        \centering
        \includegraphics[width=1.05\textwidth]{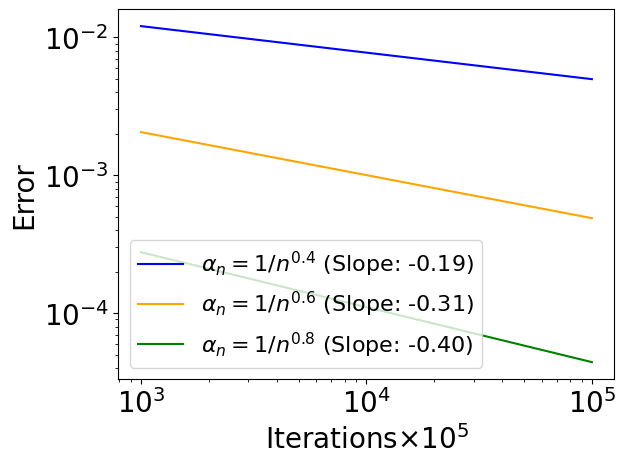}
        \caption{{Smoothed online robust regression error}}
        \label{fig:robustdimislope}
    \end{subfigure}
    \hfill
\begin{subfigure}[t]{0.23\textwidth}
        \centering
        \includegraphics[width=1.05\textwidth]{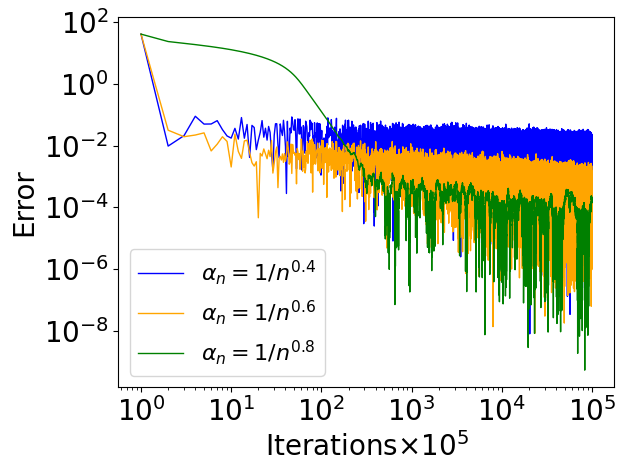}
        \caption{Online quantile regression error}
        \label{fig:quantiledimi}
    \end{subfigure}
    \hfill
    \begin{subfigure}[t]{0.23\textwidth}
        \centering
        \includegraphics[width=1.05\textwidth]{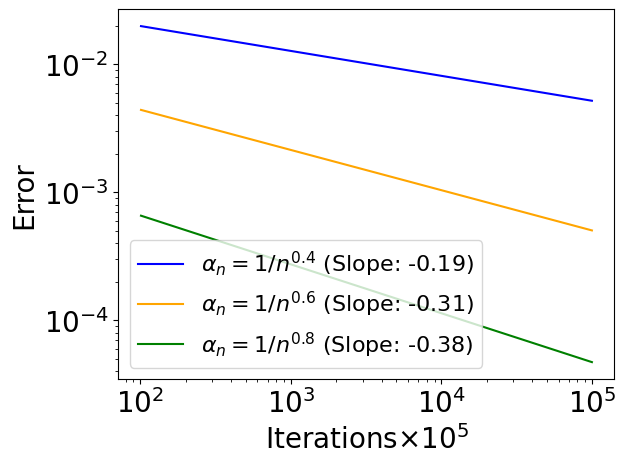}
        \caption{{Smoothed online quantile regression error}}
        \label{fig:quantiledimislope}
    \end{subfigure}

    \caption{Convergence with different diminishing stepsizes and the convergence rate}
    \label{fig:bias}
\end{figure}

First, we run both algorithms with diminishing stepsizes \( \alpha_n = 1/n^\xi\) and \(\xi \in \{0.4,0.6,0.8\} \) and an identical initial point \( \theta_0 = 40 \). For both online robust regression and online quantile regression, we perform  \( 10^{10} \) iterations. We plot the error \( |\theta_n - \theta^*| \) for both algorithms (where \( \theta^* \) denotes \( \theta^*_{\operatorname{reg}} \) or \( \theta^*_\tau \) in the corresponding settings), as shown in Figures~\ref{fig:robustdimi} and \ref{fig:quantiledimi}. We observe that for all diminishing stepsizes, both algorithms converge, and converge more rapidly after $10^8$ iterations when using a larger \( \xi \). Additionally, we smooth the last iterates using a sliding window median, approximate it with a linear function, and calculate the slope, as depicted in Figures~\ref{fig:robustdimislope} and \ref{fig:quantiledimislope}. For both algorithms, the convergence rate of \( |\theta_n - \theta^*| \) is approximately \( \xi/2 \), which aligns with Corollaries~\ref{co:robustmoments}--(3) and \ref{co:quantilemoments}--(3) for the online robust regression and online quantile regression, respectively.

For online robust regression, we also conducted experiments with constant stepsizes to verify our Markov chain results presented in Corollary~\ref{co:robustweak}. Our first experiment demonstrates the asymptotic normality of the averaged iterates of online robust regression. Using the same model as before, we consider different initializations \( \theta_0 \), various numbers of iterations \( n \), and different stepsizes \( \alpha = 0.4 \) and \( \alpha' = 0.42 \). We plot the density of \( n^{-1/2}S_n(\phi) = n^{-1/2}\sum_{k=1}^n \phi(\theta_k) \) with the test function \( \phi(\theta_k) = |\theta_k - \theta^*_{\operatorname{reg}}| \) over 4000 Monte Carlo runs.

\begin{figure}[htbp]
    \centering

    % Second Row
    \begin{subfigure}[t]{0.3\textwidth}
        \centering
        \includegraphics[width=\textwidth]{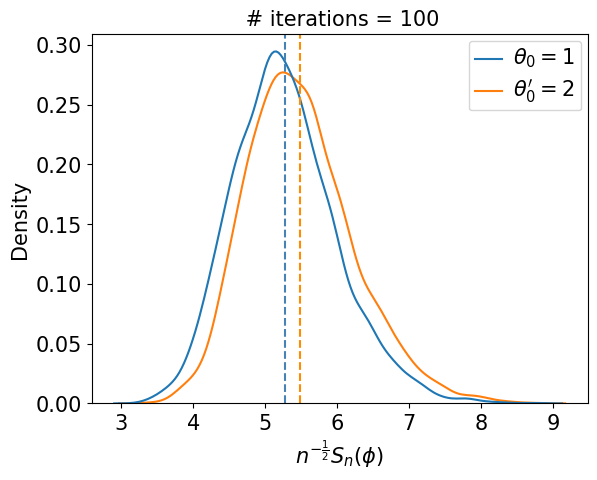}
        \caption{Different initializations with the same stepsize. {Run a small number ($10^2$) of iterations.}}
        \label{fig:clt100}
    \end{subfigure}
    \hfill
    \begin{subfigure}[t]{0.3\textwidth}
        \centering
        \includegraphics[width=\textwidth]{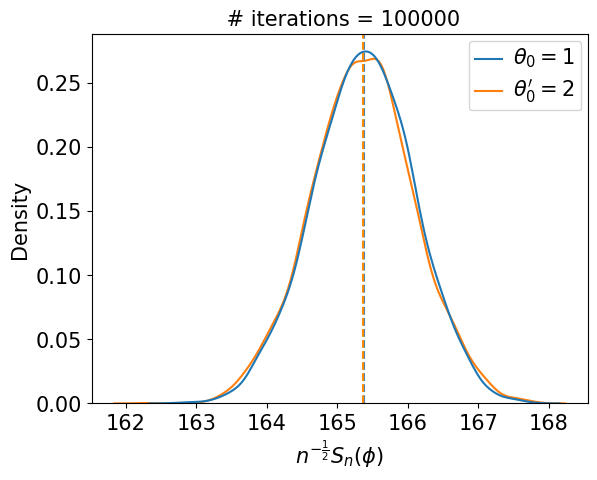}
        \caption{Different initializations with the same stepsize. {Run a large number ($10^5$) of iterations.}}
        \label{fig:clt100000}
    \end{subfigure}
    \hfill
    \begin{subfigure}[t]{0.3\textwidth}
        \centering
        \includegraphics[width=\textwidth]{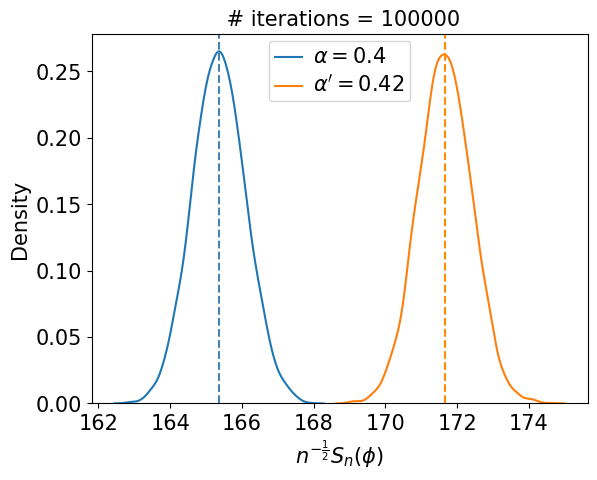}
        \caption{Different stepsizes with the same initialization. {Run a large number ($10^5$) of iterations.}}
        \label{fig:cltalpha}
    \end{subfigure}

    \caption{Asymptotic Normality for online robust regression}
    \label{fig:clt}
\end{figure}

Figure~\ref{fig:clt100} shows the effect of different initializations (represented by the blue and orange curves) on the normality of the distribution after a moderate number of iterations, specifically \( n = 10^2 \). We note that the influence of the initialization diminishes over time, as evident in Figure~\ref{fig:clt100000}, where the distribution converges towards a Gaussian form. Additionally, Figure~\ref{fig:cltalpha} illustrates the impact of different stepsizes on normality. In particular, using a larger step size \( \alpha \) (shown by the orange curve) leads to a larger mean value, i.e., larger bias. These findings are consistent with Corollary~\ref{co:robustweak}.

Our next experiment demonstrates the existence of the asymptotic bias $\E[\theta^{(\alpha)}_\infty] - \theta^*_{\operatorname{reg}}$ and the bias characterization stated in Corollary \ref{co:robustweak}--(3). In this part of experiment, we additionally consider another model with different covariate $x\sim\mathcal{N}(0,1)$. For these two models, we run algorithm \eqref{eq:robust} for $10^{10}$ iterations with constant stepsizes $\alpha \in \{0.2,0.4,0.8\}$ and diminishing stepsizes $\alpha_n = 1/n^\xi$ with $\xi \in \{0.75,0.9\}$. Furthermore, as frequently studied in previous works \cite{huo2023bias,huo2024collusion,zhang2024constant}, we also consider the tail-averaged (TA) iterates $\bar{\theta}_n^{(\alpha)} = \frac{1}{n}\sum_{k = 0}^{n-1}\theta_k^{(\alpha)}$, where the superscript $(\alpha)$ denotes the iterates driven by using the constant stepsize $\alpha,$ as well as Richardson-Romberg (RR) Extrapolated iterates $\tilde{\theta}_n^{(\alpha)} = 2\bar{\theta}_n^{(\alpha)}-\bar{\theta}_n^{(2\alpha)}$.  We plot the error of $|\bar{\theta}^{(\alpha)}_n-\theta^*_{\operatorname{reg}}|$ and $|\tilde{\theta}_n^{(\alpha)}-\theta^*_{\operatorname{reg}}|$ for constant stepsizes and the error of $|\theta_n-\theta^*_{\operatorname{reg}}|$ for diminishing stepsizes.

\begin{figure}[htbp]
    \centering
    % First Row
    \begin{subfigure}[t]{0.45\textwidth}
        \centering
        \includegraphics[width=\textwidth]{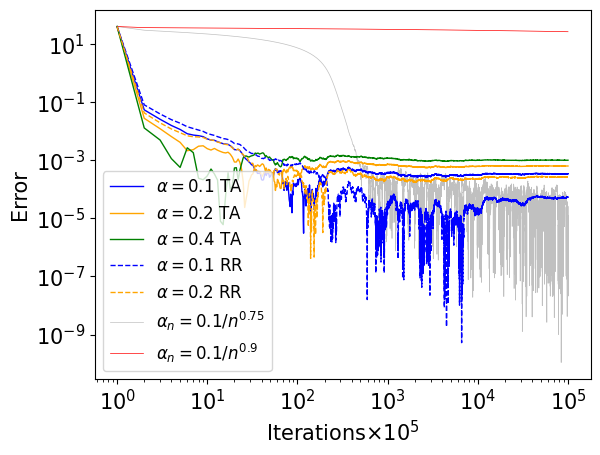}
        \caption{Non-symmetric covariate $x$}
        \label{fig:robustbiasnonsym}
    \end{subfigure}
    \hfill
    \begin{subfigure}[t]{0.45\textwidth}
        \centering
        \includegraphics[width=\textwidth]{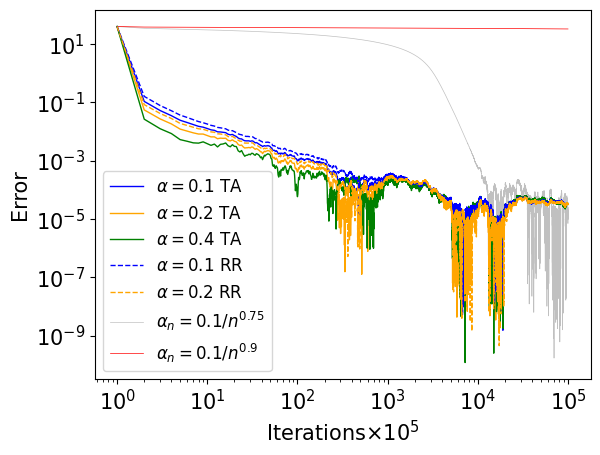}
        \caption{Symmetric covariate $x$}
        \label{fig:robustbiassym}
    \end{subfigure}

    \caption{Error of TA and RR-extrapolated iterates using constant  stepsize comparing with the error of raw iterates using diminishing stepsize for online robust regression}
    \label{fig:robustbias}
\end{figure}
Figures~\ref{fig:robustbiasnonsym} and \ref{fig:robustbiassym} show that using constant stepsizes enables faster convergence for TA iterates compared to diminishing stepsizes, with larger constant stepsizes converging faster. Notably, diminishing stepsizes with a large $\xi = 0.9$ result in slow initial convergence, as observed in the first $10^{10}$ iterations of Figures~\ref{fig:robustbiasnonsym} and \ref{fig:robustbiassym}. When \( x \) is asymmetric, \( \mathbb{E}[x|x|^2] \neq 0 \), leading to a nonzero bias term \( \alpha B \) in Corollary~\ref{co:robustweak}--(3). In Figure~\ref{fig:robustbiasnonsym}, RR-extrapolated iterates further reduce this bias for the first model (\( \mathbb{P}(x = 3) = 0.25 \) and \( \mathbb{P}(x = -1) = 0.75 \)), aligning with Corollary~\ref{co:robustweak}. Conversely--(3), when $x$ is symmetric, the leading term in the bias vanishes (with coefficient $B=0$). As shown in Figure~\ref{fig:robustbiassym}, TA and RR iterates induces similar errors without improvement from RR extrapolation. 

\begin{figure}[htbp]
    \centering
    % First Row
    \begin{subfigure}[t]{0.3\textwidth}
        \centering
        \includegraphics[width=\textwidth]{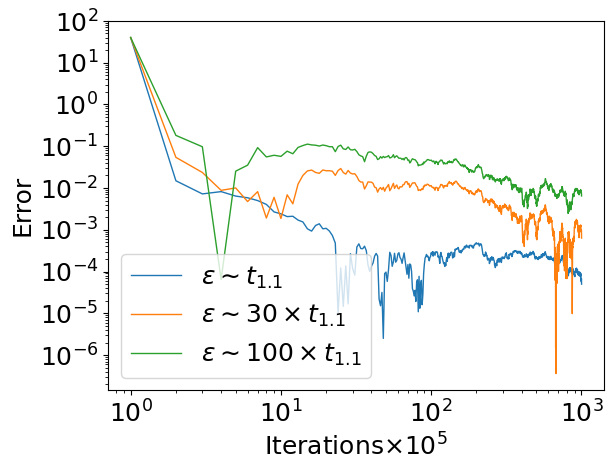}
        \caption{Robust regression with different noise and fixed corruption $s = 0$}
        \label{fig:parameterrobust0}
    \end{subfigure}
    \hfill
    \begin{subfigure}[t]{0.3\textwidth}
        \centering
        \includegraphics[width=\textwidth]{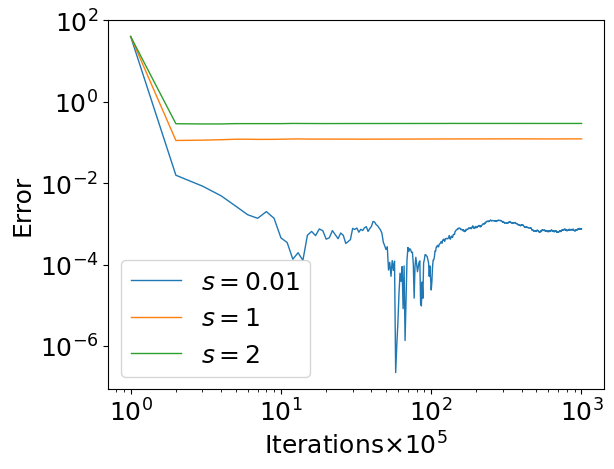}
        \caption{Robust regression with different corruption and fixed noise $\epsilon \sim t_{1.1}$}
        \label{fig:parameterrobust}
    \end{subfigure}
    \hfill
    \begin{subfigure}[t]{0.3\textwidth}
        \centering
        \includegraphics[width=\textwidth]{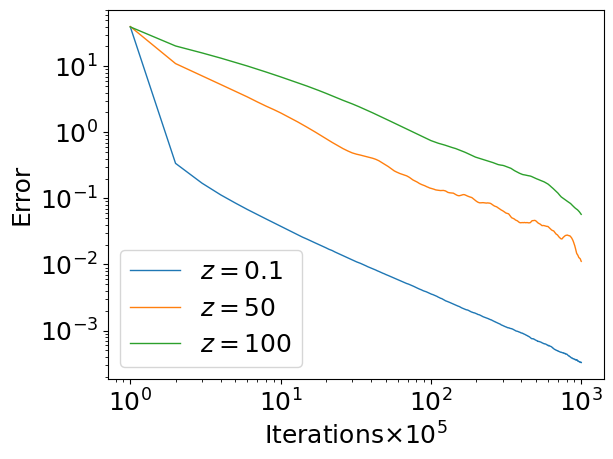}
        \caption{Quantile regression with different density}
        \label{fig:parameterquantile}
    \end{subfigure}

    \caption{Online robust regression and quantile regression with different model parameters}
    \label{fig:parameter}
\end{figure}

{In our last set of experiment, we investigate how the model parameters affect the error. 
For online robust regression, we run the algorithms with a constant stepsize 
\(\alpha = 0.4\) for \(10^8\) iterations. We consider different noise $\epsilon \in \{t_{1.1}, 30 \times t_{1.1}, 100 \times t_{1.1}\}$, where $t_{1.1}$ denotes the Student's t-distribution $t_\nu$ with degrees of freedom $\nu = 1.1$, and different corruption levels 
\(s \in \{0.01,1,2\}\), and evaluate the error of the averaged iterates 
\(\bigl|\bar{\theta}_n^{(\alpha)} - \theta^*_{\mathrm{reg}}\bigr|\). Figures~\ref{fig:parameterrobust0} and \ref{fig:parameterrobust} show that with a noise 
distribution more concentrated near zero and a lower corruption level, 
online robust regression converges faster. This result is consistent with Corollary~\ref{co:robustmoments}--(1), which indicates that a more centered noise and lower corruption level 
reduce the effective outlier proportion \(\tilde{\eta}\), 
thereby increasing \(\mu_{\operatorname{reg},2p}\). Figures~\ref{fig:parameterrobust0} and \ref{fig:parameterrobust} also suggest that a
more centered noise and lower corruption reduce bias. We note that this relationship cannot be directly inferred from Corollary~\ref{co:robustweak}--(3), as the leading term of the bias 
has a complicated form, making it unclear how noise and corruption 
levels jointly influence the bias. 
%We leave a more detailed  investigation of this phenomenon to future work.
For online quantile regression, we use a diminishing stepsize 
\(\alpha_n = 1 / n^{0.5}\) over \(10^8\) iterations and consider 
\(\epsilon \sim \mathrm{Cauchy}(-z,z)\) with \(z \in \{0.1,50,100\}\). 
By setting different values of \(z\), we ensure 
\(\mathbb{P}(\epsilon \leq 0) = 0.75\) and 
\(p_x(0) = \tfrac{1}{2\pi z}\). We then plot the error of the averaged iterates 
\(\bigl|\hat{\theta}_n - \theta^*_{\tau}\bigr|\). 
Figure~\ref{fig:parameterquantile} shows that 
with smaller \(z\)---where the conditional CDF \(F_x\) sharpens around zero--- 
online quantile regression 
converges faster, as discussed under Corollary~\ref{co:quantilemoments}. 
This observation supports the result 
\(\mathbb{E}\bigl[(\hat{\theta}_n - \theta^*_\tau)^2\bigr] \in 
\mathcal{O}\bigl(\frac{\tau(1 - \tau)}{np_1(0)^2}\bigr)\), 
which follows directly from Corollary~\ref{co:quantilemoments}--(4).} 

\section{Conclusion}
In this paper, we study sub--quadratic stochastic gradient descent (SGD) algorithms where the objective function is locally strongly convex with sub--quadratic tails. We introduce a piecewise Lyapunov function that effectively captures the behavior of sub--quadratic SGD, allowing us to relax previous assumptions on the objective function and noise. Utilizing this Lyapunov function, we provide a finer analysis of sub--quadratic SGD, including moment bounds with general stepsizes and results on weak convergence and bias characterization with constant stepsizes. We apply our results to online robust and quantile regression. For online robust regression, we consider a general corrupted linear model with sub--exponential covariates and heavy--tailed noise. Given an effective outlier proportion \( \tilde{\eta} \), we show that using diminishing stepsizes and averaged iterates achieves a convergence rate of \( \mathcal{O}\left( \frac{d}{n(1 - \tilde{\eta})^2} \right) \). We also provide a comprehensive analysis for online robust regression with constant stepsize. For online quantile regression, we remove the previous assumption that the {conditional density of the noise is continuous everywhere} and provide the first convergence rate that achieves the Cramér-Rao lower bound. One direction of immediate interest is extending our results to sub-linear SGD is an interesting direction for future work. We note that our proposed Lyapunov function does not work when \( k < 1 \), and it is not clear whether sub-linear SGD exhibits results similar to sub--quadratic SGD. Exploring weak convergence results for constant stepsize online quantile regression is another interesting future direction.

\section*{Acknowledgement}
Q.\ Xie and Y.\ Zhang are supported in part by National Science Foundation (NSF) grants CNS-1955997 and EPCN-2339794 and EPCN-2432546.
Y.\ Chen is supported in part by NSF grant CCF-2233152. Y.\ Zhang is also supported in part by NSF Award DMS-2023239.

\bibliographystyle{ACM-Reference-Format}
\bibliography{main}

%\appendixpage

\appendix
\section{Proof of Lemma \ref{lem:V}}\label{sec:V}
In this section, we prove the four properties in Lemma \ref{lem:V}.
\paragraph{Polynomial Lower Bound} Define $h(\cdot): [1,\infty]\to \R$ such that $h(x) = \exp(\frac{kx^{2-k}}{2-k})-(1-k/2)\exp(k/(2-k)) - \frac{k\exp(k/(2-k))x^2}{2}$ . Then, we have
\begin{align*}
h'(x) &= kx^{1-k}\exp(\frac{kx^{2-k}}{2-k}) - k\exp(k/(2-k))x = kxg(x).
\end{align*}
where $g(x) := x^{-k}\exp(\frac{kx^{2-k}}{2-k})-\exp(k/(2-k))$.
\begin{align*}
g'(x) &= -kx^{-k-1}\exp(\frac{kx^{2-k}}{2-k}) + kx^{1-2k}\exp(\frac{kx^{2-k}}{2-k})\\
&= -kx^{-k-1}\exp(\frac{kx^{2-k}}{2-k})(1 - x^{2-k})\geq 0.
\end{align*}
Then, we have $g(x) \geq g(1) = 0$, $h'(x) \geq 0$ and $h(x) \geq h(1) = 0$. Thus, substituting $x $ with $\frac{\|\theta-\theta^*\|}{\Delta}$ when $\|\theta-\theta^*\|\geq \Delta,$ we obtain
\begin{align*}
V_{k,0}(\theta) \geq \frac{k\exp(k/(2-k))\|\theta-\theta^*\|^2}{2\Delta^2} \quad \text{when }\|\theta-\theta^*\|\geq \Delta.
\end{align*}
Finally, by the definitions of $V_{k,0}$ and $V_{k,p}$, we have $V_{k,0}(\theta) \geq \frac{k\exp(k/(2-k))\|\theta-\theta^*\|^2}{2\Delta^2}$ and $V_{k,p}(\theta) \geq \frac{k\exp(k/(2-k))\|\theta-\theta^*\|^{2+p}}{2\Delta^2}$ for all $\theta \in \R^d.$
\paragraph{Verifying Twice Differentiability}
We first verify that $V_{k,p}(\cdot)$ is twice differentiable everywhere for all $k \in [1,2)$ and $p \geq 0$. When $p = 0$, with the explicit expression of $V_{k,0}$, we have $V'_{k,0}(\theta) = \frac{k
\exp(\frac{k\|\theta-\theta^*\|^{2-k}}{(2-k)\Delta^{2-k}})}{\Delta^{2-k}\|\theta-\theta^*\|^{k}}\cdot(\theta-\theta^*)$ when $\|\theta-\theta^*\|>\Delta$ and $V'_{k,0}(\theta) =\frac{k\exp(k/(2-k))}{\Delta^2}\cdot(\theta-\theta^*)$ when $\|\theta-\theta^*\|<\Delta$. Because $\frac{k
\exp(\frac{k\|\theta-\theta^*\|^{2-k}}{(2-k)\Delta^{2-k}})}{\Delta^{2-k}\|\theta-\theta^*\|^{k}}\cdot(\theta-\theta^*) = \frac{k\exp(k/(2-k))}{\Delta^2}\cdot(\theta-\theta^*)$ when $\|\theta-\theta^*\|=\Delta$, we have $V_{k,0}$ to be differentiable everywhere for all $k \in [1,2)$ and 
\begin{align}\label{eq:V'}
V'_{k,0}(\theta) = \left\{\begin{array}{l}
\frac{k
\exp(\frac{k\|\theta-\theta^*\|^{2-k}}{(2-k)\Delta^{2-k}})}{\Delta^{2-k}\|\theta-\theta^*\|^{k}}\cdot(\theta-\theta^*)\quad\quad\text { if } |\theta-\theta^*|>\Delta, \\
\frac{k\exp(k/(2-k))}{\Delta^2}\cdot(\theta-\theta^*)  \quad\quad\text { if } \|\theta-\theta^*\| \leq \Delta.
\end{array}\right.
\end{align}
For the second order derivative, we have $V''_{k,0}(\theta) = (1 - \frac{\Delta^{2-k}
}{\|\theta-\theta^*\|^{2-k}})\cdot\frac{k^2\exp(\frac{k\|\theta-\theta^*\|^{2-k}}{(2-k)\Delta^{2-k}})
}{\Delta^{4-2k}\|\theta-\theta^*\|^{2k}}\cdot(\theta-\theta^*)(\theta-\theta^*)^T + \frac{k
\exp(\frac{k\|\theta-\theta^*\|^{2-k}}{(2-k)\Delta^{2-k}})}{\Delta^{2-k}\|\theta-\theta^*\|^{k}}\cdot I_d$  when $\|\theta-\theta^*\|>\Delta$ and $V''_{k,0}(\theta) =\frac{k\exp(k/(2-k))}{\Delta^2}\cdot I_d$ when $\|\theta-\theta^*\|<\Delta$. Because $(1 - \frac{\Delta^{2-k}
}{\|\theta-\theta^*\|^{2-k}})\cdot\frac{k^2\exp(\frac{k\|\theta-\theta^*\|^{2-k}}{(2-k)\Delta^{2-k}})
}{\Delta^{4-2k}\|\theta-\theta^*\|^{2k}}\cdot(\theta-\theta^*)(\theta-\theta^*)^T + \frac{k
\exp(\frac{k\|\theta-\theta^*\|^{2-k}}{(2-k)\Delta^{2-k}})}{\Delta^{2-k}\|\theta-\theta^*\|^{k}}\cdot I_d = \frac{k\exp(k/(2-k))}{\Delta^2}\cdot I_d$ when $\|\theta-\theta^*\|=\Delta$, we have $V_{k,0}$ to be twice differentiable everywhere for all $k \in [1,2)$ and
\begin{align}\label{eq:V''}
V''_{k,0}(\theta) = \left\{\begin{array}{l}
(1 - \frac{\Delta^{2-k}
}{\|\theta-\theta^*\|^{2-k}})\cdot\frac{k^2\exp(\frac{k\|\theta-\theta^*\|^{2-k}}{(2-k)\Delta^{2-k}})
}{\Delta^{4-2k}\|\theta-\theta^*\|^{2k}}\cdot(\theta-\theta^*)(\theta-\theta^*)^T + \frac{k
\exp(\frac{k\|\theta-\theta^*\|^{2-k}}{(2-k)\Delta^{2-k}})}{\Delta^{2-k}\|\theta-\theta^*\|^{k}}\cdot I_d\quad\quad\text { if } |\theta-\theta^*|>\Delta, \\
\frac{k\exp(k/(2-k))}{\Delta^2}\cdot I_d  \quad\quad\text { if } \|\theta-\theta^*\| \leq \Delta.
\end{array}\right.
\end{align}
To verify that $V_{k,p}(\theta)$ is twice differentiable everywhere for all $k \in [1,2)$ and $p \geq 0$, we only need to verify $V_{k,p}(\theta)$ is twice differentiable around $\theta^*$ for all $k \in [1,2)$ and $p > 0$. When $\|\theta-\theta^*\| < \Delta$, we have $V_{k,p}(\theta) = \frac{k\exp(k/(2-k))\|\theta-\theta^*\|^{2+p}}{2\Delta^2}$. Therefore, we have
\begin{align}
V_{k,p}'(\theta) &= \frac{(2+p)k\exp(k/(2-k))\|\theta-\theta^*\|^{p}}{2\Delta^2}(\theta-\theta^*),\nonumber\\
V_{k,p}''(\theta) &= \frac{(2+p)k\exp(k/(2-k))}{2\Delta^2}(p\|\theta-\theta^*\|^{p-2}(\theta-\theta^*)(\theta-\theta^*)^T + \|\theta-\theta^*\|^{p}I_d)\label{eq:secondaroundoptimal}.
\end{align}
When $\theta = \theta^*$, we define $V_{k,p}'(\theta^*) = 0$ and $V_{k,p}''(\theta^*) = 0$ for all $k \in [1,2)$ and $p >0.$ Then, we have proved that $V_{k,p}(\theta)$ is twice differentiable everywhere for all $k \in [1,2)$ and $p \geq 0$.
\paragraph{Bounding the second derivative}
To bound $\|V_{k,p}''(\theta)\|$,  when $\|\theta-\theta^*\| \leq \Delta,$ by equation \eqref{eq:secondaroundoptimal}, we have
\begin{align*}
\|V_{k,p}''(\theta)\| \leq& \frac{(2+p)(p+1)k\exp(k/(2-k))}{2\Delta^2}\|\theta-\theta^*\|^{p}.
\end{align*}
When $\|\theta-\theta^*\| \geq \Delta,$ we have $V_{k,p}(\theta) = \|\theta-\theta^*\|^p\exp(\frac{k\|\theta-\theta^*\|^{2-k}}{(2-k)\Delta^{2-k}})-(1-k/2)\exp(k/(2-k))\|\theta-\theta^*\|^p$ and 
\begin{align*}
\|V_{k,p}''(\theta)\| \leq& c_k''(1 + p)^2\|\theta-\theta^*\|^{p+2-2k}\exp(\frac{k\|\theta-\theta^*\|^{2-k}}{(2-k)\Delta^{2-k}}),
\end{align*}
where $c_k''\geq 0$ denotes a universal constant depends only on $k$ and $\Delta$.

Therefore, there exist some universal constants $c_k,c_k' \geq 0$ that  depend  only on $k$ and $\Delta$ such that
\begin{align*}
\|V_{k,p}''(\theta)\| \leq& c_k(1 + p)^2\|\theta-\theta^*\|^{p}\exp(\frac{k\|\theta-\theta^*\|^{2-k}}{(2-k)\Delta^{2-k}})\\
\|V_{k,p}''(\theta)\| \leq& c_k'(1 + p)^2\|\theta-\theta^*\|^{p+2-2k}\exp(\frac{k\|\theta-\theta^*\|^{2-k}}{(2-k)\Delta^{2-k}}).
\end{align*}
\paragraph{Valid Lyapunov Function}When $\|\theta-\theta^*\| > \Delta$, we have
\begin{align*}
&\langle V_{k,p}'(\theta), \nabla f(\theta)\rangle  \\
=& \Big((p\|\theta-\theta^*\|^{p-2} + \frac{k\|\theta-\theta^*\|^{p-k}}{\Delta^{2-k}})\exp(\frac{k\|\theta-\theta^*\|^{2-k}}{(2-k)\Delta^{2-k}}) - p(1-k/2)\exp(k/(2-k))\|\theta-\theta^*\|^{p-2}\Big)\\
&\cdot\langle\theta-\theta^*, \nabla f(\theta)\rangle\\
\geq&  \frac{ k\|\theta-\theta^*\|^{p-k}}{\Delta^{2-k}}\exp(\frac{k\|\theta-\theta^*\|^{2-k}}{(2-k)\Delta^{2-k}}) \langle\theta-\theta^*, \nabla f(\theta)\rangle \\
\geq& \frac{b k\|\theta-\theta^*\|^{p}}{\Delta^{2-k}}\exp(\frac{k\|\theta-\theta^*\|^{2-k}}{(2-k)\Delta^{2-k}})\geq \frac{b k}{\Delta^{2-k}}V_{k,p}(\theta).
\end{align*}
When $\|\theta-\theta^*\| \leq \Delta$, we have
\begin{align*}
\langle V_{k,p}'(\theta), \nabla f(\theta)\rangle &=  \frac{(2+p)k\exp(k/(2-k))\|\theta-\theta^*\|^p}{2\Delta^2}\langle\theta-\theta^*, \nabla f(\theta)\rangle\\
&\geq \frac{\mu(2+p)k\exp(k/(2-k))\|\theta-\theta^*\|^{p+2}}{2\Delta^2} = \mu(2+p)V_{k,p}(\theta),
\end{align*}
thereby completing the proof of the last statement of Lemma \ref{lem:V}.

\section{Proof of Proposition \ref{prop:V1}}\label{sec:V1}

To prove Proposition \ref{prop:V1}, the following lemmas play a crucial role, whose proofs are deferred to Appendices \ref{sec:drift} and \ref{sec:decrease}.

\begin{lemma}\label{lem:drift}
Under Assumptions \ref{assumption:smooth} and \ref{assumption:hubor} with $k \in [1,2)$, 
 $\forall \alpha \geq 0, \theta \in \R^d$, we have
\begin{align*}
&\E[V_{k,p}(\theta-\alpha(\nabla f(\theta) + w(\theta)))]-V_{k,p}(\theta)\\
\leq&-\alpha\min(\frac{b k}{\Delta^{2-k}}, , \mu(2+p))V_{k,p}(\theta)+ \alpha^2\E[\max_{y \in [0,1]}\|V_{k,p}''(\theta - y\alpha(\nabla f(\theta)  +  w(\theta)))\|(\|\nabla f(\theta)\|^2+\|w(\theta)\|^2)]
\end{align*}
\end{lemma}

\begin{lemma}\label{lem:decrease}
Under Assumptions \ref{assumption:smooth} and \ref{assumption:hubor} with $k \in [1,2)$, when $\|\theta-\theta^*\| \geq \Delta$ and $\alpha \leq \min(2 b\Delta^{2-k}/a^2,\frac{1}{2a\Delta^{k-2}}) $, we have
\begin{align*}
\frac{1}{2}\|\theta-\theta^*\| \leq \|\theta-\theta^*-\alpha\nabla f(\theta)\| \leq \|\theta-\theta^*\|.
\end{align*}
\end{lemma}

In the following, we bound the term $\E[\max_{y \in [0,1]}\|V_{k,p}''(\theta - y\alpha(\nabla f(\theta)  +  w(\theta)))\|(\|\nabla f(\theta)\|^2+\|w(\theta)\|^2)]$ in Lemma \ref{lem:drift} for $p = 0$.
When $\|\theta-\theta^*\| > \Delta$, we have
\begin{align}
&\E[\max_{y \in [0,1]}\|V_{k,0}''(\theta - y\alpha(\nabla f(\theta)  +  w(\theta)))\|(\|\nabla f(\theta)\|^2+\|w(\theta)\|^2)]\nonumber\\
=&\E[\max_{y \in [0,1]}\|V_{k,0}''(\theta - y\alpha(\nabla f(\theta)  +  w(\theta)))\|(\|\nabla f(\theta)\|^2+\|w(\theta)\|^2)\mathds{1}_{\{\alpha\|w(\theta)\| < \|\theta-\theta^*\|/4\}}]\label{eq:k01}\\
&+\E[\max_{y \in [0,1]}\|V_{k,0}''(\theta - y\alpha(\nabla f(\theta)  +  w(\theta)))\|(\|\nabla f(\theta)\|^2+\|w(\theta)\|^2)\mathds{1}_{\{\alpha\|w(\theta)\| \geq \|\theta-\theta^*\|/4\}}].\label{eq:k02}
\end{align}
For term \eqref{eq:k01}, by Lemma \ref{lem:V}, we have
\begin{align*}
\eqref{eq:k01} \leq& c_k'\E\Big[\max_{y \in [0,1]}\|\theta - y\alpha(\nabla f(\theta)  +  w(\theta))\|^{2-2k}\exp(\frac{k\|\theta - y\alpha(\nabla f(\theta)  +  w(\theta))\|^{2-k}}{(2-k)\Delta^{2-k}})(\|\nabla f(\theta)\|^2+\|w(\theta)\|^2)\\
&\cdot\mathds{1}_{\{\alpha\|w(\theta)\| < \|\theta-\theta^*\|/4\}}\Big]\\
\overset{\text{(i)}}{\leq}&c_k'4^{2k-2}\E[\|\theta-\theta^*\|^{2-2k}\exp(\frac{k\|\theta-\theta^*\|^{2-k}}{(2-k)\Delta^{2-k}})\exp(\frac{\alpha k\|  w(\theta)\|^{2-k}}{(2-k)\Delta^{2-k}})(\|\nabla f(\theta)\|^2+\|w(\theta)\|^2)]\\
\overset{\text{(ii)}}{\leq}&c_k'4^{2k-2}\exp(\frac{k\|\theta-\theta^*\|^{2-k}}{(2-k)\Delta^{2-k}})\E[\exp(\frac{\alpha k\|  w(\theta)\|^{2-k}}{(2-k)\Delta^{2-k}})(a + \|w(\theta)\|^2\Delta^{2-2k})]\\
\overset{\text{(iii)}}{\leq}& c_k'''V_{k,0}(\theta),
\end{align*}
where (i) follows from the following Lemma \ref{lem:decrease} (with the proof deferred to Section \ref{sec:decrease}), (ii) is established by Assumption \ref{assumption:hubor} and (iii) holds since the noise sequence $\{w_n(\cdot)\}_{n \geq 0}$ is uniformly in the $\psi_{2-k}-$Orlicz space and $c_k'''$ is a constant not depending on $\alpha.$

For term \eqref{eq:k02}, by Lemma \ref{lem:V}, we have
\begin{align*}
\eqref{eq:k02}\leq&c_k\E[\max_{y \in [0,1]}\exp(\frac{k\|\theta-\theta^* - y\alpha(\nabla f(\theta)  +  w(\theta))\|^{2-k}}{(2-k)\Delta^{2-k}})(\|\nabla f(\theta)\|^2+\|w(\theta)\|^2)\mathds{1}_{\{\alpha\|w(\theta)\| \geq \|\theta-\theta^*\|/4\}}] \\
\overset{\text{(i)}}{\leq}&c_k\exp(\frac{k\|\theta-\theta^*\|^{2-k}}{(2-k)\Delta^{2-k}})\E[\exp(\frac{k\| w(\theta)\|^{2-k}}{(2-k)\Delta^{2-k}})(a^2\|\theta-\theta^*\|^{2k-2}+\|w(\theta)\|^2)\mathds{1}_{\{\alpha\|w(\theta)\| \geq \|\theta-\theta^*\|/4\}}]\\
\overset{\text{(ii)}}{\leq} &\mathcal{O}(\|\theta-\theta^*\|^{2k-2}\exp(\frac{k\|\theta-\theta^*\|^{2-k}}{(2-k)\Delta^{2-k}}))\mathbb{P}(\alpha\|w(\theta)\| \geq \|\theta-\theta^*\|/4)\\
\leq&\mathcal{O}(\|\theta-\theta^*\|^{2k-2}\exp(\frac{k\|\theta-\theta^*\|^{2-k}}{(2-k)\Delta^{2-k}}))]\cdot\mathcal{O}(\exp(-\frac{\|\theta-\theta^*\|^{2-k}}{\alpha^{2-k}})) \in \mathcal{O}(1)
\end{align*}
where (i) follows from Lemma \ref{lem:decrease} and Assumption \ref{assumption:hubor}, (ii) and (iii) holds since the noise sequence $\{w_n(\cdot)\}_{n \geq 0}$ is uniformly in the $\psi_{2-k}-$Orlicz space.

When $\|\theta-\theta^*\| \leq \Delta$, we have
\begin{align*}
&\E[\max_{y \in [0,1]}\|V_{k,0}''(\theta - y\alpha(\nabla f(\theta)  +  w(\theta)))\|(\|\nabla f(\theta)\|^2+\|w(\theta)\|^2)]\\
\leq&c_k\E[\max_{y \in [0,1]}\exp(\frac{k\|\theta-\theta^* - y\alpha(\nabla f(\theta)  +  w(\theta))\|^{2-k}}{(2-k)\Delta^{2-k}})(\|\nabla f(\theta)\|^2+\|w(\theta)\|^2)] \in \mathcal{O}(1).
\end{align*}

Therefore, there exist $\alpha_k > 0$ and a constant $c_{k,0}'$ such that
\begin{align*}
&\E[V_{k,0}(\theta-\alpha(\nabla f(\theta) + w(\theta)))]\\
\leq& \big(1-\alpha\min(\frac{b k}{\Delta^{2-k}}, , \mu(2+p)) +\alpha^2c_k'''\big)V_{k,0}(\theta) + \alpha^2c_{k,0}'\\
\leq & \big(1-\alpha\min(\frac{b k}{2\Delta^{2-k}}, , \mu(2+p)/2)\big)V_{k,0}(\theta) + \alpha^2c_{k,0}', \quad \forall \alpha \leq \alpha_k,
\end{align*}
thereby completing the proof of Proposition \ref{prop:V1}.

\subsection{Proof of Lemma \ref{lem:drift}}\label{sec:drift}
Given fixed $\theta \in \R^d$, by the fact that $\E[w(\theta)] = 0$, we obtain
\begin{align*}
& \E[V_{k,p}(\theta-\alpha(\nabla f(\theta) + w(\theta)))]-V_{k,p}(\theta)\\
=& -\alpha \langle V_{k,p}'(\theta), \nabla f(\theta)\rangle + \alpha^2\E[\int_0^1 \int_0^x \langle V_{k,p}''(\theta - y\alpha(\nabla f(\theta)  +  w(\theta)))(\nabla f(\theta) + w(\theta)), \nabla f(\theta) + w(\theta)\rangle \d y \d x]\\
\leq& -\alpha \langle V_{k,p}'(\theta), \nabla f(\theta)\rangle + \frac{\alpha^2}{2}\E[\max_{y \in [0,1]}\|V_{k,p}''(\theta - y\alpha(\nabla f(\theta)  +  w(\theta)))\|\|\nabla f(\theta) + w(\theta)\|^2]\\
\leq& -\alpha \langle V_{k,p}'(\theta), \nabla f(\theta)\rangle + \alpha^2\E[\max_{y \in [0,1]}\|V_{k,p}''(\theta - y\alpha(\nabla f(\theta)  +  w(\theta)))\|(\|\nabla f(\theta)\|^2+\|w(\theta)\|^2)]\\
\leq& -\alpha\min(\frac{b k}{\Delta^{2-k}}, , \mu(2+p))V_{k,p}(\theta)+ \alpha^2\E[\max_{y \in [0,1]}\|V_{k,p}''(\theta - y\alpha(\nabla f(\theta)  +  w(\theta)))\|(\|\nabla f(\theta)\|^2+\|w(\theta)\|^2)],
\end{align*}
where the last inequality holds by Lemma \ref{lem:V}. Therefore, we finish proving Lemma \ref{lem:drift}.

\subsection{Proof of Lemma \ref{lem:decrease}}\label{sec:decrease}
\begin{align*}
\|\theta-\theta^*-\alpha\nabla f(\theta)\| &= (\|\theta-\theta^*\|^2 - 2\alpha\langle\theta-\theta^*, \nabla f(\theta)\rangle + \alpha^2\|\nabla f(\theta)\|^2)^{1/2}\\
&\leq (\|\theta-\theta^*\|^2 - 2\alpha b\|\theta-\theta^*\|^k + \alpha^2a^2\|\theta-\theta^*\|^{2k-2})^{1/2},
\end{align*}
where the last inequality holds by Assumption \ref{assumption:hubor}. When $\|\theta-\theta^*\| \geq \Delta$ and $\alpha \leq 2 b\Delta^{2-k}/a^2 $, we have
\begin{align*}
2\alpha b\|\theta-\theta^*\|^k \geq 2\alpha b\Delta^{2-k}\|\theta-\theta^*\|^{2k-2} \geq \alpha^2a^2\|\theta-\theta^*\|^{2k-2}.
\end{align*}
Therefore, we have
\begin{align*}
    \|\theta-\theta^*-\alpha\nabla f(\theta)\| \leq \|\theta-\theta^*\|.
\end{align*}
For the other direction, we have
\begin{align*}
\|\theta-\theta^* - \alpha \nabla f(\theta)\| &\geq \|\theta-\theta^* \|-\alpha\| \nabla f(\theta)\| \\
&\geq \|\theta-\theta^* \|-a\alpha\|\theta-\theta^* \|^{k-1}\\
&\geq \|\theta-\theta^* \|(1-a\alpha\Delta^{k-2}) \\
&\geq \frac{1}{2}\|\theta-\theta^* \|, \quad \forall \alpha \leq \frac{1}{2a\Delta^{k-2}}.
\end{align*}
Therefore, we finish proving Lemma \ref{lem:decrease}.

\section{Proof of Proposition \ref{prop:V2}}\label{sec:V2}
Based on Lemma \ref{lem:drift}, when $\|\theta-\theta^*\| \geq \Delta$ and $p \geq 2$, we have
\begin{align*}
&\E[\max_{y \in [0,1]}\|V_{k,p}''(\theta - y\alpha(\nabla f(\theta)  +  w(\theta)))\|(\|\nabla f(\theta)\|^2+\|w(\theta)\|^2)]\\
\overset{\text{(i)}}{\leq}&c_k'(1+p^2)\E[\max_{y \in [0,1]}\|\theta-\theta^* - y\alpha(\nabla f(\theta)  +  w(\theta))\|^{p+2-2k}\exp(\frac{k\|\theta-\theta^* - y\alpha(\nabla f(\theta)  +  w(\theta))\|^{2-k}}{(2-k)\Delta^{2-k}})\\
&\cdot(\|\nabla f(\theta)\|^2+\|w(\theta)\|^2)] \\
\overset{\text{(ii)}}{\leq}& \mathcal{O}(\exp(\frac{k\|\theta-\theta^* \|^{2-k}}{(2-k)\Delta^{2-k}}))\E[\exp(\frac{k\|  w(\theta)\|^{2-k}}{(2-k)\Delta^{2-k}})(\|\theta-\theta^*\|^{p+2-2k}+\|w(\theta)\|^{p+2-2k})(a^2\|\theta-\theta^* \|^{2k-2}+\|w(\theta)\|^2)]\\
\overset{\text{(iii)}}{\in}& \mathcal{O}(V_{k,p}(\theta)),
\end{align*}
where (i) holds by Lemma \ref{lem:V}, (ii) is established by Lemma \ref{lem:decrease} and (iii) holds by Assumption \ref{assumption:uniformlyorlicz}.

When $\|\theta-\theta^*\| \leq \Delta$, we have
\begin{align*}
&\E[\max_{y \in [0,1]}\|V_{k,p}''(\theta - y\alpha(\nabla f(\theta)  +  w(\theta)))\|(\|\nabla f(\theta)\|^2+\|w(\theta)\|^2)] \\
\overset{\text{(i)}}{\leq}&c_k(1+p^2)\E[\max_{y \in [0,1]}\|\theta-\theta^* - y\alpha(\nabla f(\theta)  +  w(\theta))\|^{p}\exp(\frac{k\|\theta-\theta^* - y\alpha(\nabla f(\theta)  +  w(\theta))\|^{2-k}}{(2-k)\Delta^{2-k}})\\
&\cdot(\|\nabla f(\theta)\|^2+\|w(\theta)\|^2)] \\
\overset{\text{(ii)}}{\leq}& \mathcal{O}(\E[((1+\alpha L)^p\|\theta-\theta^*\|^p + \alpha^p\|w(\theta)\|^p)\exp(\frac{k\|w(\theta)\|^{2-k}}{(2-k)\Delta^{2-k}})(L^2\|\theta-\theta^*\|^2 + \|w(\theta)\|^2)]\\
\overset{\text{(iii)}}{\in} & \mathcal{O}(\|\theta-\theta^*\|^p) + \mathcal{O}(\alpha^p) \in \mathcal{O}(V_{k,p-2}(\theta)) + \mathcal{O}(\alpha^p) 
\end{align*}
where (i) holds by Lemma \ref{lem:V}, (ii) holds by Assumption \ref{assumption:hubor} and the fact that $\|\theta-\theta^*\| \leq \Delta$, and (iii) holds by Assumption \ref{assumption:uniformlyorlicz}. 

Therefore, there exist $\alpha_{k,p} \geq 0$ and constants $c_{k,p},c'_{k,p}$ not depending on $\alpha$ such that 
\begin{align*}
&\E[V_{k,p}(\theta-\alpha(\nabla f(\theta) + w(\theta)))]\\
\leq& \big(1-\alpha\min(\frac{b k}{\Delta^{2-k}}, , \mu(2+p)) +\mathcal{O}(\alpha^2)\big)V_{k,p}(\theta) + \mathcal{O}(\alpha^2)V_{k,p-2}(\theta) + \mathcal{O}(\alpha^{p+2})\\
\leq & \big(1-\alpha\min(\frac{b k}{2\Delta^{2-k}}, , \mu(2+p)/2) \big)V_{k,p}(\theta) + \alpha^2c_{k,p}V_{k,p-2}(\theta) + \alpha^{p+2}c_{k,p}',
\end{align*}
thereby completing the proof of Propsition \ref{prop:V2}.

\section{Proof of Theorem \ref{thm:constantmoment}}\label{sec:constantmoment}
To bound the $2p-$th moment of $(\theta_n-\theta^*)$, the following lemma plays a crucial role, whose proof is deferred to Appendix \ref{sec:boundV}.

\begin{lemma}\label{lem:boundV}
$\forall p \in \mathbb{N}$, given $\alpha \leq \min(\alpha_{k,0}, \alpha_{k,2}, \cdots, \alpha_{k,2p})$, where $\alpha_{k,2p}$ is defined in Propositions \ref{prop:V1} and \ref{prop:V2}, there exists a constant $d_{k,2p}'$ not depending on $\alpha$ such that
\begin{align*}
\E[V_{k,2p}(\theta_{n})] \leq (1-\alpha\mu_{k,2p})^nV_{k,2p}(\theta_0) + d_{k,2p}'\alpha^{p+1}, \quad \forall n \geq \frac{-p\ln(\alpha)}{\alpha\mu_{k,0}}.
\end{align*}
\end{lemma}
By Lemmas \ref{lem:V} and \ref{lem:boundV}, we have
\begin{align*}
\E[\|\theta_n-\theta^*\|^{2+2p}] \leq \frac{2\Delta^2}{k\exp(k/(2-k))}\E[V_{k,2p}(\theta_{n})] \leq \frac{2\Delta^2(1-\alpha\mu_{k,2p})^nV_{k,2p}(\theta_0)}{k\exp(k/(2-k))} + \frac{2\alpha^{p+1}\Delta^2d_{k,2p}}{k\exp(k/(2-k))},
\end{align*}
thereby finishing the proof of Theorem \ref{thm:constantmoment}.

\subsection{Proof of Lemma \ref{lem:boundV}}\label{sec:boundV}

By Proposition \ref{prop:V1}, when $\alpha \leq \alpha_{k,0}$, we have
\begin{align*}
\E[V_{k,0}(\theta_{n})] &\leq (1-\alpha \mu_{k,0})\E[V_{k,0}(\theta_{n-1})] + \alpha^2c_{k,0}'\\
&\leq (1-\alpha \mu_{k,0})^nV_{k,0}(\theta_0) + \sum_{i = 0}^{n-1}(1-\alpha \mu_{k,0})^i\alpha^2c_{k,0}'\\
&\leq (1-\alpha \mu_{k,0})^nV_{k,0}(\theta_0) + \alpha c_{k,0}'/\mu_{k,0}, \quad \forall n \geq 0.
\end{align*}

Assume that for $p \in [l-1]$, we have
\begin{align*}
\E[V_{k,2p}(\theta_n)] \leq& 
(1-\alpha\mu_{k,2p})^nV_{k,2p}(\theta_0) + \sum_{i = 0}^{p-1}\alpha^{p-i}(1-\alpha\mu_{k,2i-2})^nV_{k,2i}(\theta_0)\cdot\prod_{j = i+1}^pc_{k,2j}/\mu_{k,2j}\\
&+\sum_{i = 0}^{p}\alpha^{p+i+1}c_{k,2i}'/\mu_{k,2i}\prod_{j = i+1}^pc_{k,2j}/\mu_{k,2j},\quad \forall n \geq 0, \alpha \leq \min(\alpha_{k,0},\alpha_{k,2}, \cdots, \alpha_{k,2l-2}).
\end{align*}

When $p = l$, By Proposition \ref{prop:V2}, we have
\begin{align*}
\E[V_{k,2l}(\theta_{n})] \leq& (1-\alpha\mu_{k,2l})\E[V_{k,2l}(\theta_{n-1})] + \alpha ^2c_{k,2l}\E[V_{k,2l-2}(\theta_{n-1})] + \alpha^{2l+2}c_{k,2l}'\\
\leq&(1-\alpha\mu_{k,2l})^n\E[V_{k,2l}(\theta_0)] + \alpha c_{k,2l}/\mu_{k,2l}\cdot\E[V_{k,2l-2}(\theta_{n-1})] + \alpha^{2l+1}c_{k,2l}'/\mu_{k,2l}\\
\leq& 
(1-\alpha\mu_{k,2l})^nV_{k,2l}(\theta_0) + \sum_{i = 0}^{l-1}\alpha^{l-i}(1-\alpha\mu_{k,2i-2})^nV_{k,2i}(\theta_0)\cdot\prod_{j = i+1}^lc_{k,2j}/\mu_{k,2j}\\
&+\sum_{i = 0}^{l}\alpha^{l+i+1}c_{k,2i}'/\mu_{k,2i}\prod_{j = i+1}^lc_{k,2j}/\mu_{k,2j},\quad \forall n \geq 0, \alpha \leq \min(\alpha_{k,0}, \alpha_{k,2}, \cdots, \alpha_{k,2l}).
\end{align*}
where the last inequality holds by induction. Therefore, by induction, we have proved that for all $n \geq 0$ and $p \in \mathbb{N}$, when $\alpha \leq \min(\alpha_{k,0}, \alpha_{k,2}\cdots, \alpha_{k,2p})$, we have
\begin{align*}
\E[V_{k,2p}(\theta_{n})] \leq& 
(1-\alpha\mu_{k,2p})^nV_{k,2p}(\theta_0) + \sum_{i = 0}^{p-1}\alpha^{p-i}(1-\alpha\mu_{k,2i})^nV_{k,2i}(\theta_0)\cdot\prod_{j = i+1}^pc_{k,2j}/\mu_{k,2j}\\
&+\sum_{i = 0}^{p}\alpha^{p+i+1}c_{k,2i}'/\mu_{k,2i}\prod_{j = i+1}^pc_{k,2j}/\mu_{k,2j},\quad \forall n \geq 0, \alpha \leq \min(\alpha_{k,0}, \alpha_{k,2}, \cdots, \alpha_{k,2p}).
\end{align*}
In order to obtain $\alpha^{p-i}(1-\alpha\mu_{k,2i})^n \leq \alpha^{p+1}, \forall i\in[p-1]$, it is equivalent to have $n \geq \frac{(i+1)\ln(\alpha)}{\ln(1-\alpha\mu_{k,2i})}, \forall i\in[p-1]$. Notice that 
\begin{align*}
\frac{(i+1)\ln(\alpha)}{\ln(1-\alpha\mu_{k,2i})} \leq \frac{-p\ln(\alpha)}{\alpha\mu_{k,2i-2}} \leq \frac{-p\ln(\alpha)}{\alpha\mu_{k,0}}.
\end{align*}
Therefore, $\forall p \in \mathbb{N}$, given $\alpha \leq \min(\alpha_{k,0}, \alpha_{k,2}, \cdots, \alpha_{k,2p})$, there exists a constant $d_{k,2p}'$ not depending on $\alpha$ such that
\begin{align*}
\E[V_{k,2p}(\theta_{n})] \leq (1-\alpha\mu_{k,2p})^nV_{k,2p}(\theta_0) + d_{k,2p}'\alpha^{p+1}, \quad \forall n \geq \frac{-p\ln(\alpha)}{\alpha\mu_{k,0}}.
\end{align*}

\section{Proof of Theorem \ref{thm:diminishingmoment}}\label{sec:diminishingmoment}
By Proposition \ref{prop:V1}, given $\alpha_0 \leq \alpha_{k,0}$ we have
\begin{align*}
\E[V_{k,0}(\theta_n)] \leq (1-\alpha_{n-1}\mu_{k,0})\E[V_{k,0}(\theta_{n-1})] + \alpha_{n-1}^2c_{k,0}', \forall n \geq 0.
\end{align*}
Therefore, we have
\begin{align*}
\E[V_{k,0}(\theta_n)] \leq \prod_{t = 0}^{n-1}(1-\alpha_{t}\mu_{k,0})V_{k,0}(\theta_{0}) + c_{k,0}' \sum_{t = 0}^{n-1}\alpha_t^2\prod_{u = t+1}^{n-1}(1-\alpha_u\mu_{k,0})
\end{align*}
By Lemma \ref{lem:V}, we have
\begin{align*}
\E[\|\theta_n-\theta^*\|^2] \leq \frac{2\Delta^2V_{k,0}(\theta_{0})}{k\exp(k/(2-k))}\prod_{t = 0}^{n-1}(1-\alpha_{t}\mu_{k,0})+\frac{2\Delta^2c_{k,0}'}{k\exp(k/(2-k))}\sum_{t = 0}^{n-1}\alpha_t^2\prod_{u = t+1}^{n-1}(1-\alpha_u\mu_{k,0}).
\end{align*}
Therefore, by the proof of Corollaries 2.1.1 and 2.1.2 in \cite{chen2020finite}, we finish the proof of Theorem \ref{thm:diminishingmoment}.
\section{Proof of Theorem \ref{thm:convergence}}\label{sec:convergence}
In this section, we prove Theorem \ref{thm:convergence} by verifying conditions A1-A3 in \cite{qin2022geometric}.

\paragraph{Verifying Condition A1}
$\forall \theta, \theta' \in \R^d$, we have
\begin{align*}
\|\theta-\theta'\| &\leq \|\theta-\theta^*\|+\|\theta'-\theta^*\|\\
&= \frac{\Delta}{\sqrt{k\exp(k/(2-k))}}(\frac{\sqrt{k\exp(k/(2-k))}}{\Delta}\|\theta-\theta^*\|+\frac{\sqrt{k\exp(k/(2-k))}}{\Delta}\|\theta'-\theta^*\|)\\
&\leq \frac{\Delta}{\sqrt{k\exp(k/(2-k))}}(\frac{k\exp(k/(2-k))\|\theta-\theta^*\|^{2}}{2\Delta^2}+\frac{k\exp(k/(2-k))\|\theta'-\theta^*\|^{2}}{2\Delta^2}+1)\\
&\leq \frac{\Delta}{\sqrt{k\exp(k/(2-k))}}(V_{k,0}(\theta)+V_{k,0}(\theta')+1),
\end{align*}
where the last inequality holds by Lemma \ref{lem:V}.
%{Yudong: I think $g$ was used to denote something else.}

By Proposition \ref{prop:V1}, we have
\begin{align*}
PV_{k,0}(\theta)-V(\theta) &= \E[V_{k, 0}(\theta-\alpha(\nabla f(\theta) + w(\theta)))]-V_{k,0}(\theta)\\
&\leq \alpha\mu_kV_{k,0}(\theta) + \alpha^2d_{k},
\end{align*}
thereby completing verifying the condition A1.
\paragraph{Verifying Condition A2}
By Assumption \ref{assumption:A2}, we have 
\begin{align*}
W_1(\theta_1, \theta_1') \leq (1-\alpha r)\|\theta_0-\theta_0'\|, \quad \forall \theta_0,\theta_0' \in \{\theta \in \R^d: \|\theta-\theta^*\| \leq \Delta\}.
\end{align*}
By Assumptions \ref{assumption:smooth} and \ref{assumption:lipnoise}, we have
\begin{align*}
W_1(\theta_1, \theta_1') &\leq \E[\|\theta_0-\theta_0' - \alpha (\nabla f(\theta_0)-\nabla f(\theta_0')) - \alpha(w(\theta_0)-w(\theta_0'))\|]\\
&\leq (1+\alpha L + \alpha c_w)\|\theta_0-\theta_0'\|.
\end{align*}
By Lemma \ref{lem:V}, when $V_{k,0}(\theta) \leq \frac{k\exp(k/(2-k))}{2}$, $\|\theta-\theta^*\| \leq \Delta$. 

When $\alpha < \frac{k\exp(k/(2-k))\mu_k}{4d_k}$, we have
\begin{align*}
   \frac{k\exp(k/(2-k))}{2} > \frac{2\alpha^2d_k}{\alpha \mu_k}
\end{align*}.
Then, we can choose $\alpha < \min(\alpha_k, \frac{k\exp(k/(2-k))\mu_k}{4d_k}) $ and conditions A1 and A2 will be satisfied.

\paragraph{Verifying Condition A3}
We aim to verify that 
\begin{align*}
\log (1+\alpha L+\alpha c_w) \log (1+ 2 \alpha^2 d_k)<\log (\frac{1}{1-\alpha r} )\log \big(\frac{\frac{k\exp(k/(2-k))}{2}+1}{\frac{\alpha\mu_kk\exp(k/(2-k))  }{2}+ 2 \alpha^2 d_k+1} \big).
\end{align*}
For the LHS, by the fact that $\log(x) \leq x-1, \forall x >0$,  we have
\begin{align*}
\operatorname{LHS} \leq 2\alpha^3(L+c_w) d_k
\end{align*}
For the RHS, by the fact that $\log(x) \geq 1-1/x, \forall x >0$, we have
\begin{align*}
\operatorname{RHS} &\geq \alpha r \cdot \log \big(\frac{\frac{k\exp(k/(2-k))}{2}+1}{\frac{\alpha\mu_kk\exp(k/(2-k))  }{2}+ 2 \alpha^2 d_k+1} \big)\\
&\geq \alpha r \cdot \log \big(\frac{\frac{k\exp(k/(2-k))}{2}+1}{\frac{\alpha\mu_kk\exp(k/(2-k))  }{2}+ 2 \alpha_k^2 d_k+1} \big).
\end{align*}
Therefore, when $\alpha \leq \sqrt{\frac{r}{2(L+c_w)d_k}\cdot\log \big(\frac{\frac{k\exp(k/(2-k))}{2}+1}{\frac{\alpha\mu_kk\exp(k/(2-k))  }{2}+ 2 \alpha_k^2 d_k+1} \big)}$, we verify condition A3.  

Then, by Corollary 2.1 and Remark 2.3 in \cite{qin2022geometric}, we finish the proof of Theorem \ref{thm:convergence}.

\section{Proof of Theorem \ref{thm:clt}}\label{sec:clt}
In this section, we prove Theorem \ref{thm:clt} by verifying conditions A1 and A2 and leveraging Theorem 9 in \cite{jin2020central}.
\paragraph{Verifying Condition A1}
Condition A1 is satisfied by the geometric convergence rate stated in Theorem \ref{thm:convergence}.
\paragraph{Verifying Condition A2}
Let $\bar{V}_k(\theta) = V_{k,0}^2(\theta)$. Therefore, we have
\begin{align*}
\bar{V}_k(\theta) = 
\begin{cases}
\Big(\exp( \frac{k \|\theta - \theta^*\|^{2 - k}}{(2 - k) \Delta^{2 - k}} ) - (1 - \frac{k}{2}) \exp( k/(2 - k) )\Big)^2 & \text{if } \|\theta - \theta^*\| > \Delta, \\
\frac{k^2 \exp( 2k/(2-k) ) \|\theta - \theta^*\|^4}{4 \Delta^4} & \text{if } \|\theta - \theta^*\| \leq \Delta.
\end{cases}
\end{align*}
Given fixed $\theta \in \R^d$, we have
\begin{align}
&\E[\bar{V}_k(\theta-\alpha(\nabla f(\theta)+w(\theta)))]-\bar{V}_k(\theta)\nonumber\\
=& -\alpha \langle \bar{V}_k'(\theta), \nabla f(\theta)\rangle \nonumber\\
& + \alpha^2\E[\int_0^1 \int_0^x \langle \bar{V}_k''(\theta - y\alpha(\nabla f(\theta)  +  w(\theta)))(\nabla f(\theta) + w(\theta)), \nabla f(\theta) + w(\theta)\rangle \d y \d x] \nonumber\\
\leq& -\alpha \langle \bar{V}_k'(\theta), \nabla f(\theta)\rangle \label{eq:cltterm1}\\
& + \frac{\alpha^2}{2}\E[\max_{y \in [0,1]}\| \bar{V}_k''(\theta - y\alpha(\nabla f(\theta)  +  w(\theta)))\|\|\nabla f(\theta) + w(\theta)\|^2].\label{eq:cltterm2}
\end{align}
Below, we bound the terms \eqref{eq:cltterm1} and \eqref{eq:cltterm2} when $\|\theta-\theta^*\| \leq \Delta$ and $\|\theta-\theta^*\| \geq \Delta.$

When $\|\theta-\theta^*\| \leq \Delta,$ we have
\begin{align*}
\eqref{eq:cltterm1} &= - \frac{\alpha k^2 \exp( 2k/(2-k) ) \|\theta - \theta^*\|^2}{ \Delta^4}\langle \theta-\theta^*, \nabla f(\theta)\rangle\\
&\leq - \frac{\alpha k^2 \mu \exp( 2k/(2-k) ) \|\theta - \theta^*\|^4}{ \Delta^4} = -4\alpha\mu\bar{V}_k(\theta)
\end{align*}
By Assumption \ref{assumption:uniformlyorlicz}, we have $\eqref{eq:cltterm2} \in \mathcal{O}(\alpha^2).$

When $\|\theta-\theta^*\| \geq \Delta,$ we have
\begin{align*}
\eqref{eq:cltterm1} &= -\alpha\Big(\exp( \frac{k \|\theta - \theta^*\|^{2 - k}}{(2 - k) \Delta^{2 - k}} ) - (1 - \frac{k}{2}) \exp( k/(2 - k) )\Big)\frac{k\|\theta-\theta^*\|^{-k}\exp( \frac{k \|\theta - \theta^*\|^{2 - k}}{(2 - k) \Delta^{2 - k}} )}{\Delta^{2-k}}\langle \theta-\theta^*, \nabla f(\theta)\rangle\\
&\leq -\alpha\Big(\exp( \frac{k \|\theta - \theta^*\|^{2 - k}}{(2 - k) \Delta^{2 - k}} ) - (1 - \frac{k}{2}) \exp( k/(2 - k) )\Big)\frac{bk\exp( \frac{k \|\theta - \theta^*\|^{2 - k}}{(2 - k) \Delta^{2 - k}} )}{\Delta^{2-k}}\\
&\leq \frac{-\alpha bk}{\Delta^{2-k}}\bar{V}_k(\theta).
\end{align*}
Notice that 
\begin{align*}
\|\bar{V}''(\theta)\| &= \|2V_{k,0}'(\theta){V_{k,0}'(\theta)}^T + 2V_{k,0}(\theta)V_{k,0}''(\theta) \|\\
&\leq 2\|V_{k,0}'(\theta)\|^2+2|V_{k,0}(\theta)|\|V_{k,0}''(\theta) \|\\
&\in \mathcal{O}(\|\theta-\theta^*\|^{2-2k}\exp( \frac{2k \|\theta - \theta^*\|^{2 - k}}{(2 - k) \Delta^{2 - k}} )),
\end{align*}
where the last inequality holds by following equations \eqref{eq:V'} and \eqref{eq:V''}. Therefore, with the similar arguments in the Section \ref{sec:V1}, we obtain
\begin{align*}
\eqref{eq:cltterm2} \in \mathcal{O}(\alpha^2\bar{V}_k(\theta)) + \mathcal{O}(\alpha^2).
\end{align*}
Therefore, for all $\theta \in \R^d$, there exists $\alpha_c>0$ such that when $\alpha \leq \alpha_c$, we have
\begin{align*}
\E[\bar{V}_k(\theta-\alpha(\nabla f(\theta)+w(\theta)))] &\leq (1-\alpha\min(\frac{bk}{\Delta^{2-k}}, 4\alpha\mu) + \mathcal{O}(\alpha^2))\bar{V}_k(\theta) + \mathcal{O}(\alpha^2)\\
&\leq (1-\alpha\min(\frac{bk}{2\Delta^{2-k}}, 2\alpha\mu) )\bar{V}_k(\theta) + \mathcal{O}(\alpha^2),
\end{align*}
which implies for all $\theta_0 \in \R^d$, we have
\begin{align*}
\E[\bar{V}_k(\theta_n)] \leq (1-\alpha\min(\frac{bk}{2\Delta^{2-k}}, 2\alpha\mu) )^n\bar{V}_k(\theta_0) + \mathcal{O}(\alpha).
\end{align*}
Therefore, by Fatou's lemma (\cite[Theorem 1.6.5]{durrett2019probability}), we have
\begin{align*}
\E[V_{k,0}^2(\theta_\infty^{(\alpha)})] = \E[\bar{V}_k(\theta_\infty^{(\alpha)})] \in \mathcal{\alpha}<\infty,
\end{align*}
thereby verifying the condition A2.

\section{Proof of Theorem \ref{thm:robust}}\label{sec:proofrobust}

\subsection{A Few Preliminary Facts}
\begin{lemma}\label{lem:x}
Given random variable $x\in \R^d$ such that $\E[xx^T] = I_d$ and $\|x\|$ is $\sigma_x $-sub--exponential, there exists $\Delta_x = \sigma_x\ln(8\E[\|x\|^4])$ such that 
\begin{align*}
u^T\E[xx^T\mathds{1}_{\|x\| \leq \Delta_x}]u \geq \frac{1}{2}\|u\|^2, \quad \forall u \in \R^d.
\end{align*}
\end{lemma}
\begin{lemma}\label{lem:robuststrictlylargerthan0}
$\forall \theta \in \R^d, \theta \neq \theta^*_{\operatorname{reg}}$, we have
\begin{align*}
\langle \nabla f_{\operatorname{reg}}(\theta), \theta-\theta^*_{\operatorname{reg}}\rangle>0.
\end{align*}
\end{lemma}
\begin{lemma}\label{lem:robusthubor}
There exists $c_l > 0$ such that
\begin{align*}
tl'(t) \geq c_l|t|, \quad t \in \{t:|t| \geq \Delta_l\}.
\end{align*}
\end{lemma}
\begin{lemma}\label{lem:linear}
Given random variable $x \in \R^d$ such that $\E[xx^T] = I_d$ and $\E[\|x\|^4]<\infty$, for all $a \geq 0$ and $\theta \neq 0$, we have
\begin{align*}
\E[|x^T\theta|\mathds{1}_{|x^T\theta| \geq a}] \leq \frac{9}{32\E[\|x\|^4]}\|\theta\| - a
\end{align*}
\end{lemma}
\subsection{Main Proof}
\paragraph{Verifying Assumption \ref{assumption:smooth}}
$\forall \theta, \theta' \in \R^d$, by Assumption \ref{assumption:l}, we have
\begin{align}
\|\nabla f_{\operatorname{reg}}(\theta)-\nabla f_{\operatorname{reg}}(\theta')\| &= \|\E[ l'( y - x^T \theta) x ]-\E[ l'( y - x^T \theta') x ]\|\nonumber\\
&\leq \E[\|l'( y - x^T \theta) x-l'( y - x^T \theta') x\|]\nonumber\\
&\leq L_l\E[\|x\|^2]\|\theta-\theta'\|, \label{eq:robustsmooth}
\end{align}
thereby completing verifying the Assumption \ref{assumption:smooth} with $L = L_l\E[\|x\|^2]$.
\paragraph{Verifying Assumption \ref{assumption:hubor}}
By Lemma \ref{lem:x}, $\forall \theta', \theta'' \in \{\theta \in \R^d:\|\theta-\theta^*_{\operatorname{reg}}\| \leq \frac{\Delta_l-\Delta_{\epsilon,s}}{\Delta_x}\}$, we have
\begin{align}
&\langle \nabla f_{\operatorname{reg}}(\theta')-\nabla f_{\operatorname{reg}}(\theta''),\theta'-\theta''\rangle\nonumber\\
=& \langle  \E[ l'( y - x^T \theta'') x ]-\E[ l'( y - x^T \theta') x ],\theta'-\theta''\rangle \nonumber\\
=& \E[  (l'( x^T\theta^*_{\operatorname{reg}} - x^T \theta'' +\epsilon+s)- l'( x^T\theta^*_{\operatorname{reg}} - x^T \theta' +\epsilon+s)) (x^T\theta'-x^T\theta'') ]\nonumber\\
\overset{\text{(i)}}{\geq}& \E[(l'( x^T\theta^*_{\operatorname{reg}} - x^T \theta'' +\epsilon+s)- l'( x^T\theta^*_{\operatorname{reg}} - x^T \theta' +\epsilon+s)) (x^T\theta'-x^T\theta'')\mathds{1}_{|x^T\theta^*_{\operatorname{reg}} - x^T \theta' +\epsilon+s| \leq \Delta_l, |x^T\theta^*_{\operatorname{reg}} - x^T \theta'' +\epsilon+s| \leq \Delta_l}]\nonumber\\
\overset{\text{(ii)}}{\geq}&\mu_l\E[(x^T\theta'-x^T\theta'')
^2\mathds{1}_{|x^T\theta^*_{\operatorname{reg}} - x^T \theta' +\epsilon+s| \leq \Delta_l, |x^T\theta^*_{\operatorname{reg}} - x^T \theta'' +\epsilon+s| \leq \Delta_l}]\nonumber\\
\geq &\mu_l\mathbb{P}(|\epsilon+s|\leq \Delta_{\epsilon,s})\E[(x^T\theta'-x^T\theta'')
^2\mathds{1}_{|x^T\theta^*_{\operatorname{reg}} - x^T \theta'| \leq \Delta_l-\Delta_\epsilon, |x^T\theta^*_{\operatorname{reg}} - x^T \theta''| \leq \Delta_l-\Delta_\epsilon}]\nonumber\\
\geq &\mu_l\mathbb{P}(|\epsilon+s|\leq \Delta_{\epsilon,s})(\theta'-\theta'')^T\E[xx^T\mathds{1}_{\|x\| \leq \Delta_x}](\theta'-\theta'')\nonumber\\
\overset{\text{(iii)}}{\geq}& \frac{\mu_l}{2}\mathbb{P}(|\epsilon+s|\leq \Delta_{\epsilon,s})\|\theta'-\theta''\|^2, \label{eq:robuststrongconvex}
\end{align}
where (i) and (ii) hold by Assumption \ref{assumption:l}, and (iii) holds by Assumption \ref{assumption:data1}. Therefore, we have verified the local strong convexity of $f_{\operatorname{reg}}$ when $\|\theta-\theta^*_{\operatorname{reg}}\| \leq \frac{\Delta_l-\Delta_{\epsilon,s}}{\Delta_x}$ with $\mu = \frac{\mu_l}{2}\mathbb{P}(|\epsilon+s|\leq \Delta_{\epsilon,s})$.

For all $\theta \in \R^d,$ by Assumptions \ref{assumption:data1} and \ref{assumption:l}, we have
\begin{align}
\|\nabla f_{\operatorname{reg}}(\theta)\| = \|E[l'(y-x^T\theta)x]\| \leq a_l\E[\|x\|],\label{eq:boundedgradient}
\end{align}
thereby verifying the gradient $\nabla f_{\operatorname{reg}}$ is bounded with $a = a_l\E[\|x\|]$.

To verify the last property in Assumption \ref{assumption:hubor}, we have 
\begin{align*}
&\langle \nabla f_{\operatorname{reg}}(\theta) , \theta-\theta^*_{\operatorname{reg}}\rangle \\
=& \langle \E[l'(x^T\theta^*_{\operatorname{reg}}-x^T\theta + \epsilon+s)x], \theta^*_{\operatorname{reg}} - \theta\rangle\\
=& \E[l'(x^T\theta^*_{\operatorname{reg}}-x^T\theta + \epsilon+s)(x^T\theta^*_{\operatorname{reg}}-x^T\theta)]\\
=& \underbrace{\E[l'(x^T\theta^*_{\operatorname{reg}}-x^T\theta)(x^T\theta^*_{\operatorname{reg}}-x^T\theta)]}_{T_1} + \underbrace{\E[(l'(x^T\theta^*_{\operatorname{reg}}-x^T\theta + \epsilon+s)-l'(x^T\theta^*_{\operatorname{reg}}-x^T\theta))(x^T\theta^*_{\operatorname{reg}}-x^T\theta)]}_{T_2}.
\end{align*}
For term $T_1$, by Lemmas \ref{lem:robusthubor} and \ref{lem:linear} when $\theta\neq \theta^*_{\operatorname{reg}}$, we have
\begin{align*}
T_1 \geq& c_l\E[|x^T\theta^*_{\operatorname{reg}}-x^T\theta|\mathds{1}_{|x^T\theta^*_{\operatorname{reg}}-x^T\theta| \geq \Delta_l}] \\
\geq & \frac{9c_l}{32\E[\|x\|^4]}\|\theta-\theta^*_{\operatorname{reg}}\| - c_l\Delta_l.
\end{align*}

Because $l'(\cdot)$ is increasing and bounded, there exist $c_l'\geq 0$ such that $\|l'(t')-l'(t'')\| \leq \frac{9c_l}{64\E[\|x\|^4]\E[\|x\|]}$ for all $t', t'' \in \{t: t \geq c_l'\}$ and all $t', t'' \in \{t: t \leq -c_l'\}$. Therefore, for term $T_2$, we have
\begin{align*}
T_2 =& \E[(l'(x^T\theta^*_{\operatorname{reg}}-x^T\theta + \epsilon+s)-l'(x^T\theta^*_{\operatorname{reg}}-x^T\theta))(x^T\theta^*_{\operatorname{reg}}-x^T\theta)\mathds{1}_{\{|x^T\theta^*_{\operatorname{reg}}-x^T\theta| \geq 2c_l'\}}]\nonumber\\
&+\E[(l'(x^T\theta^*_{\operatorname{reg}}-x^T\theta + \epsilon+s)-l'(x^T\theta^*_{\operatorname{reg}}-x^T\theta))(x^T\theta^*_{\operatorname{reg}}-x^T\theta)\mathds{1}_{\{|x^T\theta^*_{\operatorname{reg}}-x^T\theta| \leq 2c_l'\}}]\nonumber\\
\geq& \E[(l'(x^T\theta^*_{\operatorname{reg}}-x^T\theta + \epsilon+s)-l'(x^T\theta^*_{\operatorname{reg}}-x^T\theta))(x^T\theta^*_{\operatorname{reg}}-x^T\theta)\mathds{1}_{\{|x^T\theta^*_{\operatorname{reg}}-x^T\theta| \geq 2c_l'\}}] - 2c_l'L_l(\E[|\epsilon|]+\E[|s|])\nonumber\\
=& \E[(l'(x^T\theta^*_{\operatorname{reg}}-x^T\theta + \epsilon+s)-l'(x^T\theta^*_{\operatorname{reg}}-x^T\theta))(x^T\theta^*_{\operatorname{reg}}-x^T\theta)\mathds{1}_{\{|x^T\theta^*_{\operatorname{reg}}-x^T\theta| \geq 2c_l', |\epsilon+s| \geq |x^T\theta^*_{\operatorname{reg}}-x^T\theta|/2\}}]\\
&+\E[(l'(x^T\theta^*_{\operatorname{reg}}-x^T\theta + \epsilon+s)-l'(x^T\theta^*_{\operatorname{reg}}-x^T\theta))(x^T\theta^*_{\operatorname{reg}}-x^T\theta)\mathds{1}_{\{|x^T\theta^*_{\operatorname{reg}}-x^T\theta| \geq 2c_l', |\epsilon+s| \leq |x^T\theta^*_{\operatorname{reg}}-x^T\theta|/2\}}]\\
&-  2c_l'L_l(\E[|\epsilon|]+\E[|s|])\\
\geq& -\frac{9c_l}{64\E[\|x\|^4]}\|\theta-\theta^*_{\operatorname{reg}}\|- 2c_l'L_l(\E[|\epsilon|]+\E[|s|])\\
&+\E[(l'(x^T\theta^*_{\operatorname{reg}}-x^T\theta + \epsilon+s)-l'(x^T\theta^*_{\operatorname{reg}}-x^T\theta))(x^T\theta^*_{\operatorname{reg}}-x^T\theta)\mathds{1}_{\{|x^T\theta^*_{\operatorname{reg}}-x^T\theta| \geq 2c_l', |\epsilon+s| \geq |x^T\theta^*_{\operatorname{reg}}-x^T\theta|/2\}}]\\
\geq & -\frac{9c_l}{64\E[\|x\|^4]}\|\theta-\theta^*_{\operatorname{reg}}\|- 2c_l'L_l(\E[|\epsilon|]+\E[|s|])\\
&-2a_l\E[\|x^T\theta^*_{\operatorname{reg}}-x^T\theta\|\mathds{1}_{\{|x^T\theta^*_{\operatorname{reg}}-x^T\theta| \geq 2c_l', |\epsilon+s| \geq |x^T\theta^*_{\operatorname{reg}}-x^T\theta|/2\}}]\\
\geq & -\frac{9c_l}{64\E[\|x\|^4]}\|\theta-\theta^*_{\operatorname{reg}}\|- 2c_l'L_l(\E[|\epsilon|]+\E[|s|])\\
&-2a_l\E[\|x^T\theta^*_{\operatorname{reg}}-x^T\theta\|\mathds{1}_{\{|x^T\theta^*_{\operatorname{reg}}-x^T\theta| \geq 2c_l'\}}\frac{\E[|\epsilon|]+\E[|s|]}{\|x^T\theta^*_{\operatorname{reg}}-x^T\theta\|}]\\
\geq & -\frac{9c_l}{64\E[\|x\|^4]}\|\theta-\theta^*_{\operatorname{reg}}\|- (2a_l+2c_l'L_l)(\E[|\epsilon|]+\E[|s|]).
\end{align*}
Therefore, together with the bounds for $T_1$ 
and $T_2$, we obtain
\begin{align*}
\langle \nabla f_{\operatorname{reg}}(\theta), \theta-\theta^*_{\operatorname{reg}}\rangle \geq \frac{9c_l}{64\E[\|x\|^4]}\|\theta-\theta^*_{\operatorname{reg}}\|- c,
\end{align*}
where $c = (2a_l+2c_l'L_l)(\E[|\epsilon|]+\E[|s|]) + c_l\Delta_l$
Therefore, for all $\theta \in \R^d$ such that $\|\theta-\theta^*_{\operatorname{reg}}\| \geq \frac{128\E[\|x\|^4]c}{9c_l}$, we have
\begin{align*}
\langle \nabla f_{\operatorname{reg}}(\theta), \theta-\theta^*_{\operatorname{reg}}\rangle \geq \frac{9c_l}{128\E[\|x\|^4]}\|\theta-\theta^*_{\operatorname{reg}}\|.
\end{align*}

If $\frac{128\E[\|x\|^4]c}{9c_l} \leq \frac{\Delta_l-\Delta_{\epsilon,s}}{\Delta_x}$, we have verified the last property in Assumption \ref{assumption:hubor}.

If $\frac{128\E[\|x\|^4]c}{9c_l} > \frac{\Delta_l-\Delta_{\epsilon,s}}{\Delta_x}$, by Lemma \ref{lem:robuststrictlylargerthan0}, we have
\begin{align*}
c_l''=\min_{\theta \in \{\theta\in\R^d:\frac{\Delta_l-\Delta_{\epsilon,s}}{\Delta_x}\leq\|\theta-\theta^*_{\operatorname{reg}}\| \leq \frac{128\E[\|x\|^4]c}{9c_l}\}}\langle \nabla f_{\operatorname{reg}}(\theta), \theta-\theta^*_{\operatorname{reg}}\rangle  >0,
\end{align*}
and for all $\theta\in \R^d$ such that $\|\theta-\theta^*_{\operatorname{reg}}\| \geq \frac{\Delta_l-\Delta_{\epsilon,s}}{\Delta_x}$, we have
\begin{align*}
\langle \nabla f_{\operatorname{reg}}(\theta), \theta-\theta^*_{\operatorname{reg}}\rangle \geq \min(\frac{9c_l}{128\E[\|x\|^4]}, \frac{9c_lc_l''}{128\E[\|x\|^4]c})\|\theta-\theta^*_{\operatorname{reg}}\|,
\end{align*}
thereby completing verifying the last property of Assumption \ref{assumption:hubor}.
\paragraph{Verifying Assumption \ref{assumption:uniformlyorlicz}}
$\forall \theta \in \R^d,$ by inequality \eqref{eq:boundedgradient},  we have
\begin{align*}
\|w(\theta)\| = \|\nabla f_{\operatorname{reg}}(\theta)+l'(y-x^T\theta)x\| \leq a_l\E[\|x\|] + a_l\|x\|.
\end{align*}
Because $\|x\|$ is sub--exponential by Assumption \ref{assumption:data1}, we have verifying the Assumption \ref{assumption:uniformlyorlicz} that $\|w(\cdot)\|$ is uniformly sub--exponential.
\paragraph{Verifying Assumption \ref{assumption:lipnoise}}
$\forall \theta, \theta' \in \R^d$, we have
\begin{align}
\E[\|w(\theta)-w(\theta')\|^2] =& \E[\|\nabla f_{\operatorname{reg}}(\theta)+l'(y-x^T\theta)x-\nabla f_{\operatorname{reg}}(\theta')+l'(y-x^T\theta')x\|^2]\nonumber \\
\leq& 2\|\nabla f_{\operatorname{reg}}(\theta)-\nabla f_{\operatorname{reg}}(\theta')\|^2 + 2\E[\|l'(y-x^T\theta)x-l'(y-x^T\theta')x\|^2]\nonumber\\ 
\overset{\text{(i)}}{\leq}& 4L_l^2\E[\|x\|^4]\|\theta-\theta'\|^2. \label{eq:robustlipnoise}
\end{align}
where (i) holds by following inequality \eqref{eq:robustsmooth} and Assumptions \ref{assumption:data1} and \ref{assumption:l}. Therefore, we have 
\begin{align*}
\E[\|w(\theta)-w(\theta')\|] \leq \sqrt{\E[\|w(\theta)-w(\theta')\|^2]} \leq 2L_l\sqrt{\E[\|x\|^4]}\|\theta-\theta'\|,
\end{align*}
thereby verifying the Assumption \ref{assumption:lipnoise} with $c_w = 2L_l\sqrt{\E[\|x\|^4]}$.
\paragraph{Verifying Assumption \ref{assumption:A2}}
By inequality \eqref{eq:robuststrongconvex}, $\forall \theta', \theta'' \in \{\theta \in \R^d:\|\theta-\theta^*_{\operatorname{reg}}\| \leq \frac{\Delta_l-\Delta_{\epsilon,s}}{\Delta_x}\}$, we have
\begin{align*}
&\E[\|\theta'-\alpha(\nabla f_{\operatorname{reg}}(\theta')+w(\theta') )-\theta''+\alpha(\nabla f_{\operatorname{reg}}(\theta'')+w(\theta'') ) \|]\\
\leq&\sqrt{\E[\|\theta'-\alpha(\nabla f_{\operatorname{reg}}(\theta')+w(\theta') )-\theta''+\alpha(\nabla f_{\operatorname{reg}}(\theta'')+w(\theta'') ) \|^2]}\\
=& \sqrt{\|\theta'-\theta'' - \alpha (\nabla f_{\operatorname{reg}}(\theta')-\nabla f_{\operatorname{reg}}(\theta'')) \|^2+\alpha^2\E[\|w(\theta') -w(\theta'')  \|^2]}\\
\overset{(i)}{\leq}&\sqrt{\|\theta'-\theta'' - \alpha (\nabla f_{\operatorname{reg}}(\theta')-\nabla f_{\operatorname{reg}}(\theta'')) \|^2+4L_l^2\E[\|x\|^4]\alpha^2\|\theta'-\theta''\|^2}\\
\leq&\sqrt{(1-\alpha\mu_l\mathbb{P}(|\epsilon+s|\leq \Delta_{\epsilon,s}) + \alpha^2L_l^2\E[\|x\|^2]^2)\|\theta'-\theta''\|^2 + 4L_l^2\E[\|x\|^4]\alpha^2\|\theta'-\theta''\|^2}\\
\leq&(1-\frac{1}{4}\alpha\mu_l\mathbb{P}(|\epsilon+s|\leq \Delta_{\epsilon,s}))\|\theta'-\theta''\|,
\end{align*}
where (i) holds by following inequality \eqref{eq:robustlipnoise} and the last inequality holds by choosing $\alpha \leq \frac{\mu_l\mathbb{P}(|\epsilon+s|\leq \Delta_{\epsilon,s}))}{2(L_l^2\E[\|x\|^2]^2+4L_l^2\E[\|x\|^4])}$.

Therefore, by the definition of Wasserstein 1-distance \eqref{eq:wasserstein}, we have verified Assumption \ref{assumption:A2}.

\subsection{Proof of The Preliminary Facts}
\subsubsection{Proof of Lemma \ref{lem:x}}
$\forall u \in \R^d$ and $\delta>0$, we have
\begin{align*}
u^T\E[xx^T\mathds{1}_{\|x\| \leq \delta}]u &= u^T\E[xx^T]u-u^T\E[xx^T\mathds{1}_{\|x\| > \delta}]u\\
&= \|u\|^2 - u^T\E[xx^T\mathds{1}_{\|x\| > \delta}]u\\
&\geq \|u\|^2 - \|u\|^2\E[\|x\|^2\mathds{1}_{\|x\| > \delta}]\\
&\geq \|u\|^2 - \|u\|^2\sqrt{\E[\|x\|^4]\mathbb{P}(\|x\|>\delta)}\\
&\geq \|u\|^2 - \|u\|^2\sqrt{2\E[\|x\|^4]\exp(-\delta/\sigma_x)}.
\end{align*}
Therefore, there exists $\Delta_x = \sigma_x\ln(8\E[\|x\|^4])$ such that 
\begin{align*}
u^T\E[xx^T\mathds{1}_{\|x\| \leq \Delta_x}]u \geq \frac{1}{2}\|u\|^2, \forall u \in \R^d.
\end{align*}
\subsubsection{Proof of Lemma \ref{lem:robuststrictlylargerthan0}}
$\forall \theta \in \R^d, \theta \neq 0$, we have
\begin{align*}
\langle \nabla f_{\operatorname{reg}}(\theta), \theta-\theta^*_{\operatorname{reg}}\rangle =& \E[l'(x^T(\theta^*_{\operatorname{reg}
}-\theta) + \epsilon+s)\cdot x^T(\theta^*_{\operatorname{reg}
}-\theta))].
\end{align*}
Let $u = x^T(\theta^*_{\operatorname{reg}
}-\theta)$. By Assumption \eqref{assumption:data1}, we have $u$ is independent with $\epsilon$ and $s$, $\E[u] = 0$, $\E[u^2] = \|\theta-\theta^*_{\operatorname{reg}
}\|^2 >0$ and $u$ is sub--exponential. 

Observe that
\begin{align*}
\E[(l'(u+\epsilon+s)u] &= \E[(l'(u+\epsilon+s)u] - \E[(l'(\epsilon+s)u] \\
&= \E[l'(u+\epsilon+s)-l'(\epsilon+s))((u+\epsilon+s)-\epsilon-s)]\\
&\geq \E[l'(u+\epsilon+s)-l'(\epsilon+s))((u+\epsilon+s)-\epsilon-s)\mathds{1}_{\{|\epsilon+s| \leq \Delta_{\epsilon,s}\}}]\\
&\geq \E[(\mu_lu^2\mathds{1}_{|u| \leq \Delta_l-\Delta_{\epsilon,s}}+\mu_l (\Delta_l-\Delta_{\epsilon,s})^2\mathds{1}_{\{|u| > \Delta_l-\Delta_{\epsilon,s}\}})\mathds{1}_{\{|\epsilon+s| \leq \Delta_{\epsilon,s}\}}]\\
&= P(|\epsilon+s| \leq \Delta_{\epsilon,s}) \E[\mu_lu^2\mathds{1}_{|u| \leq \Delta_l-\Delta_{\epsilon,s}}+\mu_l (\Delta_l-\Delta_{\epsilon,s})^2\mathds{1}_{\{|u| > \Delta_l-\Delta_{\epsilon,s}\}}].
\end{align*}

If $\E[(l'(u+\epsilon+s)u] = 0$, we have $\mathbb{P}(u \neq 0, |u| \leq \Delta_l-\Delta_{\epsilon,s}) = 0$ and $\mathbb{P}(|u| > \Delta_l-\Delta_{\epsilon,s}) = 0,$ which implies $\mathbb{P}(u = 0) = 1.$ By \cite[Proposition 2.16]{folland1999real}, we have $\E[u^2] = 0$, which contradicts with the fact that $\E[u^2] >0.$ Therefore, we have proved that $\E[l'(u+\epsilon+s)u]>0$, thereby finishing the proof of Lemma \ref{lem:robuststrictlylargerthan0}.
\subsubsection{Proof of Lemma \ref{lem:robusthubor}}
By Assumption \ref{assumption:l}, we have $-l'(-\Delta_l) \geq \mu_l\Delta_l>0$ and $l'(\Delta_l)\geq \mu_l\Delta_l>0$. Because $l'(\cdot)$ is increasing, we have
\begin{align*}
tl'(t) \geq \min(-l'(-\Delta_l),l'(\Delta_l))|t| \geq \mu_l\Delta_l|t|,
\end{align*}
thereby completing the proof of Lemma \ref{lem:robusthubor} with $c_l = \mu_l\Delta_l>0.$
\subsubsection{Proof of Lemma \ref{lem:linear}}
\begin{align*}
\E[|x^T\theta|\mathds{1}_{|x^T\theta| \geq a}] =& \E[|x^T\theta|]-\E[|x^T\theta|\mathds{1}_{|x^T\theta| \leq a}]\\
\geq& \E[|x^T\theta|] - a\\
=& \int_{t = 0}^\infty \mathbb{P}(|x^T\theta| \geq t)\d t - a\\
\geq & \int_{t = 0}^{\|\theta\|/2} \mathbb{P}(|x^T\theta| \geq t)\d t - a\\
\geq & \frac{1}{2}\mathbb{P}(|x^T\theta| \geq \|\theta\|/2)\|\theta\| -a\\
= & \frac{1}{2}\mathbb{P}(|x^T\theta|^2 \geq \|\theta\|^2/4)\|\theta\| -a\\
\overset{\text{(i)}}{\geq}& \frac{9}{32} \frac{\E[|x^T\theta|^2]^2}{\E[|x^T\theta|^4]}\|\theta\| - a\\
=& \frac{9}{32} \frac{\|\theta\|^4}{\E[|x^T\theta|^4]}\|\theta\| - a\\
\geq & \frac{9}{32\E[\|x\|^4]}\|\theta\| - a,
\end{align*}
where (i) holds by following Paley–Zygmund inequality \cite{zygmund2002trigonometric}.
\section{Proof of Corollary \ref{co:robustmoments}}
In this section, we verify the Assumption $H_S$ in \cite{gadat2023optimal} for online robust regression. Recall that $w_{\operatorname{reg}}(\theta) = -\nabla f_{\operatorname{reg}}(\theta) - l'(y - x^T\theta)x$. Define $S(\theta) = \E[w_{\operatorname{reg}}(\theta)w_{\operatorname{reg}}(\theta)^T]$ and we aim to prove that $S(\cdot)$ is lip-continuous.

Observe that 
\begin{align*}
S(\theta) &= \E[(-\nabla f_{\operatorname{reg}}(\theta) - l'(y - x^T\theta)x)(-\nabla f_{\operatorname{reg}}(\theta) - l'(y - x^T\theta)x)^T]\\
&= \E[(l'(y - x^T\theta)x)( l'(y - x^T\theta)x)^T] - \nabla f_{\operatorname{reg}}(\theta)\nabla f_{\operatorname{reg}}(\theta) ^T.
\end{align*}
Therefore, for all $\theta,\theta' \in \R^d$, we have
\begin{align*}
&\|S(\theta)-S(\theta)'\|\\
\leq& \|\E[(l'(y - x^T\theta)^2-l'(y - x^T\theta')^2)xx^T]\| + \|\nabla f_{\operatorname{reg}}(\theta)\nabla f_{\operatorname{reg}}(\theta) ^T-\nabla f_{\operatorname{reg}}(\theta')\nabla f_{\operatorname{reg}}(\theta') ^T\|\\
\overset{\text{(i)}}{\leq}& 2a_l\E[\|x\|^3]\|\theta-\theta'\| + \|\nabla f_{\operatorname{reg}}(\theta)(\nabla f_{\operatorname{reg}}(\theta) -  \nabla f_{\operatorname{reg}}(\theta'))^T\|\\
&+\|(\nabla f_{\operatorname{reg}}(\theta) -  \nabla f_{\operatorname{reg}}(\theta'))\nabla f_{\operatorname{reg}}(\theta') ^T\|\\
\overset{\text{(ii)}}{\leq}& (2a_l\E[\|x\|^3]+2a_lL_l\E[\|x\|^2]\E[\|x\|])\|\theta-\theta'\|,
\end{align*}
where (i) holds by Assumption \ref{assumption:l} and (ii) holds by following Theorem \ref{thm:robust}. Therefore, we verify the Assumption $H_S$ in \cite{gadat2023optimal} for online robust regression.

\section{Proof of Theorem \ref{thm:quantile}}\label{sec:quantileproof}
\subsection{A Few Preliminary Facts}
{In this subsection, we present some useful preliminary lemmas. 
The proofs of Lemmas~\ref{lem:quantilestrictlylargerthan0} and~\ref{lem:quantilehubor} 
are similar to those of Lemmas~\ref{lem:robuststrictlylargerthan0} and~\ref{lem:robusthubor}; 
hence, we omit them.}

\begin{lemma}\label{lem:quantilestrictlylargerthan0}
$\forall \theta \in \R^d, \theta \neq \theta^*_\tau$, we have
\begin{align*}
\langle \nabla f_\tau(\theta), \theta-\theta^*_\tau\rangle>0.
\end{align*}
\end{lemma}
\begin{lemma}\label{lem:quantilehubor}
There exists $c_\tau > 0$ such that
\begin{align*}
tF_x(t) \geq c_\tau|t|, \quad \forall x \in \R^d, t \in \{t:|t| \geq \Delta_\tau\}.
\end{align*}
\end{lemma}
\subsection{Main Proof}
In this section, we prove the Theorem \ref{thm:quantile} by verifying the Assumptions \ref{assumption:smooth}-\ref{assumption:uniformlyorlicz}.

\paragraph{Verifying Assumption \ref{assumption:smooth}} 
For all $\theta, \theta' \in \R^d$, by Assumption \ref{assumption:quantile}, we have
\begin{align*}
\|\nabla f_\tau(\theta) - \nabla f_\tau(\theta')\| &=  \|\E[F_x(x^T\theta - x^T\theta^*_\tau)x]-\E[F_x(x^T\theta' - x^T\theta^*_\tau)x]\| \\
&\leq L_\tau\E[\|x\|^2]\|\theta-\theta'\|,
\end{align*}
thereby verifying the Assumption \ref{assumption:smooth} with $L = L_\tau\E[\|x\|^2]$.
\paragraph{Verifying Assumption \ref{assumption:hubor}}
Recall $\Delta_x = \sigma_x\ln(8\E[\|x\|^4])$ defined in Lemma \ref{lem:x}. For all $\theta', \theta'' \in \{ \theta \in \mathbb{R}^d : \|\theta - \theta^*_\tau\| \leq \Delta_\tau/\Delta_x \}$, by Assumption \ref{assumption:quantile}, we have
\begin{align*}
&\langle \theta'-\theta'', \nabla f_\tau(\theta')-\nabla f_\tau(\theta'')\rangle\\
=& \E[(F_x(x^T\theta'-x^T\theta^*)-F_x(x^T\theta''-x^T\theta^*))(x^T\theta'-x^T\theta'')]\\
\geq& \mu_\tau\E[(x^T\theta'-x^T\theta'')^2\mathds{1}_{\{|x^T\theta'-x^T\theta^*| \leq \Delta_\tau , |x^T\theta''-x^T\theta^*| \leq \Delta_\tau\}}]\\
\geq& \mu_\tau (\theta'-\theta'')^T\E[xx^T\mathds{1}_{\{\|x\| \leq \Delta_x \}}](\theta'-\theta'')\\
\geq& \frac{\mu_\tau
}{2}\|\theta'-\theta''\|^2.
\end{align*}
For all $ \theta  \in \R^d$, we have
\begin{align*}
\|\nabla f_\tau(\theta)\| = \|\E[(F_x(x^T\theta - x^T\theta^*_\tau)-\tau)x]\| \leq (1+\tau)E[\|x\|].
\end{align*}
To verify the last property in Assumption \ref{assumption:hubor}, {by Lemma \ref{lem:quantilehubor}}, when $\theta \neq \theta^*_\tau$, we have
\begin{align*}
\langle \theta - \theta^*_\tau , \nabla f_\tau(\theta)\rangle &= \langle \E[(\mathds{1}_{\{y-x^T\theta \leq 0\}}-\tau)x], \theta-\theta^*_\tau \rangle\\
&= \E[(F_x(x^T\theta-x^T\theta^*_\tau)-F_x(0))(x^T\theta-x^T\theta^*_\tau)]\\
&\geq \E[(F_x(x^T\theta-x^T\theta^*_\tau)-F_x(0))(x^T\theta-x^T\theta^*_\tau)\mathds{1}_{\{|x^T\theta-x^T\theta^*_\tau|\geq \Delta_\tau\}}]\\
&\geq c_\tau\E[|x^T\theta-x^T\theta^*_\tau|\mathds{1}_{\{|x^T\theta-x^T\theta^*_\tau|\geq \Delta_\tau\}}]\\
&\geq \frac{9c_\tau}{32\E[\|x\|^4]}\|\theta-\theta^*_\tau\| - c_\tau\Delta_\tau,
\end{align*}
where the last inequality holds by Lemma \ref{lem:linear}. 

Therefore, when $\|\theta-\theta^*_\tau\| \geq \frac{64\E[\|x\|^4]}{9\Delta_\tau}$, we have
\begin{align*}
\langle \theta - \theta^*_\tau , \nabla f_\tau(\theta)\rangle  \geq \frac{9c_\tau}{64\E[\|x\|^4]}\|\theta-\theta^*_\tau\|.
\end{align*}

If $\frac{64\E[\|x\|^4]}{9\Delta_\tau} \leq \frac{\Delta_\tau}{\Delta_x}$, we have verified the last property in Assumption \ref{assumption:hubor}.

If $\frac{64\E[\|x\|^4]}{9\Delta_\tau} > \frac{\Delta_\tau}{\Delta_x}$, by Lemma \ref{lem:quantilestrictlylargerthan0}, we have
\begin{align*}
c_\tau'=\min_{\theta \in \{\theta\in\R^d:\frac{\Delta_\tau}{\Delta_x} \leq\|\theta-\theta^*_\tau\| \leq \frac{64\E[\|x\|^4]}{9\Delta_\tau}\}}\langle \nabla f_\tau(\theta), \theta-\theta^*_\tau \rangle  >0,
\end{align*}
and for all $\theta\in \R^d$ such that $\|\theta-\theta^*_\tau\| \geq \frac{\Delta_\tau}{\Delta_x}$, we have
\begin{align*}
\langle \nabla f_\tau(\theta), \theta-\theta^*_\tau \rangle \geq \min(\frac{9c_\tau}{64\E[\|x\|^4]}, \frac{9c_l'\Delta_\tau}{64\E[\|x\|^4]})\|\theta-\theta^*_\tau\|,
\end{align*}
thereby verifying the last property of Assumption \ref{assumption:hubor}.
\paragraph{Verifying Assumption \ref{assumption:uniformlyorlicz}}By the definition in Section \ref{sec:quantile}, we have
\begin{align*}
\|w_\tau(\theta)\| = \|(\mathds{1}_{\{y-x^T\theta \leq 0\}}-\tau)x - \nabla f_\tau(\theta)\| \leq 2\|x\| + (1+\tau)\E[\|x\|],
\end{align*}
which implies that the noise sequence is uniformly in the $\psi_1-$Orlicz space.

\section{Discussions on the Lack of Weak Converge for Quantile Regression}\label{sec:quantileweak}
In this section, we focus on the recursive quantile estimation problem, which is a special case of online quantile regression when $x = 1$. We try to illustrate that it is not clear if constant stepsize recursive quantile estimation has the weak convergence results.

The constant stepsize recusive quantile estimation has the following update
\begin{align*}
\theta_{n+1} = \theta_n - \alpha ( \mathds{1}_{\{ y_n   \leq \theta_n \}} - \tau ).
\end{align*}
Because $\theta_n \in \R$ for all $n \geq 0$, we use the following formula to calculate the Wasserstain-1 distance:
\begin{align*}
W_1\left(\mu_1, \mu_2\right)=\int_{\mathbb{R}}\left|F_1(x)-F_2(x)\right| \mathrm{d} x,
\end{align*}
where $F_i$ denotes the cumulative probability function of $\mu_i$ for $i = 1,2.$ 

Without loss of generality, we assume that $\theta'< \theta< \theta'+\alpha/4$ and $\theta, \theta' \in [-\Delta_\tau,\Delta_\tau]$. We have
\begin{align*}
&W_1(\law(\theta - \alpha ( \mathds{1}_{\{ X \leq \theta \}} - \tau )), \law(\theta' - \alpha ( \mathds{1}_{\{ X \leq \theta' \}} - \tau ))) \\
=& (\theta-\theta') F(\theta') + (\theta' - \theta+\alpha  )(F(\theta)-F(\theta')) + (\theta-\theta')(1-F(\theta))\\
=& \theta-\theta' + (\alpha-2 (\theta-\theta'))(F(\theta)-F(\theta'))\\
\geq& (\theta-\theta') + \alpha/4(F(\theta)-F(\theta'))\\
\geq& (1+\alpha\mu_\tau/4)(\theta-\theta').
\end{align*}
Consequently, we demonstrate that two iterates near zero diverge in the \( W_1 \) distance, indicating that the weak convergence result may not hold for recursive quantile estimation. However, this does not rule out the possibility of weak convergence under a metric weaker than \( W_1 \), which we leave for future work.

\section{Proof of Corollary \ref{co:quantilemoments}}\label{sec:quantilehs}
In this section, we verify the Assumption $H_S$ in \cite{gadat2023optimal} for online quantile regression. Recall that $w_\tau(\theta) = (\mathds{1}_{\{y-x^T\theta \leq 0\}}-\tau)x - \nabla f_\tau(\theta)$. Define $S(\theta) = \E[w_\tau(\theta)w_\tau(\theta)^T]$ and we aim to prove that $S(\cdot)$ is lip-continuous.

Notice that 
\begin{align*}
S(\theta) &= \E[((\mathds{1}_{\{y-x^T\theta \leq 0\}}-\tau)x - \nabla f_\tau(\theta))((\mathds{1}_{\{y-x^T\theta \leq 0\}}-\tau)x - \nabla f_\tau(\theta))^T]\\
&= \E[(\mathds{1}_{\{y-x^T\theta \leq 0\}}-\tau)^2xx^T] - \nabla f_\tau(\theta)\nabla f_\tau(\theta)^T\\
&= \E[(1-2\tau)F_x(x^T\theta - x^T\theta^*_\tau)xx^T] + \tau^2\E[xx^T]- \nabla f_\tau(\theta)\nabla f_\tau(\theta)^T.
\end{align*}
Therefore, for all $\theta, \theta' \in \R$, we have
\begin{align*}
&\|S(\theta)-S(\theta')\| \\
\leq& (1-2\tau)\E[|F_x(x^T\theta - x^T\theta^*_\tau)-F_x(x^T\theta' - x^T\theta^*_\tau)|\|x\|^2]+ \|\nabla f_\tau(\theta)\nabla f_\tau(\theta)^T-\nabla f_\tau(\theta')\nabla f_\tau(\theta')^T\|\\
\leq& (1-2\tau)L_\tau\E[\|x\|^3]\|\theta-\theta'\| + \|\nabla f_\tau(\theta)\|\|\nabla f_\tau(\theta)-\nabla f_\tau(\theta')\| + \|\nabla f_\tau(\theta')\|\|\nabla f_\tau(\theta)-\nabla f_\tau(\theta')\|\\
\leq & ((1-2\tau)L_\tau\E[\|x\|^3]+2(1+\tau)L_\tau\E[\|x\|]\E[\|x\|^2])\|\theta-\theta'\|
\end{align*}
where the last inequality holds by Theorem \ref{thm:quantile}. Therefore, we verify the Assumption $H_S$ in \cite{gadat2023optimal} for online quantile regression.

\end{document}